\documentclass[11pt]{article}

\usepackage[T1]{fontenc}
\usepackage[utf8]{inputenc}
\usepackage[english]{babel}

\usepackage[sectionbib]{natbib}

\usepackage{graphicx}
\usepackage{subcaption}
\usepackage{caption}
\usepackage{float}
\usepackage{placeins}
\usepackage{pdfpages} 

\usepackage{geometry}
\usepackage{setspace}
\usepackage{booktabs}
\usepackage{multirow}
\usepackage{xcolor}
\usepackage{import}
\usepackage{afterpage} 
\usepackage{emptypage} 

\usepackage[reqno,tbtags,sumlimits]{amsmath}
\usepackage{amssymb}
\usepackage{amsthm}
\usepackage{amsxtra}
\usepackage{mathrsfs}
\usepackage{mathtools}
\usepackage{bigints}
\usepackage{empheq}  
\usepackage[mathcal,mathscr]{eucal}
\usepackage{siunitx}

\usepackage{hyperref}
\usepackage{xurl} 
\usepackage{microtype} 
\usepackage{blindtext} 

\theoremstyle{remark}

\theoremstyle{definition}

\renewcommand{\epsilon}{\varepsilon}
\renewcommand{\theta}{\vartheta}
\renewcommand{\phi}{\varphi}

\usepackage{titlesec}

\newcount\Comments  
\usepackage{color}
\definecolor{darkgreen}{rgb}{0,0.5,0}
\definecolor{purple}{rgb}{1,0,1}
\newcommand{\kibitz}[2]{\ifnum\Comments=1\textcolor{#1}{#2}\fi}

\title{\huge A Machine Learning Approach to Forecasting Honey Production with Tree-Based Methods\\ \vspace{5mm}}

\usepackage{authblk}
\author[1\footnote{Corresponding author}]{Alessio Brini}
\author[2]{{Elisa Giovannini}}
\author[3]{Elia Smaniotto}

\affil[1]{\footnotesize{Duke University Pratt School of Engineering, 305 Teer Engineering Building Box 90271, Durham, NC 27708 (USA). E-mail: alessio.brini@duke.edu}}
\affil[2]{\footnotesize{Department of Economics and Management, University of Florence, Via delle Pandette 32, 50127 Firenze FI (Italy). E-mail: elisa.giovannini@unifi.it}}
\affil[3]{\footnotesize{DiMSEFA, Catholic University of the Sacred Heart, via Necchi 9, 20123 Milan, MI, (Italy). E-mail: elia.smaniotto@unicatt.it}}

\date{}

\begin{document}
\maketitle

\section*{Abstract}
The beekeeping sector has experienced significant production fluctuations in recent years, largely due to increasingly frequent adverse weather events linked to climate change. These events can severely affect the environment, reducing its suitability for bee activity. We conduct a forecasting analysis of honey production across Italy using a range of machine learning models, with a particular focus on weather-related variables as key predictors. Our analysis relies on a dataset collected in 2022, which combines hive-level observations with detailed weather data. We train and compare several linear and nonlinear models, evaluating both their predictive accuracy and interpretability. By examining model explanations, we identify the main drivers of honey production. We also ensemble models from different families to assess whether combining predictions improves forecast accuracy. These insights support beekeepers in managing production risks and may inform the development of insurance products against unexpected losses due to poor harvests.

\vspace{3mm}

\noindent
\textbf{Keywords}: machine learning, tree-based models, hive weight forecasting, climate risk, beekeeping analytics, precision agriculture


\section{Introduction}

Honeybees are essential pollinators, playing a vital role in ecosystems and supporting food production by enabling the reproduction of many crop species. While staple cereals such as wheat and rice are primarily self-pollinated, pollinators still influence the yield and quality of over 70\% of the world’s most important crops, contributing to an estimated 35\% of the global human food supply \citep{TSCHARNTKE2012}. The estimated annual economic value of honeybee pollination amounts to several billion dollars annually, especially due to the key role in enhancing agriculture production and ensuring plant reproduction \citep{delaplane2000crop,millennium2005ecosystems,klein2007importance,garibaldi2014research,fao2018bees}. Therefore, the decline in honey production threatens food security, as less pollination leads to a reduced crop yield. Besides their role in crop pollination, honeybees are essential producers of honey, which is widely consumed for its health benefits. Honey production contributes significantly to the economy, with the global honey market valued at approximately USD 8.94 billion in 2023, according to a report by Fortune Business Insights. In Europe, the honey market generated revenues of around USD 3.36 billion in the same year, as estimated by Grand View Research.

Europe represents a major hub for honey production and trade.  With 20 million beehives and 218000 tons in 2022\footnote{\url{https://agriculture.ec.europa.eu/farming/animal-products/honey\_en}}, the European Union is the second largest honey producer after China. However, the former also imports a surplus of around 40\% of the amount of honey produced to cover domestic consumption, making imports greater than exports.
The largest honey production is mainly located in Southern Europe, where climatic conditions are more favorable to beekeeping, as reported by the European Commission.

Honey production depends on three major categories: climate, pests and diseases, and beekeeping practices. For instance, honeybees need adequate forage and water to produce honey, and a lack of either can result in reduced production. Additionally, changes in temperature and rainfall patterns can affect flower blooming, leading to a decline in nectar and pollen production. Pests and diseases such as Varroa mites also negatively impact honey production. These external parasites feed on honeybees, weakening their immune systems and increasing their vulnerability to infections. The use of pesticides and insecticides in agriculture also contributes to the decline in honeybee populations, as they can kill bees and disrupt their behavior. Beekeeping practices can also influence honey production. Proper hive management, such as regular inspections and adequate feeding, can increase honey production. On the other hand, poor management practices, such as overcrowding and improper ventilation, can lead to hive diseases and a decline in honey production \citep{underwood2019beekeeping}.

In this paper, we approach the study of honey production drivers by focusing specifically on the climate effect. It is undoubtedly clear that climate change will cause major modifications to the depicted framework of honey production\citep{holmes_2002-weatherinfluence, gordo2006temporal, leconte2008climate, switanek2017modelling, flores2019climate-bees, solovev2020influence, calovi2021}. Due to its geographical location, mainly related to climatic sensitivity and diverse microclimates, Italy is one of the most affected countries as Italian beekeepers recorded substantial variations in honey production with losses of up to 70\% in some regions \citep{porrini2016status,gray2019}. 
It is essential to understand the climate aspects to gain insights into the beekeeping system's efficiency to decrease the risk of losses and maximize their contribution to the ecosystem. Extended periods of rain and sudden temperature increases have severe impacts on spring plants and bees' health, implicitly causing an impact on total honey production. 

The contributions of this paper are threefold:
\begin{itemize}
    \item We analyze the key weather-related drivers of honey production using interpretable machine learning models. Understanding these drivers can help beekeepers better manage their operations and anticipate periods of reduced productivity.
    
    \item By forecasting production outcomes, we provide a quantitative tool that can support beekeepers in making critical decisions, such as where and when to relocate hives to mitigate the impact of adverse weather conditions.
    
    \item The analysis also has applications in risk management. Forecasted honey production can inform the design of machine learning-based insurance products aimed at protecting beekeepers from unexpected losses due to poor harvests.
\end{itemize}

Recent technological advances make these contributions possible and enable the continuous monitoring of beehive conditions and the collection of large-scale data suitable for quantitative analysis. Apiculture, the technical term for beekeeping, has increasingly adopted precision beekeeping technologies, a branch of precision agriculture focused on improving apiary management. These systems use smart, connected devices to track individual colonies and generate detailed datasets that support both predictive modeling and strategic decision-making \citep{zachepins2012precisionbeeekeeping}.

Our analysis leverages the data from an Italian technology company, 3Bee S.R.L. \footnote{\url{https://www.3bee.com/en/}}. The company develops intelligent monitoring and diagnostic systems for bee health, bringing together numerous beekeepers. The analyzed dataset includes daily hive-level observations from 209 smart beehives deployed across Italy during the 2022 honey season, collected through 3Bee’s precision beekeeping technology. We integrate this data with daily weather features extracted from the ERA5-Land reanalysis dataset by Copernicus, aligning meteorological conditions with hive locations to effectively model the impact of climate variability on honey production.

We approach the forecasting task using a diverse set of predictive models, including linear regression and its regularized variants, ensemble tree-based methods, and a feedforward neural network. This broad landscape allows us to assess both linear and nonlinear relationships in the data, capturing interactions that simpler statistical models may miss. In particular, tree-based models such as Random Forest, XGBoost, and LightGBM strike a favorable balance between forecasting accuracy and interpretability, making them well-suited for analyzing complex patterns in hive dynamics. We further explore ensemble strategies that combine predictions from models across different families to enhance forecast accuracy. Beyond prediction, we analyze model explainability to identify the key features, especially weather-related variables, that drive honey production, offering actionable insights for beekeepers and informing the design of risk management tools. Among the evaluated models, out-of-sample performance is led by Random Forest and a highly regularized linear model such as Elastic Net across multiple evaluation metrics, although other tree-based ensembles remain competitive.

The remainder of the paper is organized as follows. Sec.~\ref{Sec:literature} reviews the existing literature on modeling and forecasting honey production. Sec.~\ref{sec:dataset} introduces the dataset and outlines the feature engineering approach. Sec.~\ref{sec:models} presents the predictive models used in our analysis, while Sec.~\ref{sec:empirical} discusses the empirical findings. In Sec.~\ref{sec:interpretability}, we interpret the models through feature importance analyses, and in Sec.~\ref{Sec:ensemble}, we evaluate ensemble strategies that combine predictions across model families. Finally, Sec.~\ref{sec:conclus} concludes the paper and highlights potential directions for future work.

\section{Related Literature}\label{Sec:literature}
The effect of weather and environment on beehive weight variations has been studied for over a century \citep{hambleton}. Many works relate the influence of seasonal weather conditions on honey productivity and the health conditions of bees. \cite{szabo1980effect-taylorfrancis} find a positive correlation between the bees' flight activity and the temperature. \cite{holmes_2002-weatherinfluence} perform a multivariate regression analysis involving 21 variables to investigate the factors influencing honey production.
\cite{bhusal2006response} study honey production using a Randomized Complete Block Design, an experimental setup commonly used in agriculture to control for variability across groups and isolate the effects of key production factors. \cite{flores2019climate-bees,catania2020application, gounari2022greece} provide a more recent outlook on climate change impacts on bees' activity in the Mediterranean area. Over the years, the evolution of technologies and data availability has allowed the development of sophisticated statistical methods for studying bees' behavior, pollen foraging, and weather impact. \cite{clarke2018predictive} investigate the relationship between the foraging activity of honey bees and local weather conditions in the United Kingdom using generalized least squares regression, whereas \cite{karaboga2011novel} employ a cluster analysis for simulating the intelligent foraging behavior of a honey bee swarm. \cite{nasr2014cluster} conduct an additional cluster analysis to estimate factors related to honey production. \cite{overtuf2022regression} conduct a Canada-based research demonstrating a strong correlation between winter weather and honey losses with standard regression and spatial analysis. \cite{becsi2021biophysical} present a novel approach to quantify the effects of weather conditions on Austrian honey bee colony winter mortality by defining 
biophysics-based weather indicators. \cite{dainat} study the spread of Varroa infestation and its correlation with high bee mortality under specific temperature conditions. Other works estimate honey production through spatial regression \citep{tassinari2013spatial}, fuzzy inference methods \citep{hastono2017fuzzy}.

Several studies employ alternative machine learning approaches to estimate honey production. These include clustering algorithms \citep{RAFAELBRAGA2020}, k-nearest neighbors \citep{yesugade2018machine}, and tree-based methods such as decision trees and ensemble models \citep{calovi2021, quinlan2022grassy, braga2020}. \cite{Karadas2017} compare the predictive performance of several data mining algorithms and neural networks to identify the main factors influencing average honey yield per beehive. \cite{ALVES2020} adopt convolutional neural networks to detect cells in comb images and classify their contents into seven categories, including cells occupied by eggs, larvae, capped brood, pollen, nectar, and honey. \cite{campbell2020} use regression trees to estimate the honey harvests in South West Australia based on weather and vegetation-related information obtained from satellite sensors. \cite{ngo2021} show the correlation between environmental data and pollen foraging with a neural network-powered imaging system, emphasizing that temperature, relative humidity, wind speed, rain level, and light intensity influence colony activity. 

Despite a growing body of research on honeybee activity and its environmental drivers, a comprehensive modeling analysis focused on the Italian context is still lacking. Our contribution addresses this gap by comparing a wide range of model families, including linear models, tree ensembles, and neural networks, for forecasting hive-level honey production. We emphasize tree-based methods due to their strong performance on structured tabular data and ability to capture nonlinear relationships, even in relatively low-dimensional settings. These models balance predictive accuracy and interpretability, offering a valuable alternative to traditional linear techniques and more data-hungry neural network approaches. Through this comparative framework, we aim to uncover the key drivers of honey production in Italy and provide actionable insights for for beekeepers and stakeholders supporting their interests, such as insurers and policy advisors.

\section{Data Description and Feature Engineering} \label{sec:dataset}

We employ a dataset that combines two distinct sources: data from precision beekeeping technologies and meteorological data. Precision beekeeping is a growing branch of agriculture that aims to optimize resource allocation and maximize bee productivity through connected smart beehives \citep{catania2020application, cardell2022webee, PB2022precisionbeekeeping}. These technologies, coupled with the voluntary collaboration of beekeepers, enable the collection of extensive data on hive conditions.

To complement the beekeeping data, we incorporate meteorological data sourced from the Copernicus Climate Data Store\footnote{\url{https://cds.climate.copernicus.eu/}}, which provides high-quality weather information. We align these datasets to enable a comprehensive analysis of how environmental factors influence hive conditions and productivity.

The following subsections describe the two data sources and explain their integration in detail.

\subsection{Beehive Data}\label{subsec:hive_data}

We obtain precision beekeeping data from 3Bee, an agri-tech company specializing in intelligent monitoring devices and bee health diagnostic systems. Their technology allows beekeepers to gather real-time information about their hives, enhancing production efficiency and minimizing risks related to disease and other issues. The 3Bee 2022 dataset contains unevenly distributed intraday hive weight recordings. We select 209 hives suitable for analysis and follow a rigorous preprocessing and filtering procedure, as described below.

To construct the weight time series, we focus on the period from April 1, 2022, to September 30, 2022, corresponding to the main honey production season during spring and summer. We resample the weight recordings to a daily frequency, using the last available observation between 5:00 p.m. and 8:00 p.m. Central European Time (CET) as the daily weight measurement. If no observation falls within this time window, we treat the day as missing.

We apply additional preprocessing steps to refine the dataset:
\begin{itemize}
    \item We exclude hives with missing daily observations throughout the spring–summer production period.
    \item We replace spurious weight measurements below 20 kilograms (the minimum expected weight of the wooden hive structure) with a smoothed value computed using an exponentially weighted moving average (EWMA) with a smoothing factor $\alpha = 0.5$. The EWMA is calculated recursively over the previous three days, giving more weight to recent observations.
    \item We flag daily weight variations exceeding 1 kg in absolute value as potential measurement errors and replace them with a smoothed value computed using an EWMA of the previous three days’ hive weight variations. The smoothing factor is set to $\alpha = 0.5$, applying the same recursive formulation as in the previous step.

\end{itemize}

\begin{figure}[t]
\centering
\includegraphics[width=0.65\textwidth]{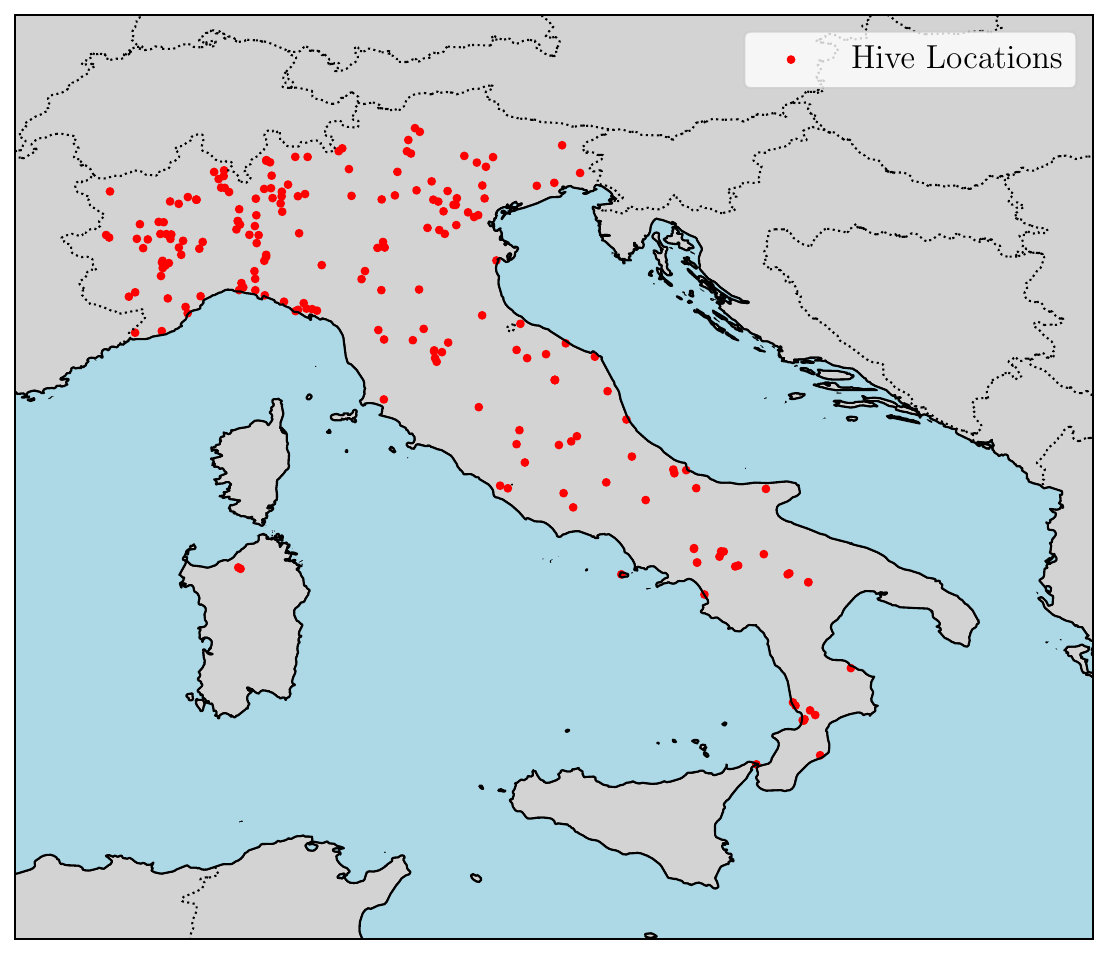} 
\caption{Geographical distribution of the hives over the Italian territory.}
\label{Fig:map}
\end{figure}

\begin{figure}[t]
    \centering
    \begin{subfigure}{0.35\textwidth}
        \centering
        \includegraphics[width=\linewidth]{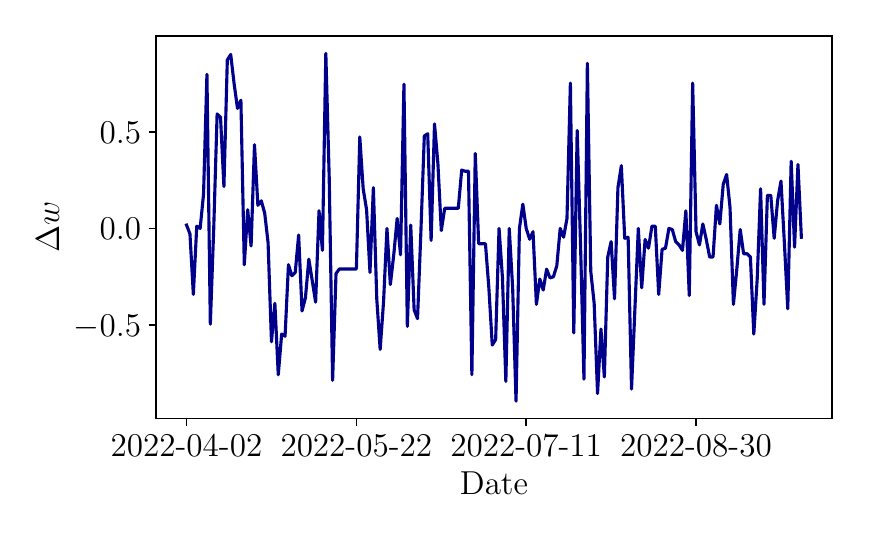}
    \end{subfigure}
    \hspace{0.04\textwidth}
    \begin{subfigure}{0.35\textwidth}
        \centering
        \includegraphics[width=\linewidth]{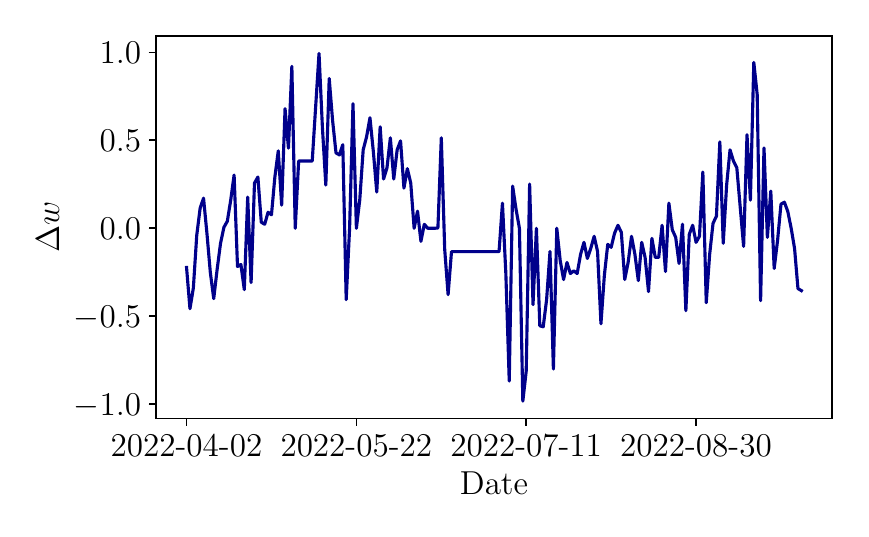}
    \end{subfigure}
    
    \vspace{0.5em}
    
    \begin{subfigure}{0.35\textwidth}
        \centering
        \includegraphics[width=\linewidth]{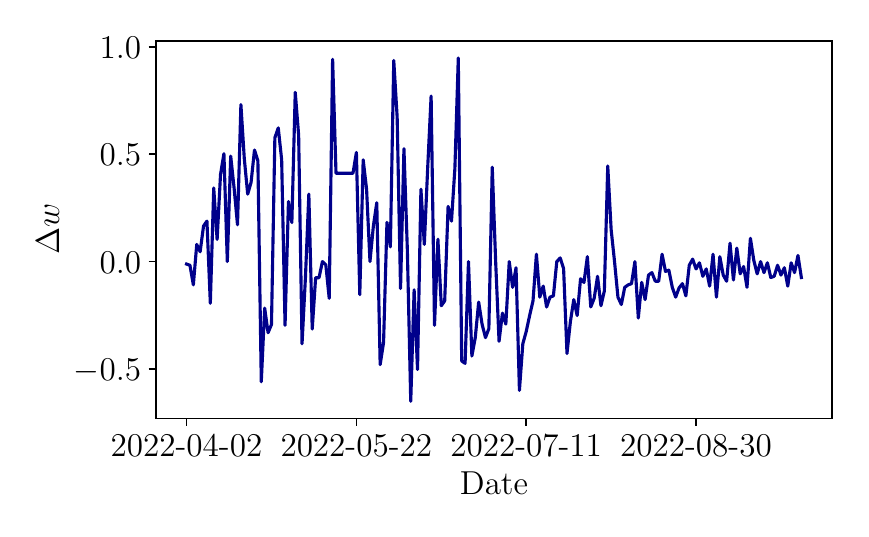}
    \end{subfigure}
    \hspace{0.04\textwidth}
    \begin{subfigure}{0.35\textwidth}
        \centering
        \includegraphics[width=\linewidth]{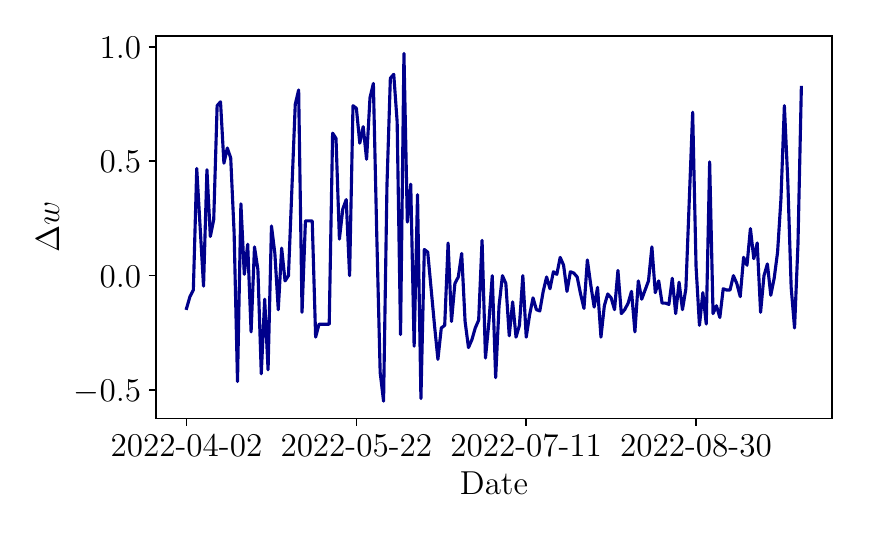}
    \end{subfigure}
    
    \caption{Daily weight variation $\Delta w$ for a random selection of four hives.}
    \label{Fig:hive_diffw_examples}
\end{figure}

Finally, we test the daily hive weight variation series for stationarity using the Augmented Dickey-Fuller (ADF) test at the 5\% significance level. Test results indicate that all series are stationary at the 5\% level. Fig. \ref{Fig:hive_diffw_examples} illustrates examples of the preprocessed daily hive weight variation series used in the analysis. Fig. \ref{Fig:map} visualizes the geographical position of these hives across the Italian territory, with most hives located in the northern regions of the country. We also include latitude and longitude as input features in the forecasting models. Since beekeepers can relocate hives over time, these coordinates are subject to change, though such movements typically involve short-range transitions across neighboring regions within the country.

\FloatBarrier

\FloatBarrier

\subsection{Weather Data}\label{subsec:weather_data}

We source meteorological data from the ERA5-Land reanalysis dataset,\footnote{The ERA5-Land dataset is an open-access resource developed by the European Centre for Medium-Range Weather Forecasts (ECMWF) as part of the EU-funded Copernicus Climate Change Service (C3S). It provides high-quality hourly weather estimates \citep{ERA5-Land2021, copernicus2019, copernicus2021}.} which is available through the Copernicus Climate Data Store.\footnote{\url{https://cds.climate.copernicus.eu/datasets/reanalysis-era5-single-levels?tab=overview}} Reanalysis data combine past short-range weather forecasts with modern numerical weather prediction models, addressing the lack of uniformly distributed meteorological stations across the considered territory.

We extract hourly weather data for each of the 209 hives using their average geographical coordinates (latitude and longitude) over the observation period to account for potential hive relocations. Tab. \ref{Tab:weather_variables} lists the extracted variables, along with the descriptive statistics computed when resampling the hourly information to daily frequency. This resampling aligns the meteorological data with the beehive dataset to ensure compatibility for analysis. We refer to the ERA5-Land documentation\footnote{\url{https://confluence.ecmwf.int/display/CKB/ERA5\%3A+data+documentation}} for the unit of measure and detailed definitions of each variable extracted.

\begin{table}[h]
\begin{tabular}{lc}
\toprule
\textbf{Weather Variable} & \textbf{Aggregated Features} \\
\midrule
Dewpoint temperature      & Mean, Min, Max              \\
Temperature at 2 meters   & Mean, Min, Max              \\
Surface pressure          & Mean                        \\
Total precipitation       & Sum                         \\
Maximum temperature at 2 meters & Max                  \\
Minimum temperature at 2 meters & Min                  \\
Surface solar radiation   & Sum, Mean                   \\
Maximum temperature precipitation & Max, Mean          \\
Minimum temperature precipitation & Min, Mean          \\
\bottomrule
\end{tabular}
\centering
\caption{Weather Variables and Aggregated Features}
\label{Tab:weather_variables}
\end{table}

\subsection{Data Exploration and Feature Engineering}

\begin{figure}[h]
    \centering
    \includegraphics[width=0.9\textwidth]{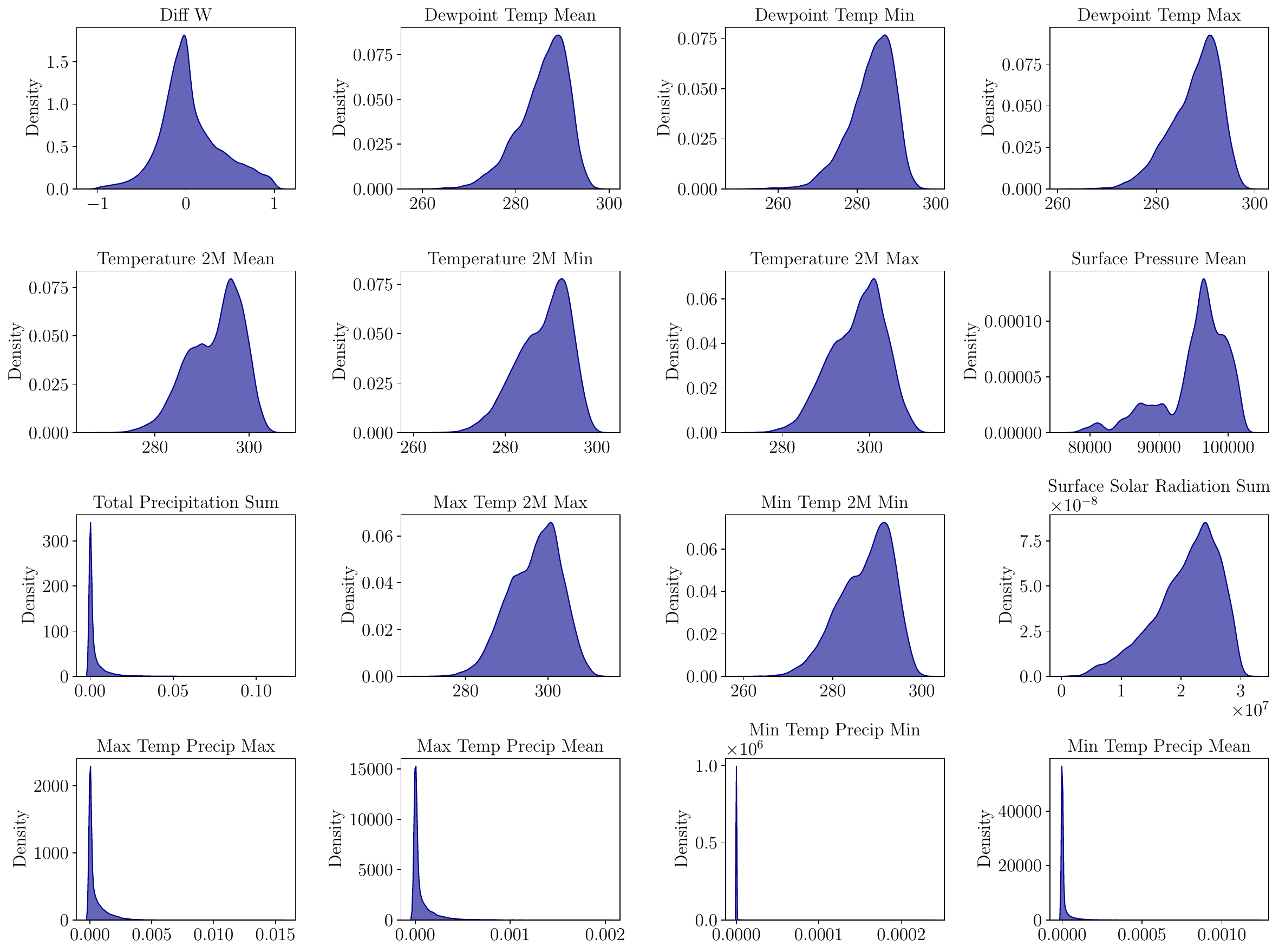}
    \caption{Empirical densities of the features outlined in Tab \ref{Tab:descriptive_stats}.}
    \label{Fig:densities_features}
\end{figure}

After integrating the hive weight and meteorological datasets, we obtain 18 variables. These variables are outlined in Tab. \ref{Tab:descriptive_stats}, which provides their numerical descriptive statistics, and in Fig. \ref{Fig:densities_features}, which displays their empirical density distributions. We compute these statistics across the cross-sectional dataset, aggregating both hive IDs and time.

To further analyze the statistical properties of the variables, we assess their stationarity using the ADF test at the 5\% level. Tab.~\ref{Tab:stationary} reports the percentage of hive IDs for which each variable passes the stationarity test. When more than 40\% of hives fail the test for a given variable, we attribute this to the intrinsic properties of the measurement, such as seasonality or structural trends, that may affect its statistical behavior.

Temperature-related variables, for instance, often exhibit clear trends due to seasonal cycles and external climatic factors. Differencing these variables would remove meaningful variations that may be crucial for predictive modeling. For this reason, we preserve their raw values in the dataset.

Precipitation-related variables present a different challenge. Their distributions are typically sparse with occasional peaks, and differencing may not necessarily improve their statistical properties. As with temperature, we retain these variables in their raw form, relying on the models’ ability to handle zero-inflated and skewed inputs effectively.

\begin{figure}[h]
    \centering
    \includegraphics[width=0.9\textwidth]{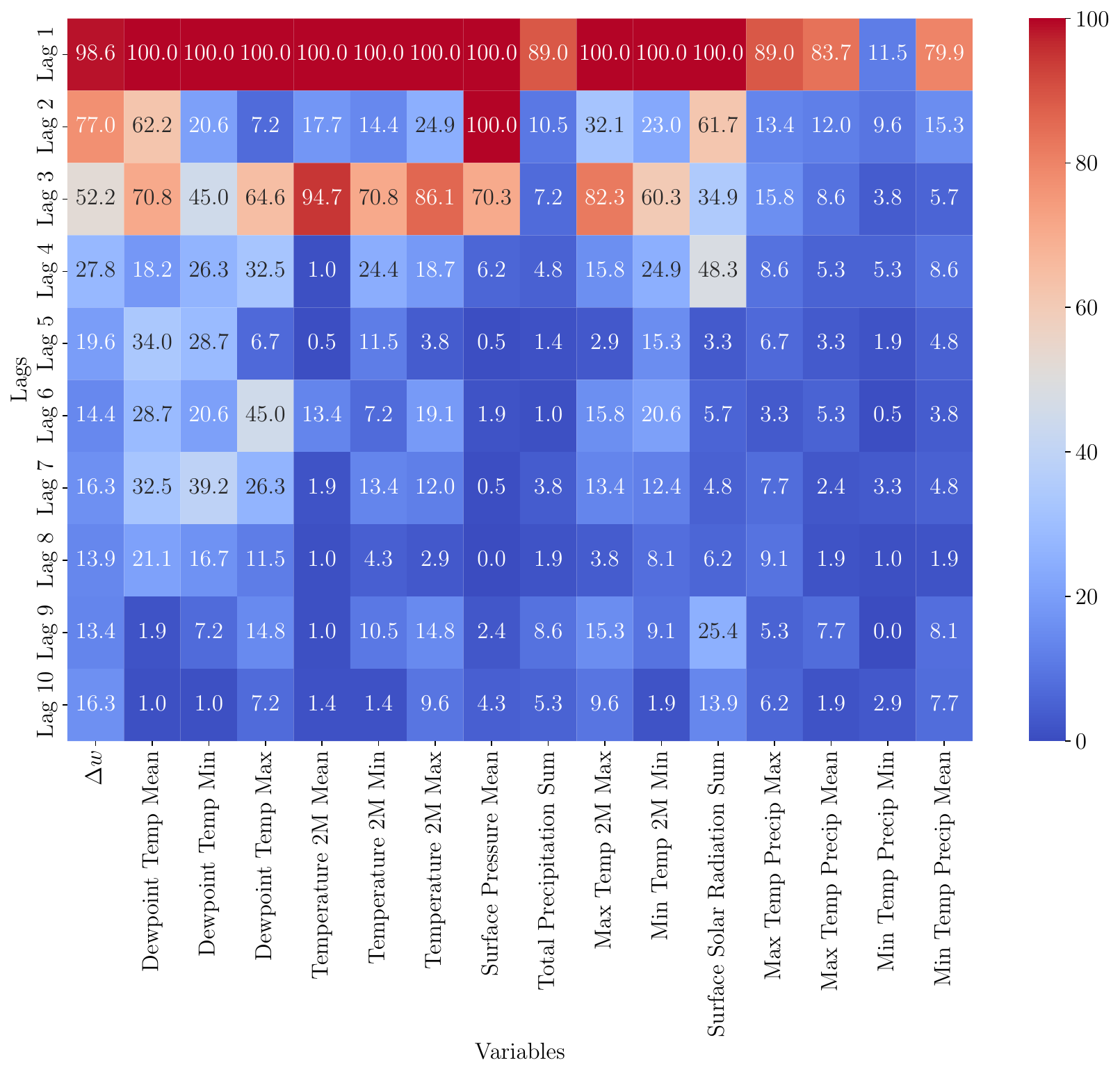}
    \caption{Heatmap of the percentage of variables that exhibit autocorrelation at different lags. Each entry of the matrix represents the percentage of hive IDs, measured over the cross-section, that exhibit such linear autocorrelation.}
    \label{fig:acorr_matrix}
\end{figure}

Lastly, we examine the presence of autocorrelation in the dataset. Fig.~\ref{fig:acorr_matrix} presents a heatmap showing the percentage of hive IDs for which each variable exhibits significant autocorrelation at different daily lags. This analysis provides critical insight into temporal dependencies and guides the selection of lagged predictors in the forecasting models. Based on these findings, in our empirical analysis, we explore models incorporating lags structures ranging from one to four time steps for all variables.

\begin{table}[h]
\centering
\begin{tabular}{lllllll}
\toprule
 & Mean & Std. Dev. & Min & 25\% & 75\% & Max \\
\midrule
$\Delta w$ & 4.77e-02 & 3.44e-01 & -1.00e+00 & -1.60e-01 & 2.29e-01 & 1.00e+00 \\
Dewpoint Temp Mean & 2.86e+02 & 5.26e+00 & 2.59e+02 & 2.83e+02 & 2.90e+02 & 2.98e+02 \\
Dewpoint Temp Min & 2.84e+02 & 5.72e+00 & 2.51e+02 & 2.80e+02 & 2.88e+02 & 2.97e+02 \\
Dewpoint Temp Max & 2.88e+02 & 5.01e+00 & 2.62e+02 & 2.85e+02 & 2.92e+02 & 2.99e+02 \\
Temperature 2M Mean & 2.93e+02 & 5.70e+00 & 2.68e+02 & 2.89e+02 & 2.97e+02 & 3.06e+02 \\
Temperature 2M Min & 2.88e+02 & 5.67e+00 & 2.62e+02 & 2.84e+02 & 2.93e+02 & 3.01e+02 \\
Temperature 2M Max & 2.97e+02 & 6.18e+00 & 2.72e+02 & 2.93e+02 & 3.02e+02 & 3.12e+02 \\
Surface Pressure Mean & 9.53e+04 & 4.72e+03 & 7.74e+04 & 9.39e+04 & 9.86e+04 & 1.03e+05 \\
Total Precipitation Sum & 2.67e-03 & 5.78e-03 & 0.00e+00 & 8.10e-06 & 2.60e-03 & 1.16e-01 \\
Max Temp 2M Max & 2.97e+02 & 6.23e+00 & 2.69e+02 & 2.92e+02 & 3.01e+02 & 3.13e+02 \\
Min Temp 2M Min & 2.87e+02 & 5.93e+00 & 2.60e+02 & 2.83e+02 & 2.92e+02 & 3.01e+02 \\
Surface Solar Radiation Sum & 2.10e+07 & 5.38e+06 & 1.68e+06 & 1.77e+07 & 2.50e+07 & 3.10e+07 \\
Max Temp Precip Max & 4.36e-04 & 7.46e-04 & 0.00e+00 & 1.91e-07 & 5.76e-04 & 1.55e-02 \\
Max Temp Precip Mean & 6.05e-05 & 1.21e-04 & 0.00e+00 & 1.04e-08 & 6.63e-05 & 2.01e-03 \\
Min Temp Precip Min & 1.44e-07 & 2.82e-06 & 0.00e+00 & 0.00e+00 & 0.00e+00 & 2.39e-04 \\
Min Temp Precip Mean & 1.27e-05 & 3.87e-05 & 0.00e+00 & 0.00e+00 & 5.93e-06 & 1.22e-03 \\
Latitude & 4.41e+01 & 1.98e+00 & 3.83e+01 & 4.33e+01 & 4.56e+01 & 4.67e+01 \\
Longitude & 1.10e+01 & 2.57e+00 & 6.98e+00 & 8.93e+00 & 1.27e+01 & 1.69e+01 \\
\bottomrule
\end{tabular}
\caption{Descriptive statistics across each hive ID and each daily observation for a total of 37202 observations.}
\label{Tab:descriptive_stats}
\end{table}
\FloatBarrier

\FloatBarrier

\begin{table}[h]
\centering
\begin{tabular}{lr}
\toprule
{} &  Percentage \\
\midrule
Diff W                      &       90.91 \\
Dewpoint Temp Mean          &       51.67 \\
Dewpoint Temp Min           &       79.43 \\
Dewpoint Temp Max           &       60.29 \\
Temperature 2M Mean         &        3.35 \\
Temperature 2M Min          &       26.32 \\
Temperature 2M Max          &       11.96 \\
Surface Pressure Mean       &       99.52 \\
Total Precipitation Sum     &       87.56 \\
Max Temp 2M Max             &       20.10 \\
Min Temp 2M Min             &       21.05 \\
Surface Solar Radiation Sum &       48.80 \\
Max Temp Precip Max         &       99.04 \\
Max Temp Precip Mean        &       91.87 \\
Min Temp Precip Min         &       81.34 \\
Min Temp Precip Mean        &       83.25 \\
\bottomrule
\end{tabular}
\caption{Percentage of stationary features among all the hive IDs.}
\label{Tab:stationary}
\end{table}
\FloatBarrier

\FloatBarrier

\section{Models for Predicting Hive Weight Variations}\label{sec:models}

We now turn to the predictive modeling of hive weight dynamics using the processed and feature-engineered dataset introduced in the previous section. We denote the resulting feature matrix by $X \in \mathbb{R}^{n \times p}$, where $n$ is the number of hive-day samples and $p$ is the number of predictors. Each row of $X$ corresponds to a unique hive at a given day, while each column represents a lagged feature. We include $\ell$ lagged time steps for each variable, so $X_{t-1:t-\ell}$ denotes the predictor vector that concatenates all features from day $t-1$ back to $t-\ell$. For simplicity, we omit the hive identifier from the notation, as we treat each hive as an independent observational unit. We do not construct features that model inter-hive interactions, and we assume a balanced panel structure where each hive contributes the same number of observations, as detailed in Sec.~\ref{sec:dataset}. Our target variable is the next-day variation in hive weight, denoted by $\Delta w_{t+1}$, which we aim to predict based on information available up to time $t-1$ for each hive. The learning task is thus formulated as a supervised regression problem, where the goal is to approximate the mapping
\[
f: X_{t-1:t-\ell} \rightarrow \Delta w_{t+1}.
\]

In the following, we describe the modeling strategies adopted for this task, including linear regression and its regularized variants, a set of tree-based machine ensemble models, and a feedforward neural network. Each method is evaluated based on its ability to predict future hive weight variations, capturing potential nonlinear interactions and temporal dependencies within the data.

We tune the hyperparameters of each model and report the procedures and selected configurations in Appendix~\ref{App:hyper}.

\subsection{Linear Models and Regularization}

For our predictive task, we employ a set of linear models that serve both as competitive baselines and as more interpretable alternatives to machine learning models. Specifically, we consider Ordinary Least Squares (OLS), Ridge regression \citep{hoerl1970ridge}, LASSO \citep{tibshirani1996regression}, and Elastic Net \citep{zou2005regularization}. All models in this family aim to estimate a vector of coefficients $\beta \in \mathbb{R}^p$ such that the predicted values $\hat{\Delta w}_{t+1} = X_{t-1:t-\ell} \beta$ approximate $\Delta w_{t+1}$.

This is achieved by minimizing the following penalized loss function:
\begin{equation}
\hat{\beta} = \arg\min_{\beta} \left\{ \| \Delta w_{t+1} - X_{t-1:t-\ell} \beta \|_2^2 + \lambda_1 \|\beta\|_1 + \lambda_2 \|\beta\|_2^2 \right\}
\end{equation}

This general formulation allows us to recover specific models based on the values of the regularization parameters:
\begin{itemize}
    \item \textit{Ordinary Least Squares (OLS):} $\lambda_1 = \lambda_2 = 0$
    \item \textit{Ridge Regression:} $\lambda_1 = 0$, $\lambda_2 > 0$
    \item \textit{LASSO Regression:} $\lambda_1 > 0$, $\lambda_2 = 0$
    \item \textit{Elastic Net:} $\lambda_1 > 0$, $\lambda_2 > 0$
\end{itemize}

While OLS minimizes prediction error without any constraint on the size or number of the coefficients, regularized models introduce penalties to improve generalization, especially in high-dimensional settings or when multicollinearity is present. Ridge regression shrinks all coefficients towards zero without forcing sparsity, whereas LASSO performs variable selection by shrinking some coefficients exactly to zero. Elastic Net combines both penalties, benefiting from the stability of Ridge and the sparsity of LASSO.

\subsection{Tree-Based Ensemble Methods}

In addition to linear models, we consider tree-based ensemble methods known for their ability to capture nonlinear relationships and complex interactions among features. In particular, we employ Random Forests (RF) \citep{breiman2001}, Extreme Gradient Boosting (XGBoost) \citep{chen2016xgboost}, and Light Gradient Boosting Machine (LightGBM) \citep{ke2017lightgbm}, each of which leverages regression trees as base learners.

Regression trees are non-parametric models that recursively partition the input space into axis-aligned rectangular regions and fit a constant value in each region. Given a sample $\left(X_{t-1:t-\ell}^{(i)}, \Delta w_{t+1}^{(i)}\right)$ for $i=1,\ldots,n$, a regression tree estimates a function $f(x)$ of the form
\begin{equation}
f(x) = \sum_{m=1}^{M} c_m \cdot \mathbb{I}(x \in R_m),
\end{equation}
where $R_m$ are disjoint regions of the input space and $c_m$ are the predicted values within each region, typically set to the average target value in that region.

\textit{Random Forest.} RFs build an ensemble of $M$ regression trees $\{f_1(x), \ldots, f_M(x)\}$ trained on bootstrap samples of the training data. At each split within a tree, a random subset of features is considered to introduce further variation. The final prediction is the average of all individual trees:
\begin{equation}
\hat{\Delta w}_{t+1}^{\text{RF}} = \frac{1}{M} \sum_{m=1}^{M} f_m(X_{t-1:t-\ell}).
\end{equation}
This bagging strategy reduces variance and prevents overfitting, especially in high-dimensional settings.

\textit{Extreme Gradient Boosting.} XGBoost employs a boosting strategy where we train regression trees sequentially to correct the residuals of the previous ensemble. The model is initialized with a constant prediction and updated iteratively:
\begin{equation}
\hat{\Delta w}_{t+1}^{\text{XGB}} = \sum_{m=1}^{B} f_m(X_{t-1:t-\ell}), \quad f_m \in \mathcal{F},
\end{equation}
where $\mathcal{F}$ is the space of regression trees. At each iteration, $f_m$ minimizes a regularized loss function via gradient descent. XGBoost incorporates both $L_1$ and $L_2$ regularization to avoid overfitting and is optimized for efficiency through sparsity-aware splits and parallel computation.

\textit{Light Gradient Boosting Machine.} LightGBM is another gradient boosting implementation that improves upon XGBoost in terms of computational efficiency and scalability. Unlike traditional level-wise tree growth strategies used by XGBoost, LightGBM adopts a leaf-wise strategy with depth constraints, allowing it to grow trees that focus more effectively on reducing loss:
\begin{equation}
\hat{\Delta w}_{t+1}^{\text{LGB}} = \sum_{m=1}^{B} g_m(X_{t-1:t-\ell}), \quad g_m \in \mathcal{G},
\end{equation}
where $\mathcal{G}$ denotes the class of leaf-wise grown trees. We use $f_m$ to denote trees in the XGBoost ensemble and $g_m$ for trees in the LightGBM model, as each method follows a different splitting strategy.

\medskip

We train these ensemble models on the same feature matrix $X_{t-1:t-\ell}$ used in the linear models and evaluated on their ability to predict $\Delta w_{t+1}$. Their flexibility and built-in regularization mechanisms make them well-suited to capture the complex, nonlinear patterns often present in high-dimensional and noisy time series data such as those observed in hive weight dynamics.

\subsection{Feedforward Neural Network}

We also employ a feedforward neural network (FNN) and train the network on the same lagged feature matrix $X_{t-1:t-\ell}$ used in previous models. The goal remains the same---to predict the next-day hive weight variation $\Delta w_{t+1}$.

A standard FNN defines a mapping of the form:
\[
\hat{\Delta w}_{t+1} = f(X_{t-1:t-\ell}; \theta),
\]
where $\theta$ denotes the weights and biases of the network, which are learned by minimizing a loss function over the training data, typically the mean squared error.

The input layer receives the lagged feature matrix $X_{t-1:t-\ell}$, followed by one or more hidden layers that transform the data through learned nonlinearities, and a final output layer returning the prediction. We train the network using backpropagation with gradient descent-based optimization.

We intentionally opt for a simple feedforward architecture and avoid more complex sequential models such as Recurrent Neural Networks or Long Short-Term Memorynetworks. As discussed in Sec.~\ref{sec:dataset}, most features exhibit a limited autocorrelation structure, often vanishing after a few lags. This motivates our decision to engineer lagged predictors explicitly rather than passing sequences to recurrent models.

FNNs are among the most commonly used neural architectures in supervised learning and form the backbone of more complex deep learning models \citep[see, for example,][]{goodfellow2016deep}. We emphasize their suitability for this task due to their flexibility, compatibility with tabular data, and robustness when combined with regularization techniques such as dropout and early stopping.

\section{Empirical Findings}\label{sec:empirical}

This section presents the predictive performance of the models described in Sec.~\ref{sec:models}\footnote{The code used in this study will be publicly available on GitHub upon acceptance. The hive-level data are proprietary and cannot be shared. However, a synthetic sample dataset can be provided for the purpose of reproducing the analysis and running the code.}
. All models are trained and evaluated under a consistent data-splitting procedure to ensure comparability. Specifically, we partition the dataset chronologically, allocating the most recent 30\% of observations to the test set. The remaining 70\% is also split chronologically, with the earliest 90\% used for training and the latest 10\% reserved for validation during hyperparameter tuning. This procedure yields the following partition sizes: 23,199 observations for training, 2,717 for validation, and 11,286 for testing. OLS is the only model not requiring hyperparameter tuning and is trained directly on the training set.

To assess the contribution of temporal dynamics to model performance, we conduct a series of experiments in which each model is trained under different configurations, progressively increasing the number of lagged features included. More precisely, for each forecasting method, we define four feature sets, including up to one, two, three, and four lags of the variables described in Sec.~\ref{sec:dataset}. This results in feature matrices comprising 18, 34, 50, and 66 features, respectively. The expansion occurs by adding one additional lag per variable at each step, while keeping the two spatial features, latitude and longitude, unchanged.

To assess and compare predictive accuracy across models and input configurations, we report four different loss metrics for a sample of size $N$:
\begin{itemize}
    \item \textbf{Root Mean Squared Error (RMSE)}: $\sqrt{\frac{1}{N} \sum_{i=1}^N (\Delta w_i - \hat{\Delta w}_i)^2}$
    
    \item \textbf{Normalized Root Mean Squared Error (NRMSE)}: RMSE normalized by the mean of the observed values, to remove scale dependency
    
    \item \textbf{Mean Absolute Error (MAE)}: $\frac{1}{N} \sum_{i=1}^N |\Delta w_i - \hat{\Delta w}_i|$
    
    \item \textbf{Log-Cosh Loss}: $\sum_{i=1}^N \log(\cosh(\hat{\Delta w}_i - \Delta w_i))$
\end{itemize}

Each metric captures a distinct aspect of model performance. RMSE penalizes large deviations more heavily and is suited for identifying extreme prediction errors. NRMSE facilitates cross-scale comparisons, especially valuable when seasonal shifts alter production levels. MAE offers robustness to outliers and ease of interpretation. Log-Cosh balances RMSE’s sensitivity with MAE’s robustness, making it useful when occasional large errors may occur.

Tab.~\ref{Tab:model_performance_test} reports the performance of each model across all feature sets and evaluation metrics on the test set. We include the corresponding results on the validation set in Appendix~\ref{App:val}. We structure Tab.~\ref{Tab:model_performance_test} to illustrate how predictive accuracy evolves as the information set expands. As described earlier, feature sets grow with the inclusion of lagged variables, increasing the number of predictors from 18 to 66.

Examining the results, we observe that expanding the feature set by including additional lags of the features generally improves predictive accuracy. This trend is consistent across all models and evaluation metrics, as models trained with four lags outperform their counterparts trained with fewer lags.

On the test set, Elastic Net with four lags achieves the best performance across three of the four evaluation metrics: RMSE, NRMSE, and Log-Cosh. When Elastic Net does not deliver the top score, the second-best model is typically either a variant of Elastic Net with fewer lags or the RF model. We note that RF outperforms Elastic Net under MAE. This may reflect the different error sensitivities of the metrics: Elastic Net minimizes a regularized squared error loss, making it more sensitive to large deviations, while RF's ensemble structure may yield more stable median-like predictions, which can align better with the MAE criterion in the presence of asymmetric or heavy-tailed errors.

When considering the feature set with four lags, the tree-based ensemble models, RF, XGB, and LightGBM, deliver predictive performance that remains highly competitive with the best-performing linear alternatives. This result underscores the ability of nonlinear models to exploit richer temporal information, particularly in low-dimensional tabular settings, yet also shows that Elastic Net can outperform them in terms of overall error, while ensemble methods remain robust and well-suited to capturing complex input patterns.

In contrast, the FNN struggles to match the predictive accuracy of either Elastic Net or the ensemble tree methods across all feature sets and evaluation metrics. This outcome suggests that simple dense architectures may be less effective in capturing the structure of this specific dataset, likely because they require more data to learn meaningful patterns effectively.

Overall, comparing the performance of competing models to that of OLS reveals substantial improvements from using either highly regularized linear methods or nonlinear ensembles. These gains are consistent across all metrics, underscoring the importance of incorporating regularization or flexible modeling approaches in this forecasting task.

\begin{figure}[h]
    \centering
    \includegraphics[width=0.35\textwidth]{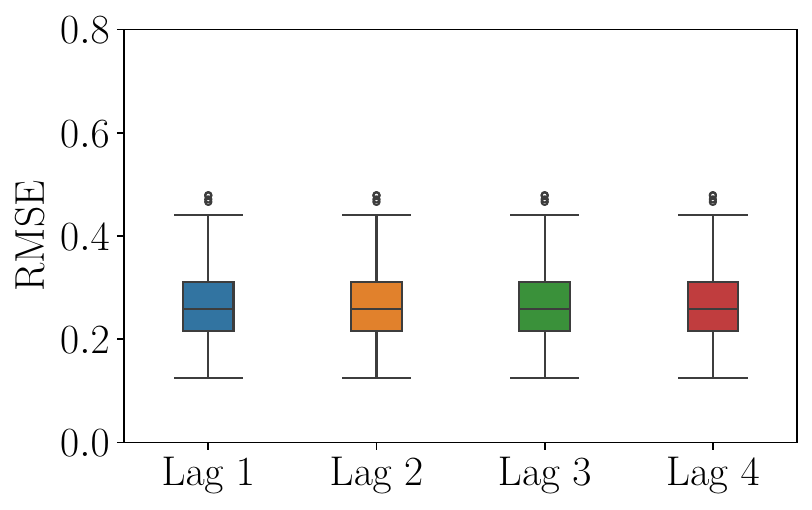}
    \hspace{0.04\textwidth}
    \includegraphics[width=0.35\textwidth]{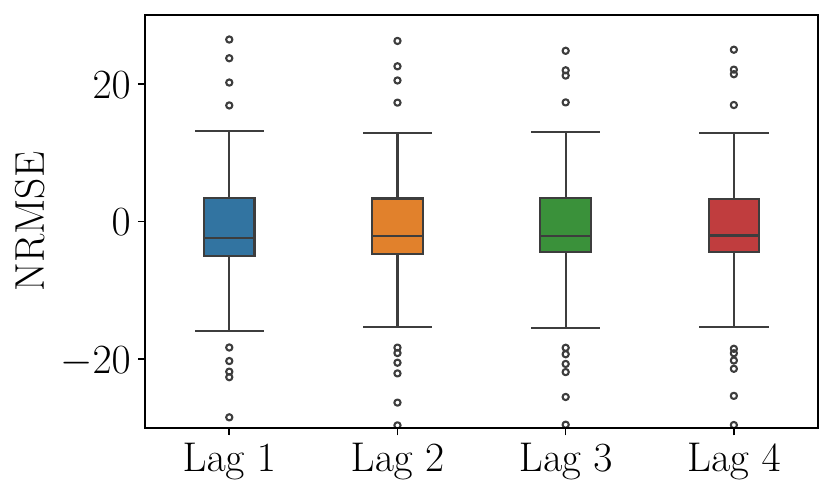}
    
    \vskip\baselineskip
    
    \includegraphics[width=0.35\textwidth]{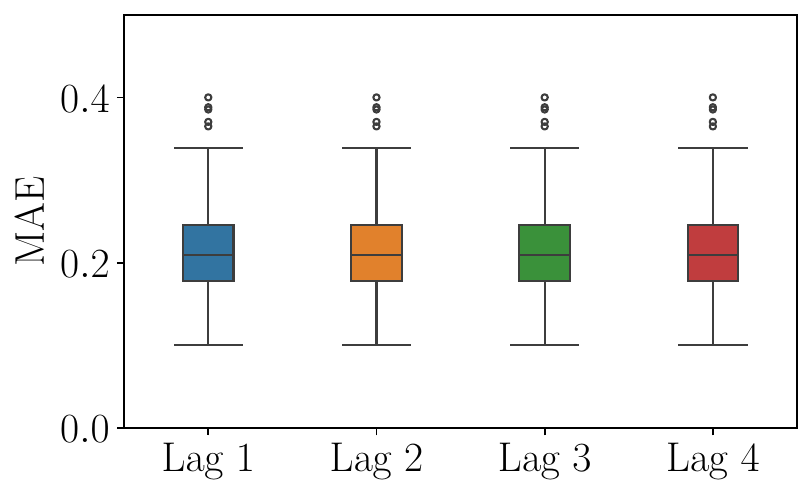}
    \hspace{0.04\textwidth}
    \includegraphics[width=0.35\textwidth]{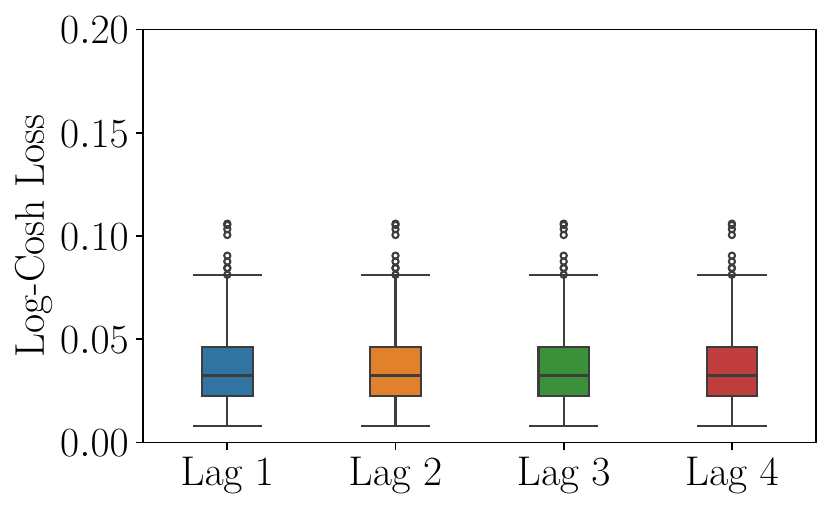}
    \caption{Prediction error distributions on the test set for Elastic Net across the four evaluation metrics.}
    \label{fig:elasticnet_test_errors}
\end{figure}

\begin{figure}[h]
    \centering
    \includegraphics[width=0.35\textwidth]{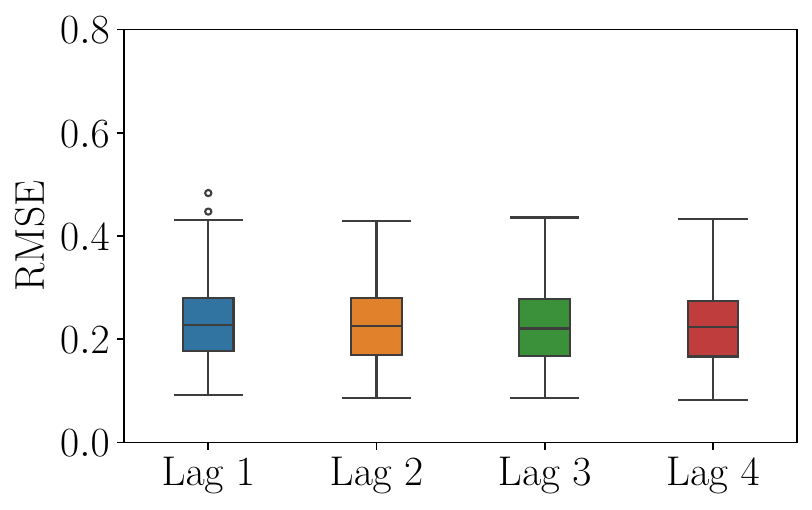}
    \hspace{0.04\textwidth}
    \includegraphics[width=0.35\textwidth]{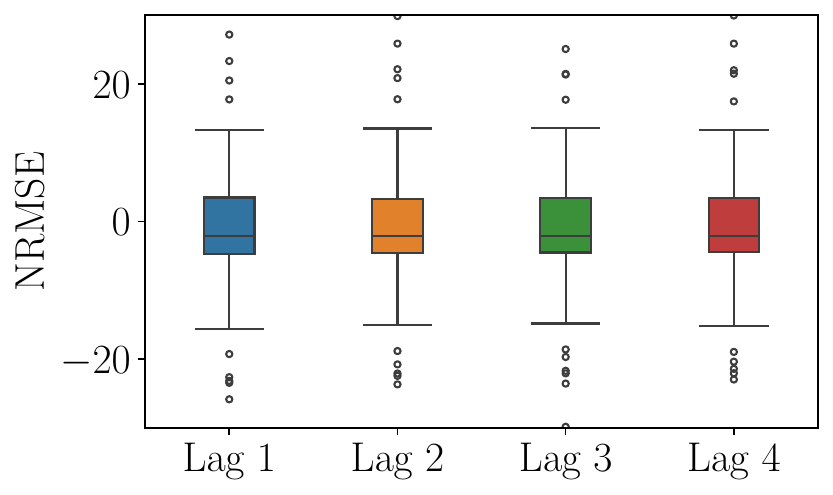}
    
    \vskip\baselineskip
    
    \includegraphics[width=0.35\textwidth]{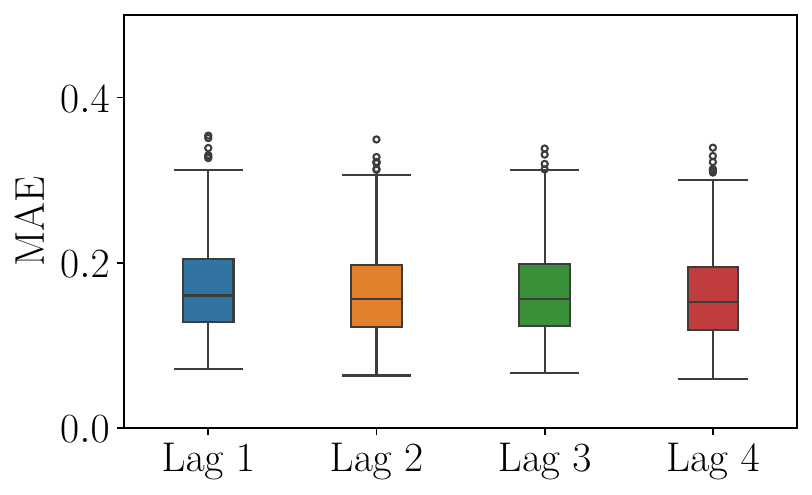}
    \hspace{0.04\textwidth}
    \includegraphics[width=0.35\textwidth]{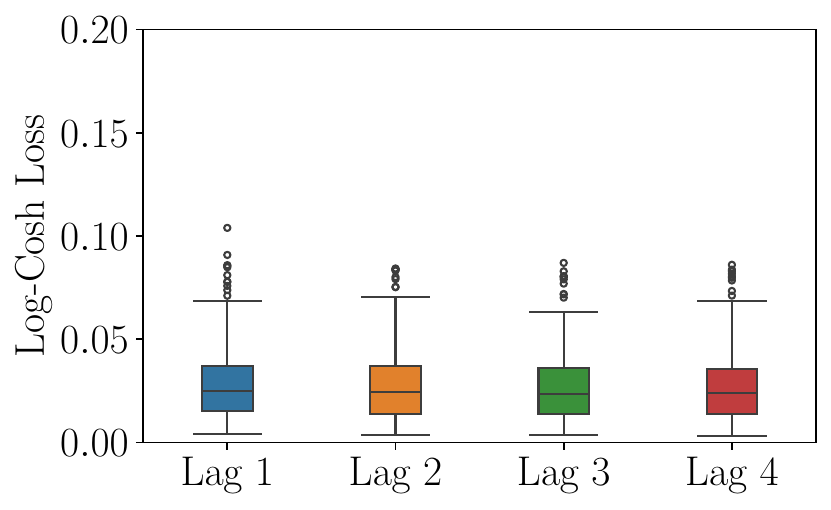}
    \caption{Prediction error distributions on the test set for RF across the four evaluation metrics.}
    \label{fig:rf_test_errors}
\end{figure}

To further illustrate the effect of expanding the feature space through additional lags, Fig.~\ref{fig:elasticnet_test_errors} and Fig.~\ref{fig:rf_test_errors} show the distribution of prediction errors for Elastic Net and RF, respectively, across different lag configurations. In both cases, we observe a clear reduction in the dispersion and number of outliers as more temporal information is incorporated. Additional results for other models are provided in Appendix~\ref{App:error_distributions}.

\begin{table}[htbp]
\centering
\begin{tabular}{c c | cccc}
\toprule
Model & Lag & RMSE & NRMSE & MAE & Log-Cosh \\
\midrule
\multirow{4}{*}{OLS} & 1 & 0.253406 & 4.420324 & 0.174299 & 0.030278 \\
& 2 & 0.248937 & 4.611122 & 0.170637 & 0.029260 \\
& 3 & 0.249703 & 4.738134 & 0.172267 & 0.029438 \\
& 4 & 0.250934 & 4.803010 & 0.173628 & 0.029724 \\
\midrule
\multirow{4}{*}{LASSO} & 1 & 0.253168 & 4.411062 & 0.174052 & 0.030223 \\
& 2 & 0.248374 & 4.588200 & 0.170080 & 0.029131 \\
& 3 & 0.248879 & 4.705112 & 0.171372 & 0.029245 \\
& 4 & 0.249599 & 4.750250 & 0.172221 & 0.029413 \\
\midrule
\multirow{4}{*}{Ridge} & 1 & 0.253374 & 4.418625 & 0.174267 & 0.030270 \\
& 2 & 0.248887 & 4.608976 & 0.170585 & 0.029248 \\
& 3 & 0.249627 & 4.735033 & 0.172179 & 0.029420 \\
& 4 & 0.250813 & 4.798764 & 0.173497 & 0.029696 \\
\midrule
\multirow{4}{*}{Elastic Net} & 1 & 0.250060 & \textbf{4.098937} & 0.182627 & 0.029780 \\
& 2 & 0.242750 & 4.157926 & 0.172201 & 0.028030 \\
& 3 & 0.239220 & 4.146437 & 0.166685 & 0.027195 \\
& 4 & \textbf{0.237344} & \textit{4.139143} & \textit{0.163867} & \textbf{0.026758} \\
\midrule
\multirow{4}{*}{RF} & 1 & 0.247960 & 4.223913 & 0.170646 & 0.029091 \\
& 2 & 0.241428 & 4.348248 & 0.165003 & 0.027630 \\
& 3 & 0.240391 & 4.437630 & 0.165636 & 0.027416 \\
& 4 & \textit{0.238295} & 4.229303 & \textbf{0.161771} & \textit{0.026926} \\
\midrule
\multirow{4}{*}{XGBoost} & 1 & 0.250489 & 4.254226 & 0.179589 & 0.029851 \\
& 2 & 0.241799 & 4.409599 & 0.165927 & 0.027707 \\
& 3 & 0.239937 & 4.424339 & 0.164782 & 0.027306 \\
& 4 & 0.238975 & 4.297915 & 0.164232 & 0.027104 \\
\midrule
\multirow{4}{*}{LightGBM} & 1 & 0.248754 & 4.267907 & 0.171617 & 0.029261 \\
& 2 & 0.241845 & 4.390921 & 0.165906 & 0.027717 \\
& 3 & 0.240080 & 4.455639 & 0.164700 & 0.027334 \\
& 4 & 0.239354 & 4.290254 & 0.163884 & 0.027175 \\
\midrule
\multirow{4}{*}{FNN} & 1 & 0.247933 & 4.264736 & 0.169998 & 0.029079 \\
& 2 & 0.242204 & 4.494661 & 0.166968 & 0.027817 \\
& 3 & 0.243344 & 4.565688 & 0.169432 & 0.028110 \\
& 4 & 0.243297 & 4.741252 & 0.168119 & 0.028057 \\
\bottomrule
\end{tabular}
\caption{Model Performance Comparison on the Test Set. The table reports the predictive performance of each model across different feature sets defined by the number of included lags. For each model and feature configuration, we report RMSE, NRMSE, MAE, and Log-Cosh evaluated on the test set. The best-performing model for each metric is bolded, while the second-best is italicized.}
\label{Tab:model_performance_test}
\end{table}
\FloatBarrier

\FloatBarrier

\FloatBarrier

\begin{figure}[H]
    \centering
    \includegraphics[width=0.35\textwidth]{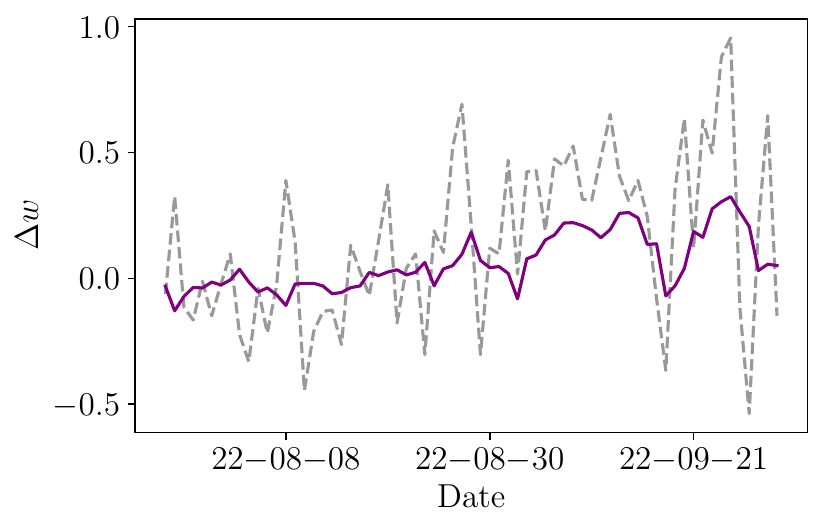}
    \hspace{0.04\textwidth}
    \includegraphics[width=0.35\textwidth]{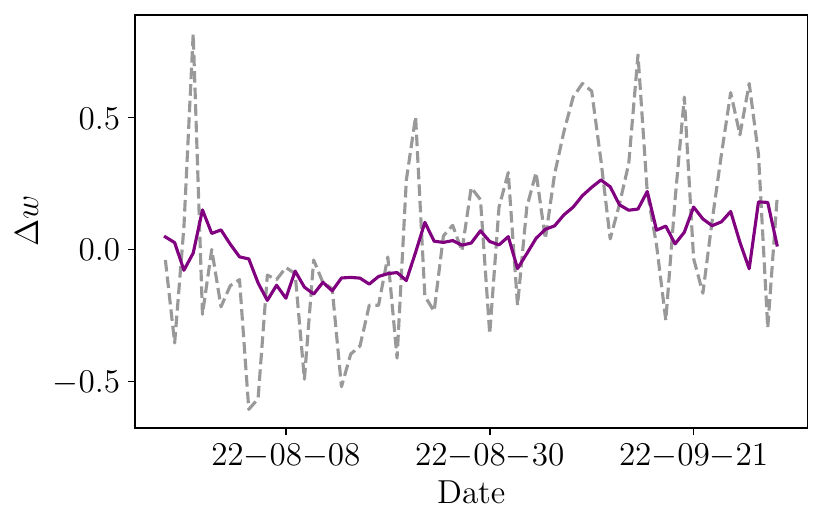}
    
    \vskip\baselineskip
    
    \includegraphics[width=0.35\textwidth]{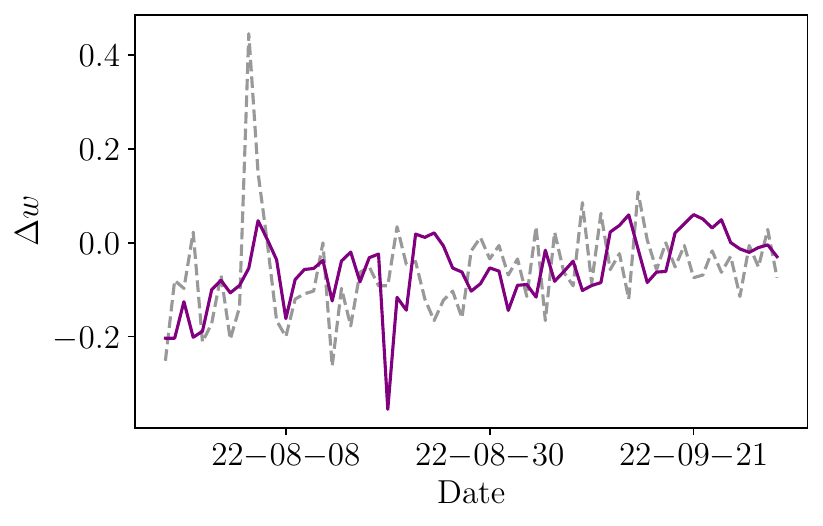}
    \hspace{0.04\textwidth}
    \includegraphics[width=0.35\textwidth]{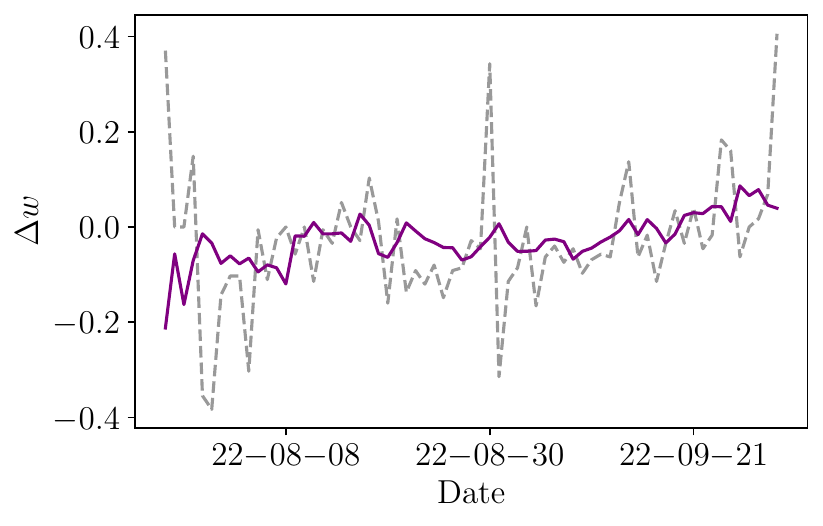}

    \caption{Predicted versus actual hive weights for selected hives using Elastic Net. Each panel shows results for a different hive in the dataset.}
    \label{Fig:elasticnet_hive_predictions}
\end{figure}

\begin{figure}[H]
    \centering
    \includegraphics[width=0.35\textwidth]{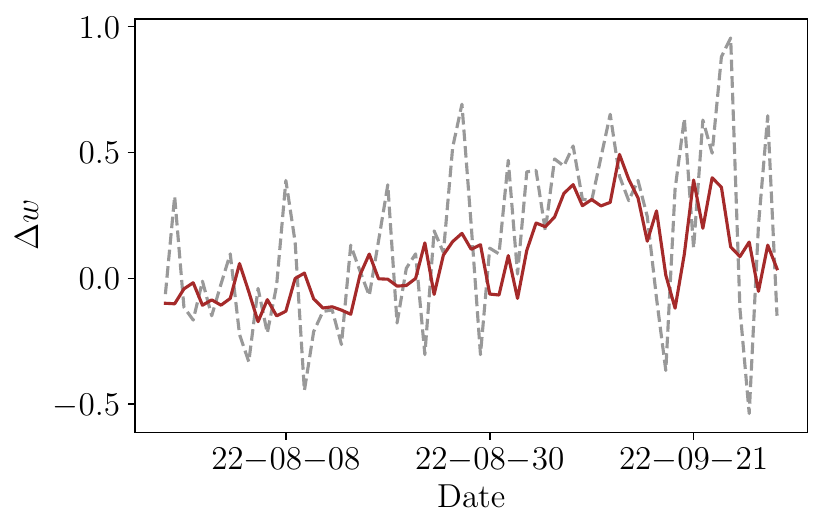}
    \hspace{0.04\textwidth}
    \includegraphics[width=0.35\textwidth]{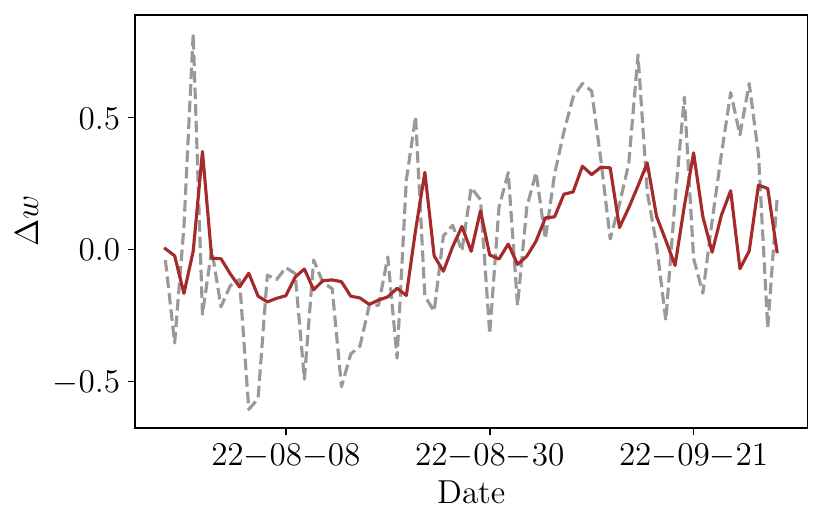}
    
    \vskip\baselineskip
    
    \includegraphics[width=0.35\textwidth]{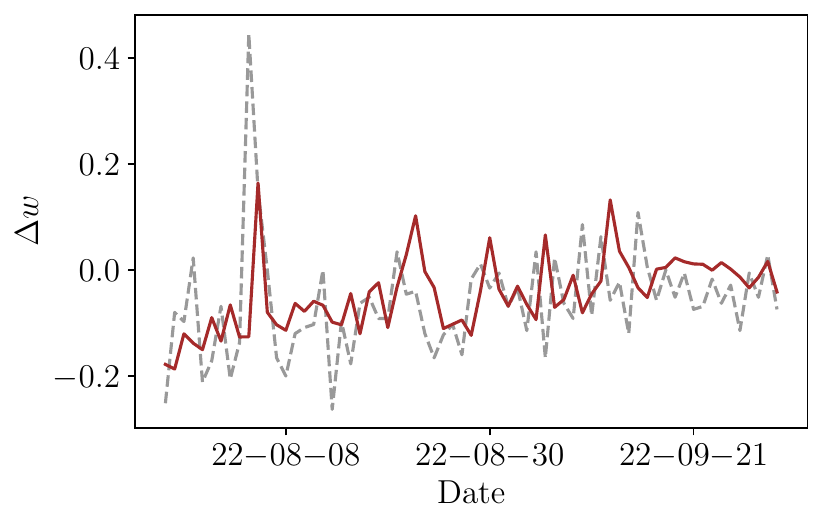}
    \hspace{0.04\textwidth}
    \includegraphics[width=0.35\textwidth]{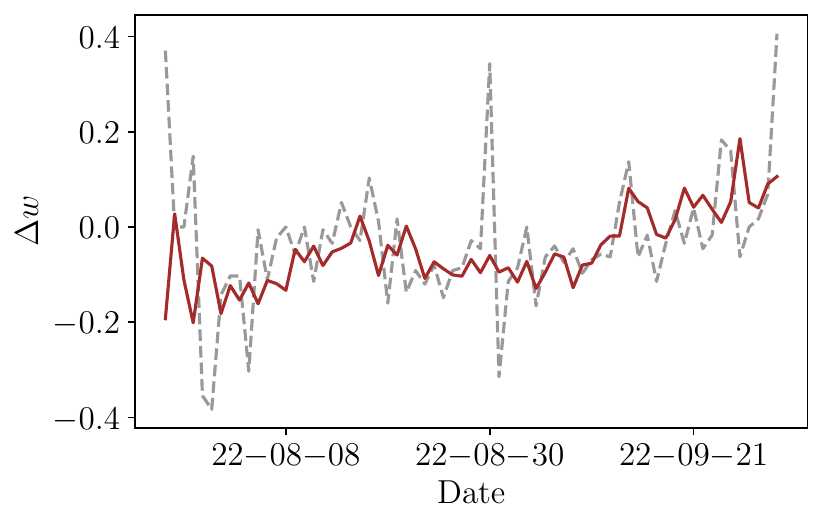}

    \caption{Predicted versus actual hive weights for selected hives using RF. Each panel shows results for a different hive in the dataset.}
    \label{Fig:rf_hive_predictions}
\end{figure}

Figs~\ref{Fig:elasticnet_hive_predictions} and~\ref{Fig:rf_hive_predictions} illustrate the model predictions for a representative hive across different lag configurations. In both cases, the inclusion of additional temporal information leads to visibly improved tracking of the true hive weights, with reduced prediction volatility and a closer alignment between predicted and actual values as more lags are incorporated. Notably, RF appears better at capturing the peaks in hive weight variations, which is also reflected in its lower MAE compared to Elastic Net.

\FloatBarrier

\FloatBarrier


\section{Feature Importance and Model Interpretability}
\label{sec:interpretability}

After analyzing the predictive performance of the models, we now turn our attention to understanding the underlying decision processes of the trained algorithms. Given the importance of model transparency in informing risk management strategies and supporting the potential design of insurance tools for beekeepers, we investigate the drivers behind model predictions through an extensive feature importance analysis.

Among the models evaluated, Elastic Net emerged as one of the best performing approaches. Being a highly regularized linear model, its interpretability is direct, as the magnitude and sign of its coefficients provide a straightforward indication of feature relevance. 

\begin{figure}[H]
    \centering
    \begin{tabular}{cc}
        \includegraphics[width=0.45\textwidth]{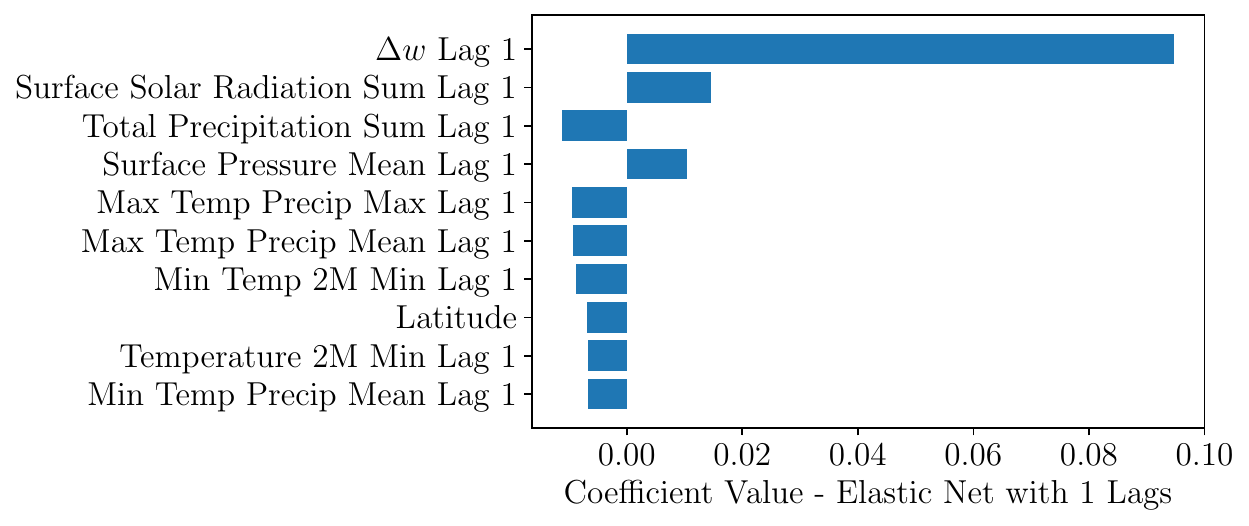} &
        \includegraphics[width=0.45\textwidth]{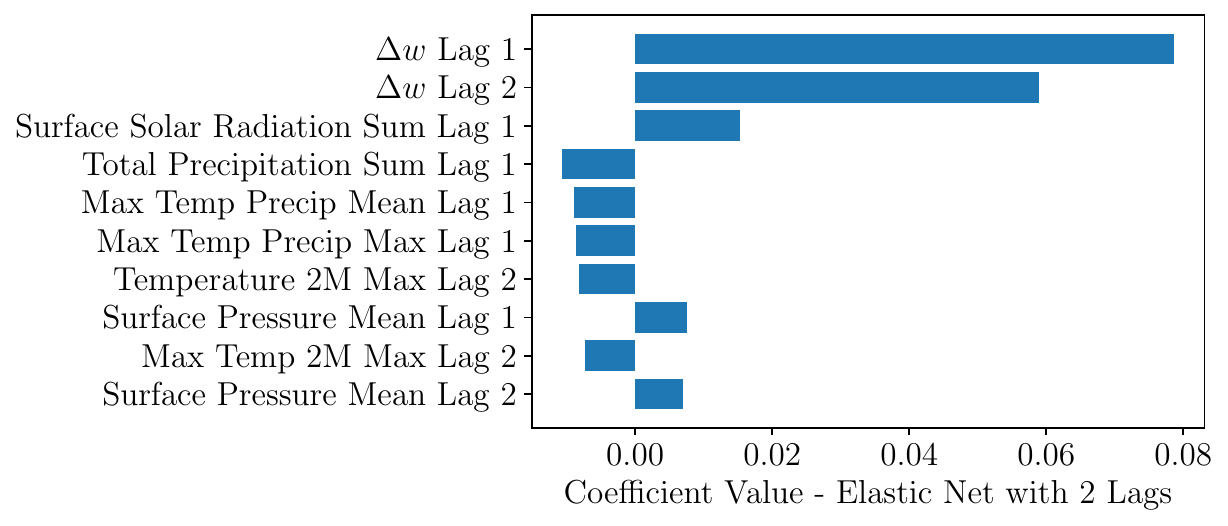} \\
        \includegraphics[width=0.45\textwidth]{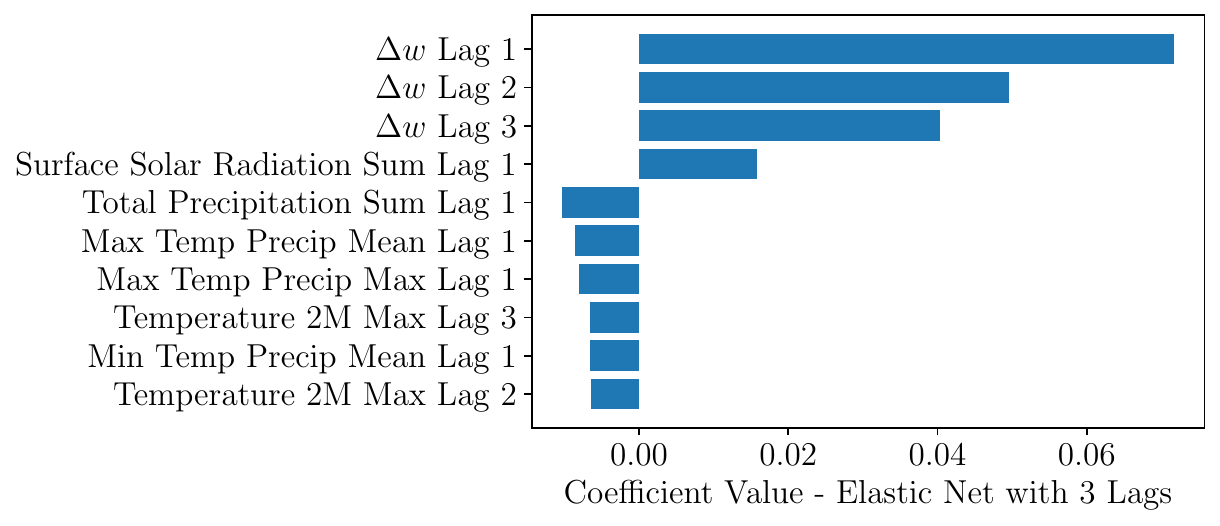} &
        \includegraphics[width=0.45\textwidth]{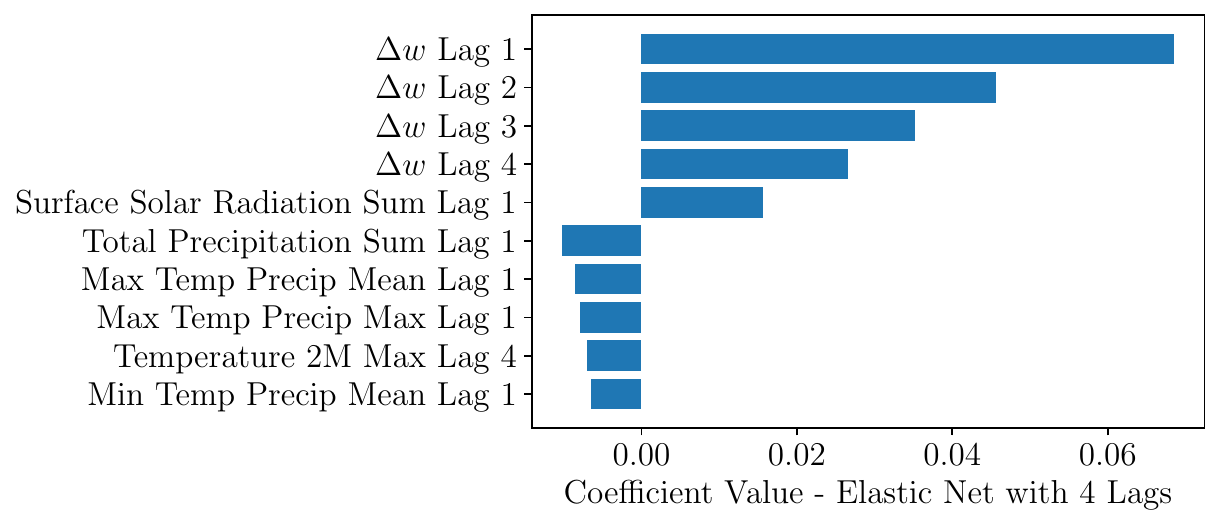}
    \end{tabular}
    \caption{Top 10 absolute coefficients estimated by Elastic Net across different lag configurations.}
    \label{fig:elasticnet_coef_grid}
\end{figure}
\FloatBarrier

Fig.~\ref{fig:elasticnet_coef_grid} reports the ten largest coefficients in absolute value estimated by Elastic Net for each lag configuration. These coefficients offer a direct interpretation of feature relevance within a linear framework. Across all configurations, the autoregressive components of hive weight variation, particularly the first and second lags, consistently appear as dominant predictors. Weather-related variables, especially surface solar radiation, temperature extremes, and precipitation indicators, also emerge with non-negligible weights, reinforcing their role in modulating short-term hive dynamics. To complement this and explore the behavior of more complex nonlinear models, we carry out a detailed feature importance analysis for the tree-based ensemble methods.

Specifically, we focus on RF and LightGBM, applying three complementary techniques to assess feature importance: impurity-based importance, permutation importance, and SHapley Additive exPlanations (SHAP). This comparison with the coefficients estimated by Elastic Net offers a comprehensive view of the most influential factors in predicting hive weight variations.

\begin{figure}[H]
    \centering
    \begin{tabular}{cc}
        \includegraphics[width=0.45\textwidth]{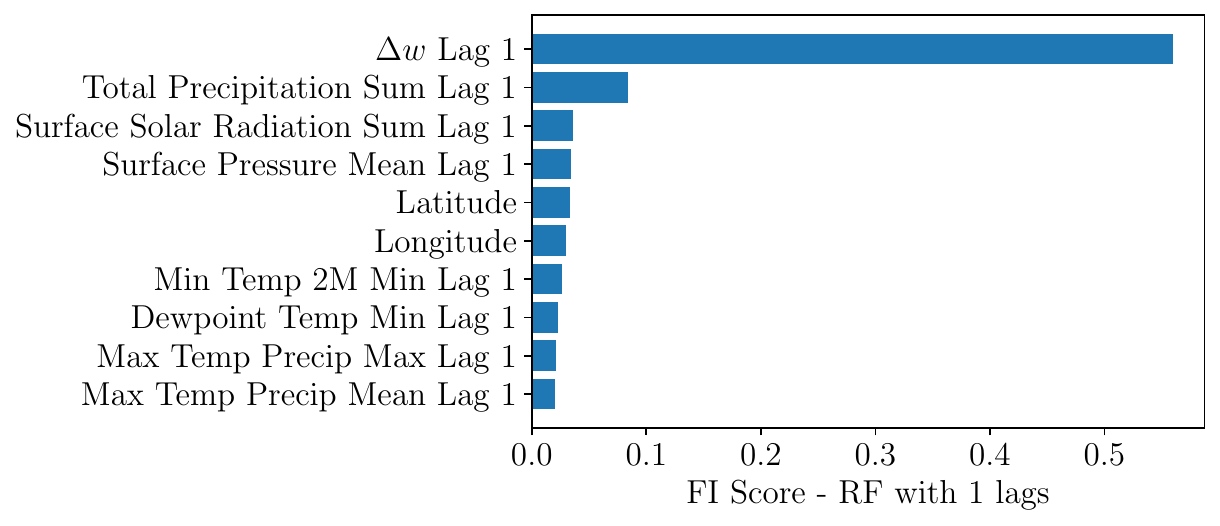} &
        \includegraphics[width=0.45\textwidth]{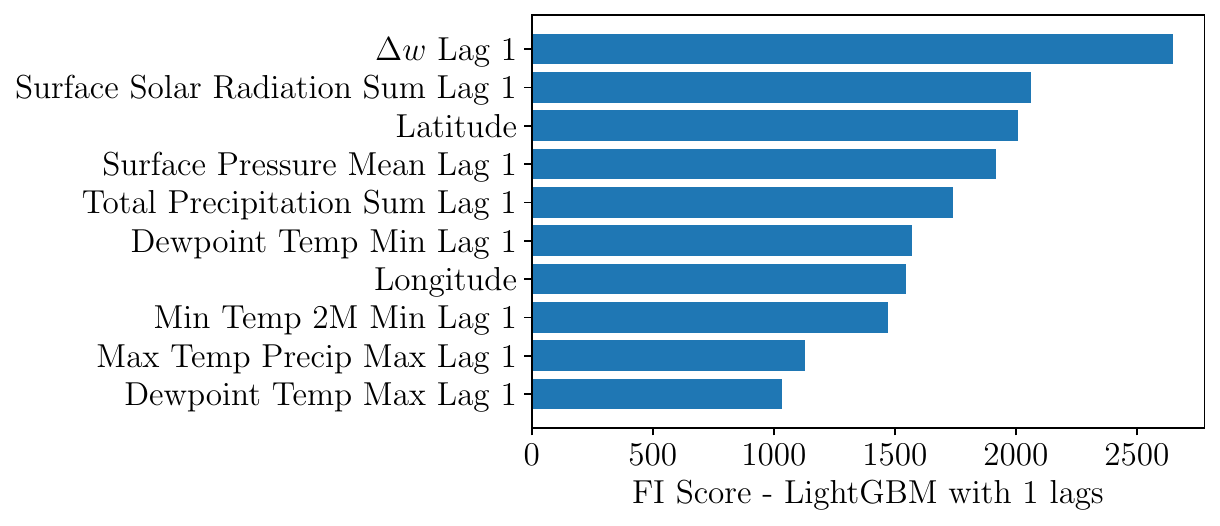} \\
        \includegraphics[width=0.45\textwidth]{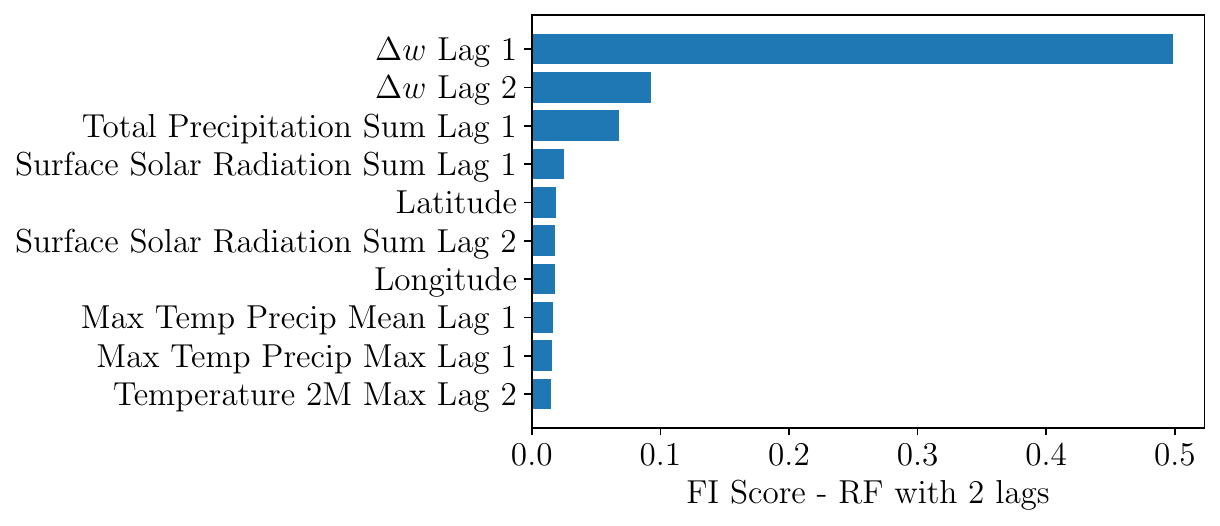} &
        \includegraphics[width=0.45\textwidth]{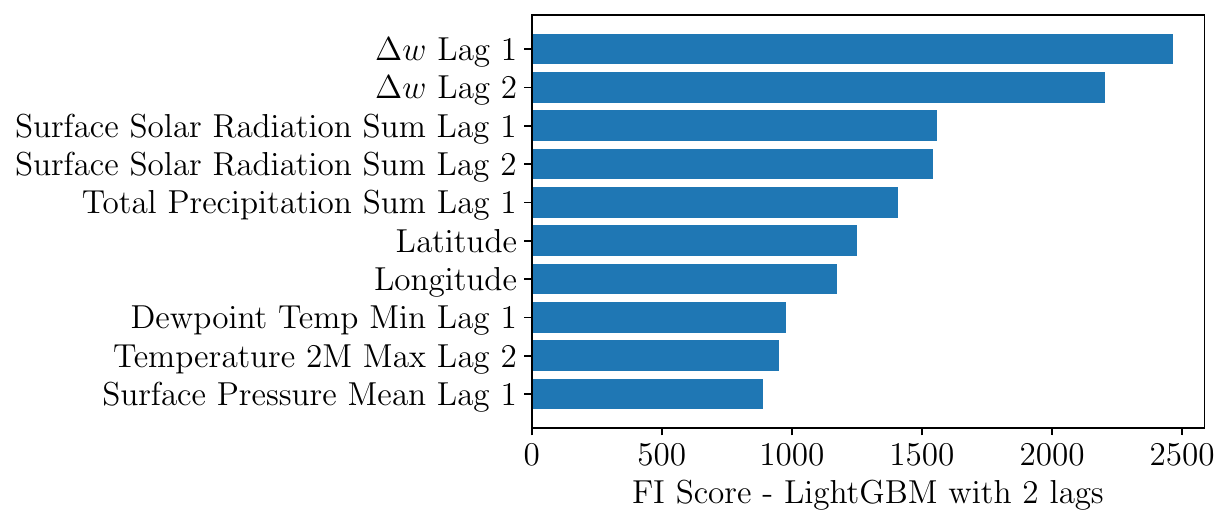} \\
        \includegraphics[width=0.45\textwidth]{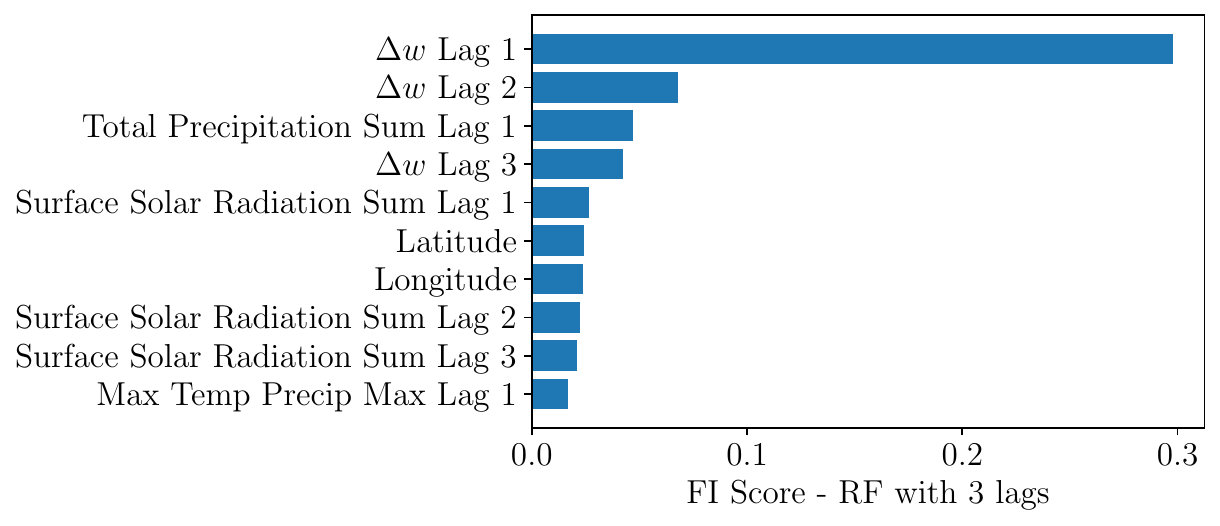} &
        \includegraphics[width=0.45\textwidth]{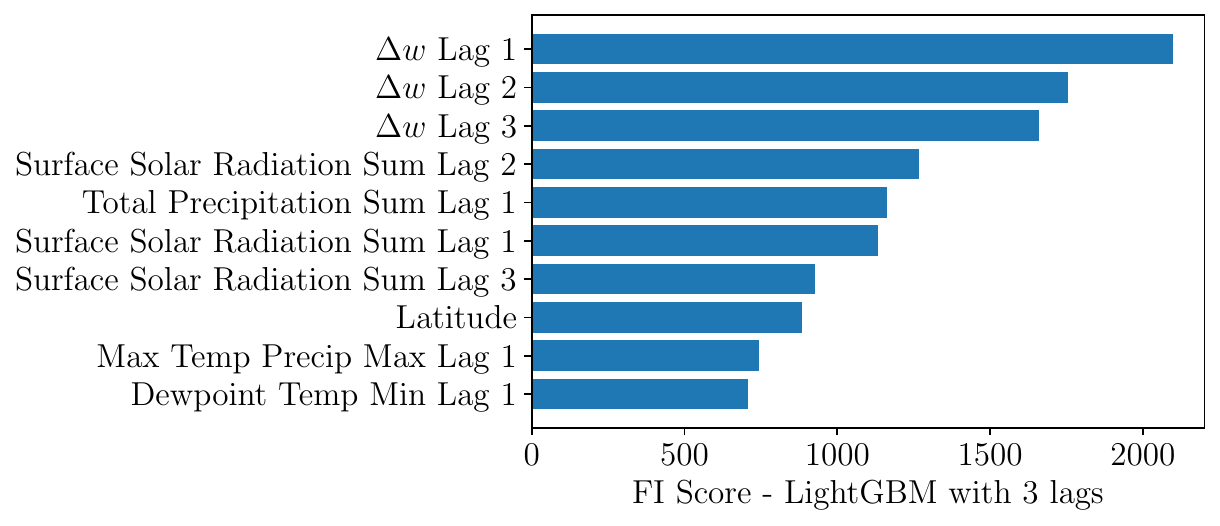} \\
        \includegraphics[width=0.45\textwidth]{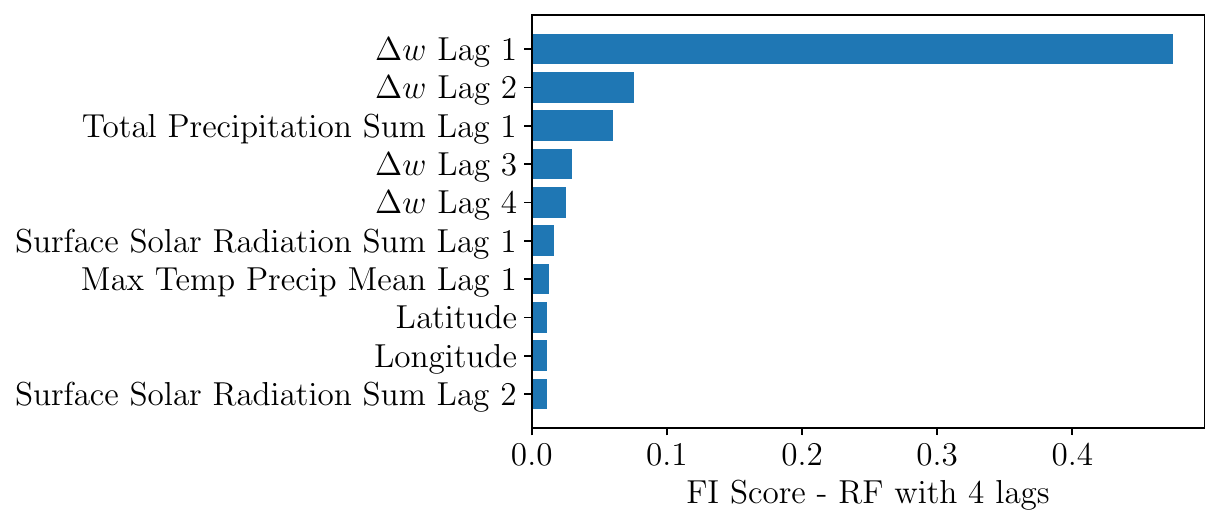} &
        \includegraphics[width=0.45\textwidth]{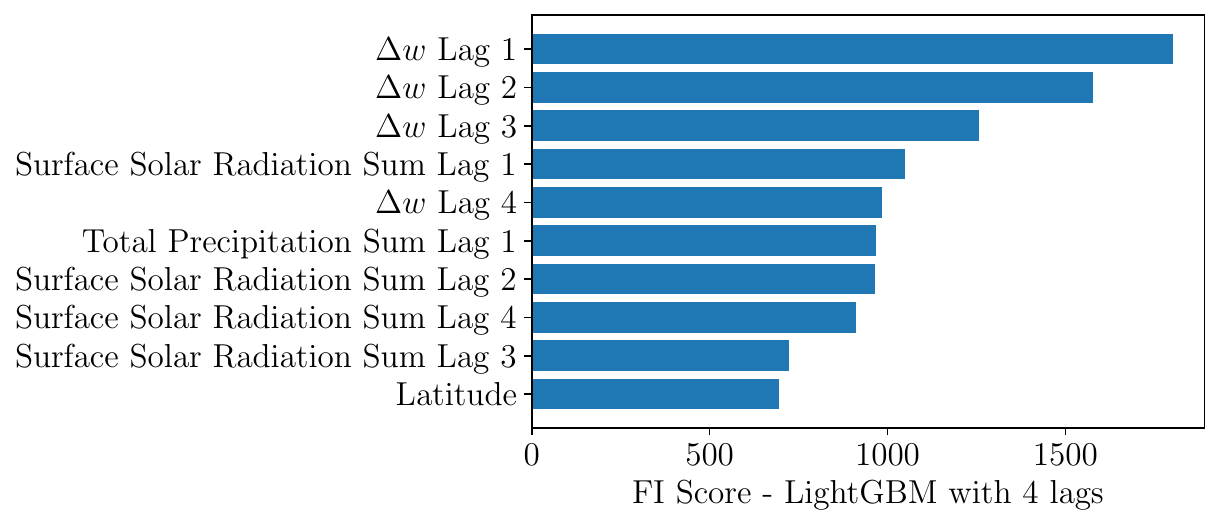}
    \end{tabular}
    \caption{Impurity-based feature importance for RF (left) and LGBM (right) across different lag configurations (top to bottom: 1 to 4 lags).}
    \label{fig:impurity_grid}
\end{figure}

Fig.~\ref{fig:impurity_grid} reports impurity-based feature importance scores for RF and LGBM models trained across different lag configurations. The importance of a feature, also called impurity-based importance, is the total reduction in the splitting criterion (e.g., variance or MSE) attributed to that feature across all trees in the ensemble. The squared relative importance of a feature is computed as the sum of the squared improvements on all internal nodes in which it was selected as the splitting variable \citep{hastie2009libro}. The higher the mean decrease in impurity over all parallel (RF) or sequential (LightGBM) trees, the more important the feature is to obtain accurate results. Lagged variations in hive weight continue to dominate feature importance rankings across models and lag structures, confirming the findings already observed with Elastic Net. Meteorological variables such as surface solar radiation, precipitation and minimum temperature maintain complementary relevance.

It is worth noting that impurity-based importance scores are not directly comparable in magnitude between RF and LightGBM. This discrepancy arises because RF computes normalized importance values, scaling them so they sum to one across all features, whereas LightGBM reports unnormalized scores, typically based on the total reduction in loss (gain) across all splits involving a given feature. As a result, while the absolute scale of the scores differs across models, their relative ordering within each model remains informative and comparable.

Across models, we also observe that impurity-based scores tend to be more evenly distributed across features in LightGBM compared to RF. This difference stems from LightGBM's leaf-wise tree growth and gradient-based split selection, which favor broader usage of variables across the tree structure. In contrast, RF's level-wise splitting and reliance on early impurity reduction often cause dominant features, such as the first lag of $\Delta w$, to concentrate most of the importance, especially when they yield large gains in the initial splits.

\FloatBarrier

\begin{figure}[H]
    \centering
    \begin{tabular}{cc}
        \includegraphics[width=0.45\textwidth]{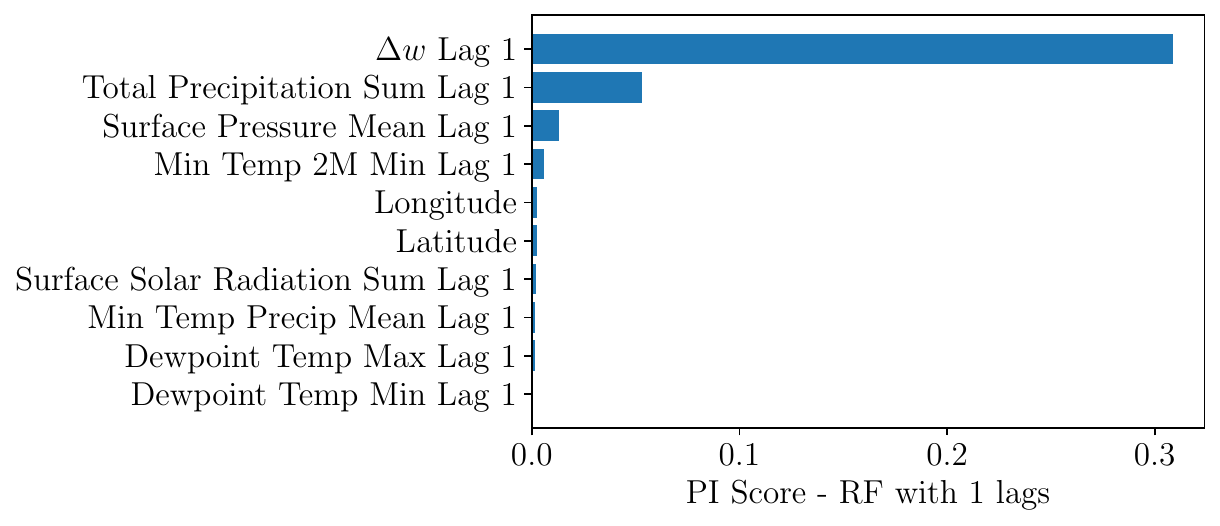} &
        \includegraphics[width=0.45\textwidth]{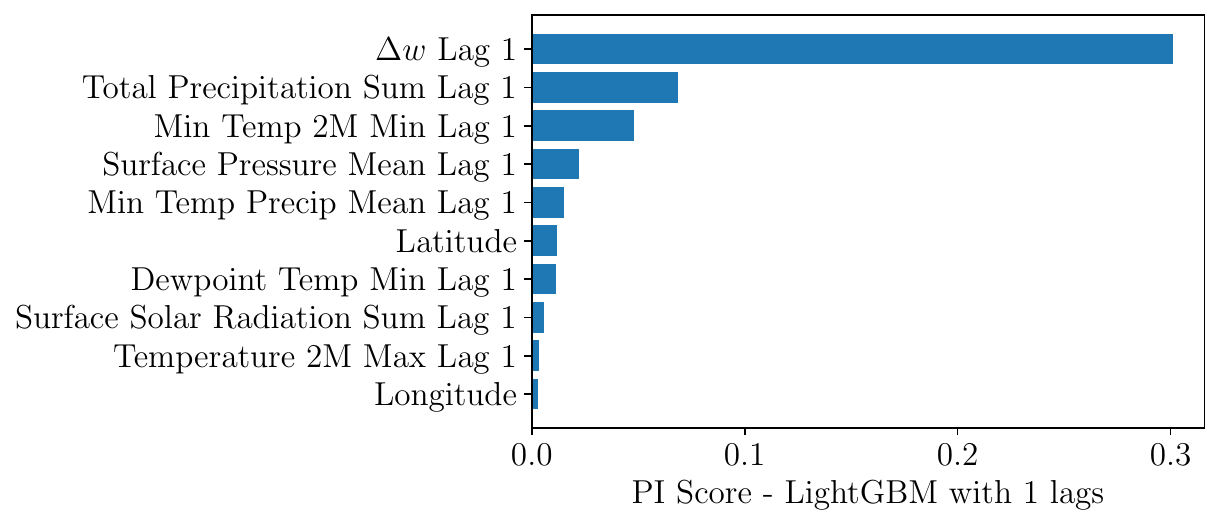} \\
        \includegraphics[width=0.45\textwidth]{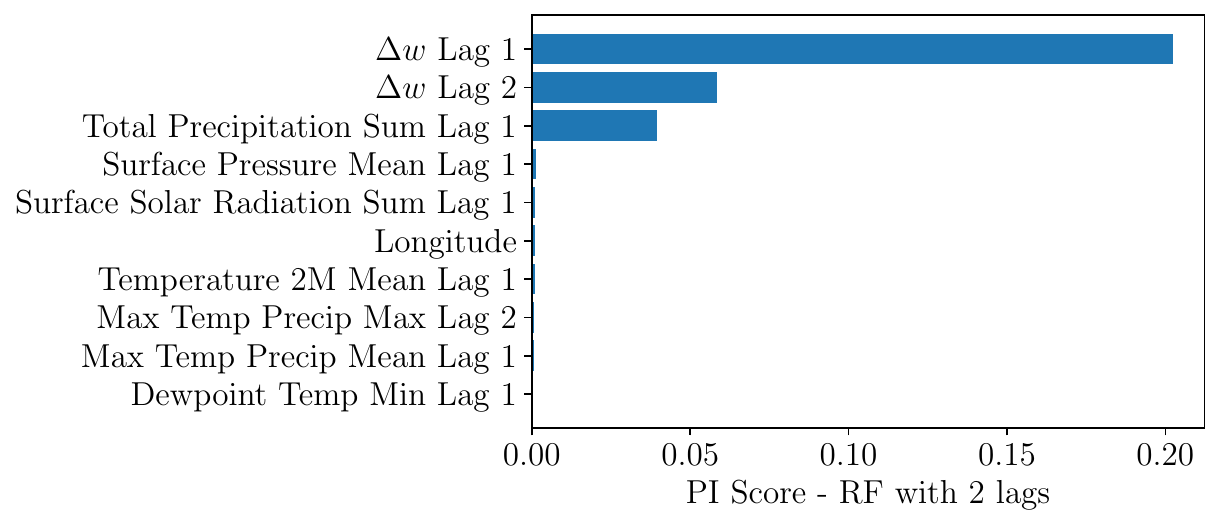} &
        \includegraphics[width=0.45\textwidth]{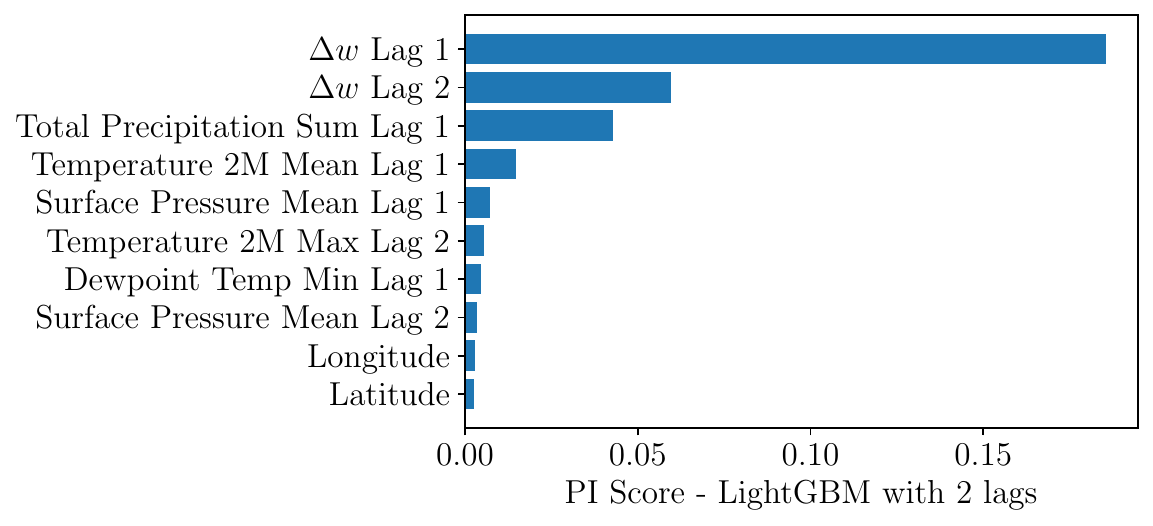} \\
        \includegraphics[width=0.45\textwidth]{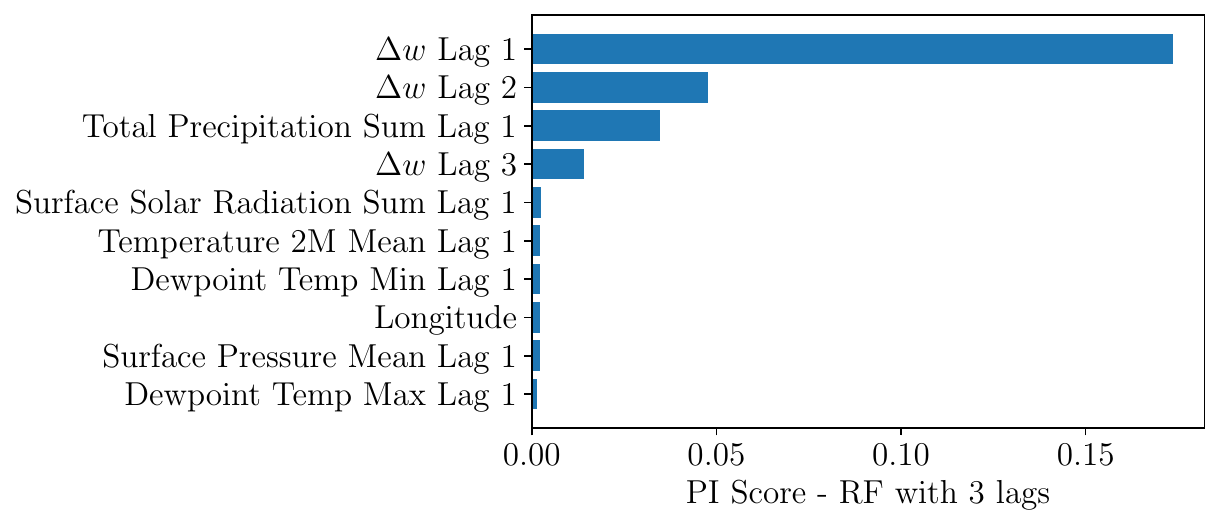} &
        \includegraphics[width=0.45\textwidth]{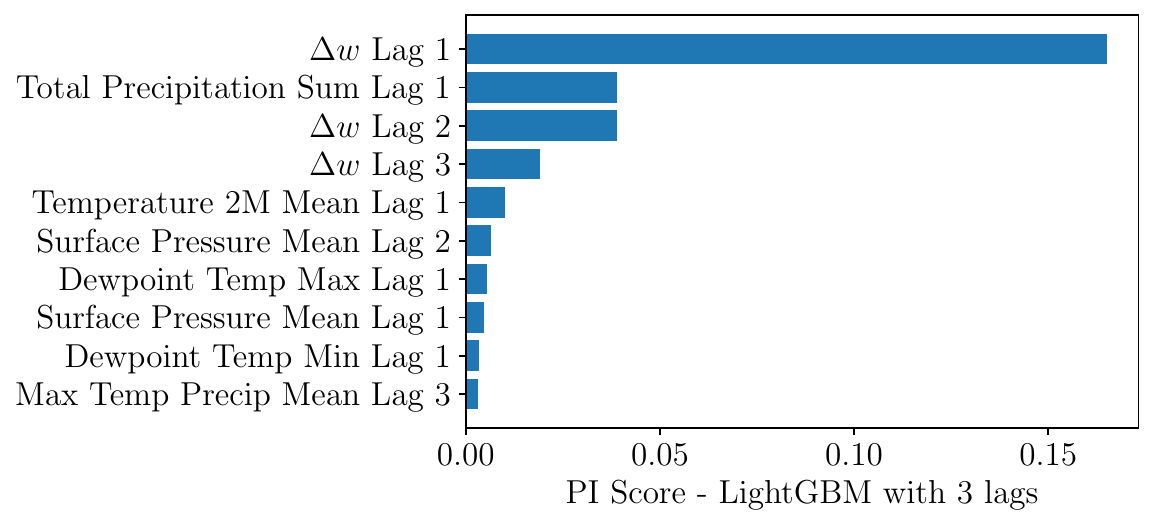} \\
        \includegraphics[width=0.45\textwidth]{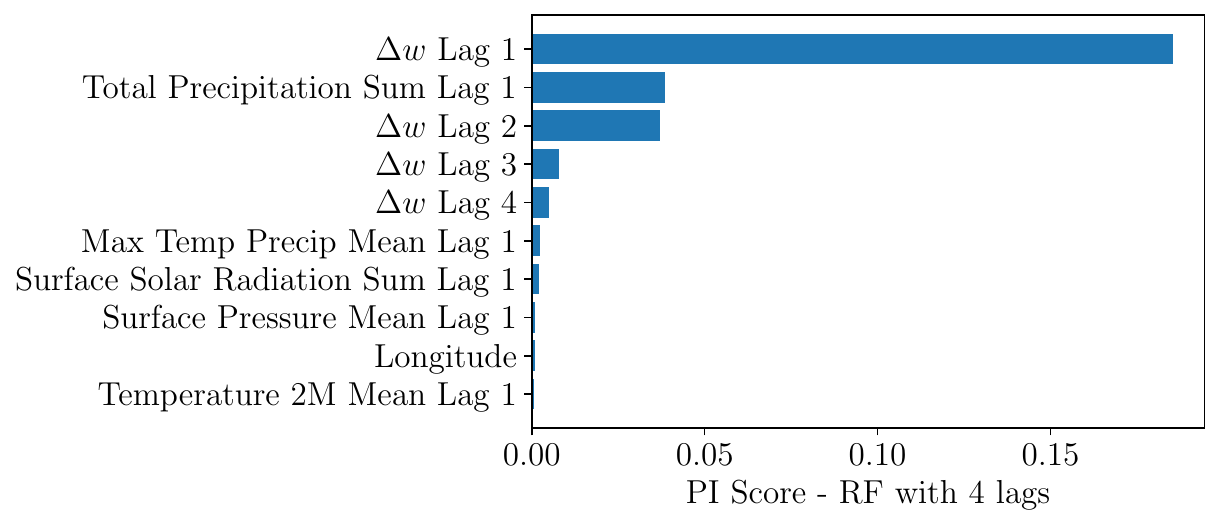} &
        \includegraphics[width=0.45\textwidth]{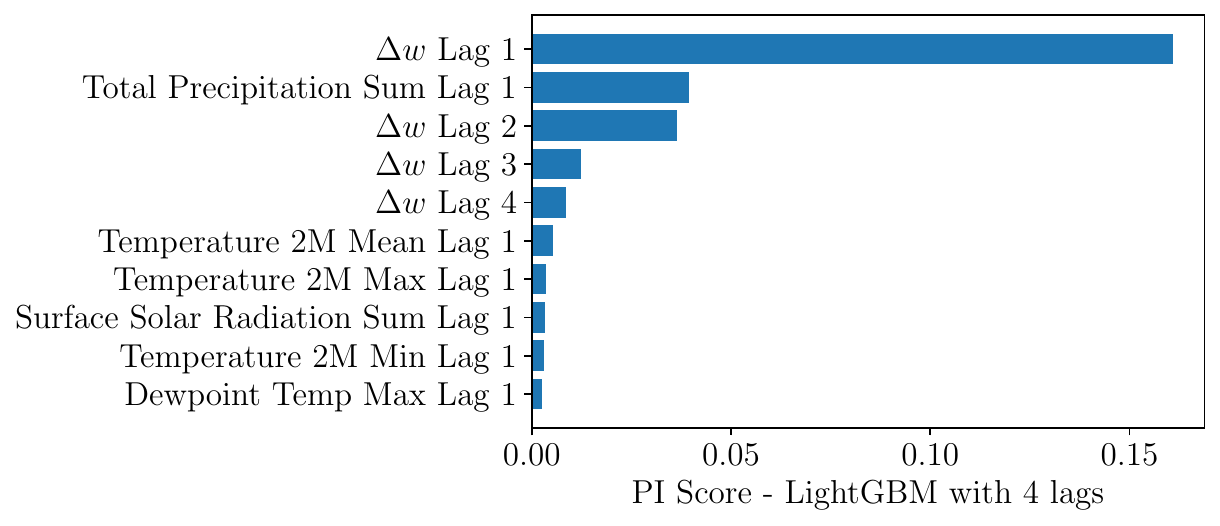}
    \end{tabular}
    \caption{Permutation-based feature importance for RF (left) and LGBM (right) across different lag configurations (top to bottom: 1 to 4 lags).}
    \label{fig:permutation_grid}
\end{figure}

While impurity-based measures highlight how often and how effectively features are used for splitting, they do not capture the actual impact of features on predictive performance. To address this limitation, we complement our analysis with permutation importance, reported in Fig.~\ref{fig:permutation_grid}. This method evaluates the importance of each feature by randomly permuting its values and measuring the resulting decrease in the model's accuracy. A feature is considered important if permuting its values significantly drops the model's performance. This technique provides a more comprehensive view of feature importance, considering the feature's effect on the tree structure and its impact on the model's predictive accuracy. Combining both techniques usually provides a more comprehensive understanding of the relative importance of each feature in the model. The permutation results reinforce the dominant role of lagged hive weight variations and, to a lesser extent, weather variables.

It is worth noting that, unlike the impurity-based scores, the permutation importance rankings appear more consistent across RF and LightGBM. This is expected, as permutation importance directly measures the predictive contribution of each feature by assessing the performance degradation caused by random shuffling. As a result, it tends to yield similar rankings when different models depend on the same set of informative predictors, even if their internal tree-building strategies differ.

\FloatBarrier

As a final instrument to shed light on the driver of the predictive capabilities of our machine learning models, we use the SHapley Additive exPlanation (SHAP) framework (see \citep{shapley201617, lundberg2017unified}). This approach explains a complex nonlinear model by shedding light on the contribution of each input feature to the output formation. For each input vector $x \in \mathbb{R}^K$, where $K$ is the number of features, and a model $f$, the SHAP value $\phi_i(f,x)$ for $i = 1, \ldots, K$ quantifies the effect (in a sense, the importance) of the $i$-th feature on the output $f(x)$. To compute this effect, for any subset $S\subseteq{1,\ldots,K}$, we define $f_S(x)$ as the model output when only the features in $S$ are used (with missing features marginalized or set to a baseline). The contribution of the $i$-th feature is then given by $f_{S\cup{i}}(x) - f_S(x)$.The SHAP value is defined as the weighted average 
\begin{equation}
\phi_{i}(f, x)=\sum_{S\subseteq \{1,\ldots, K\}\setminus \{i\}}\frac{\left|S\right| !\left(K-\left|S\right|-1\right) !}{K !}\left[f_{S\cup\{i\}}(x)-f_S(x)\right],
\end{equation}
where the weights ensure that $\sum_i \phi_i = f(x)$.

\begin{figure}[htbp]
    \centering
    \begin{tabular}{cc}
        \includegraphics[width=0.45\textwidth]{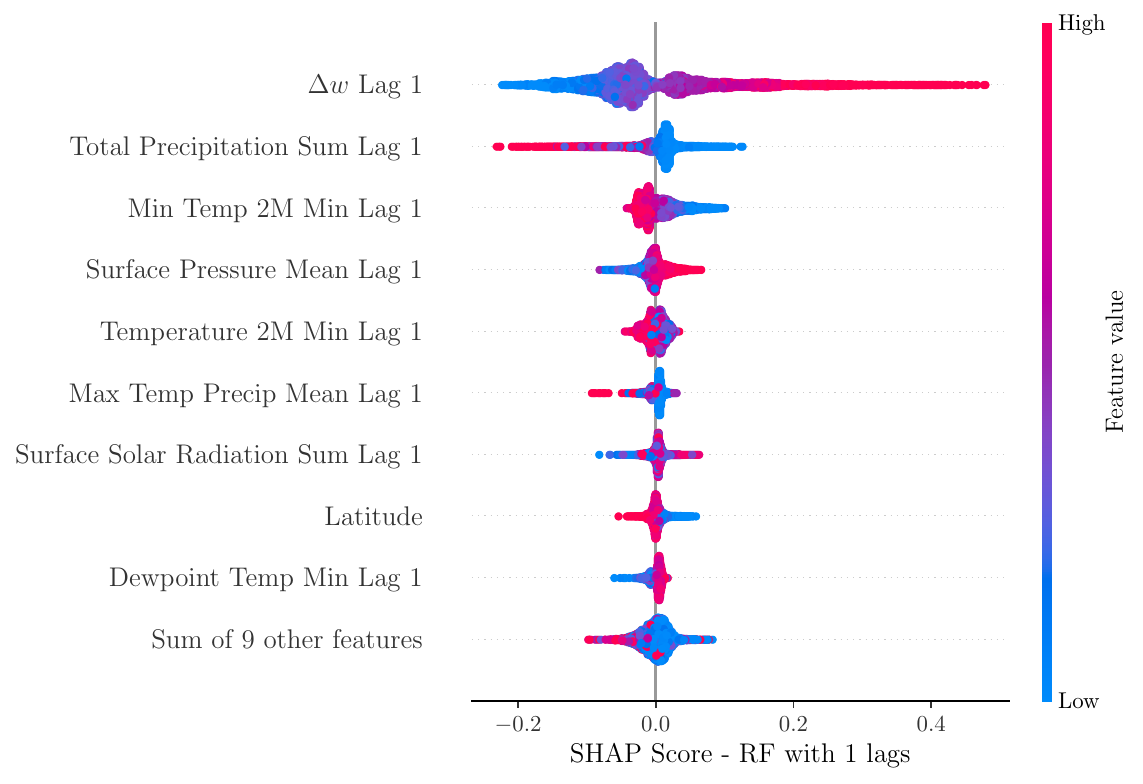} &
        \includegraphics[width=0.45\textwidth]{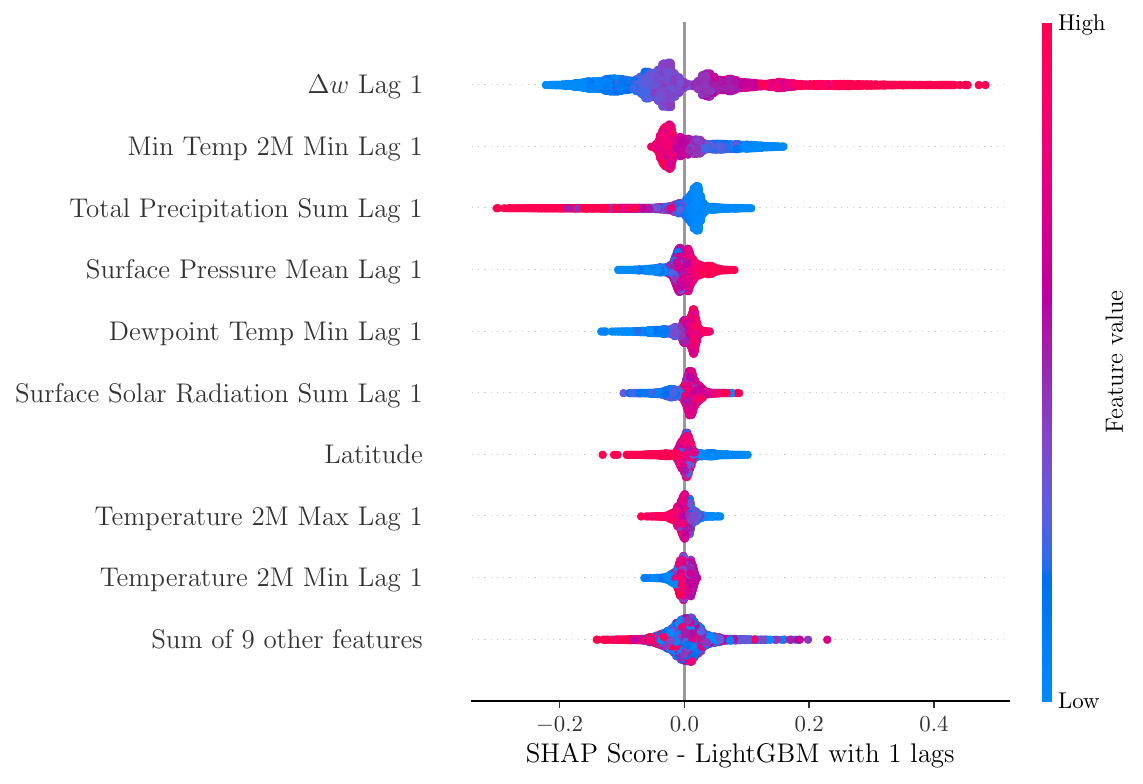} \\
        \includegraphics[width=0.45\textwidth]{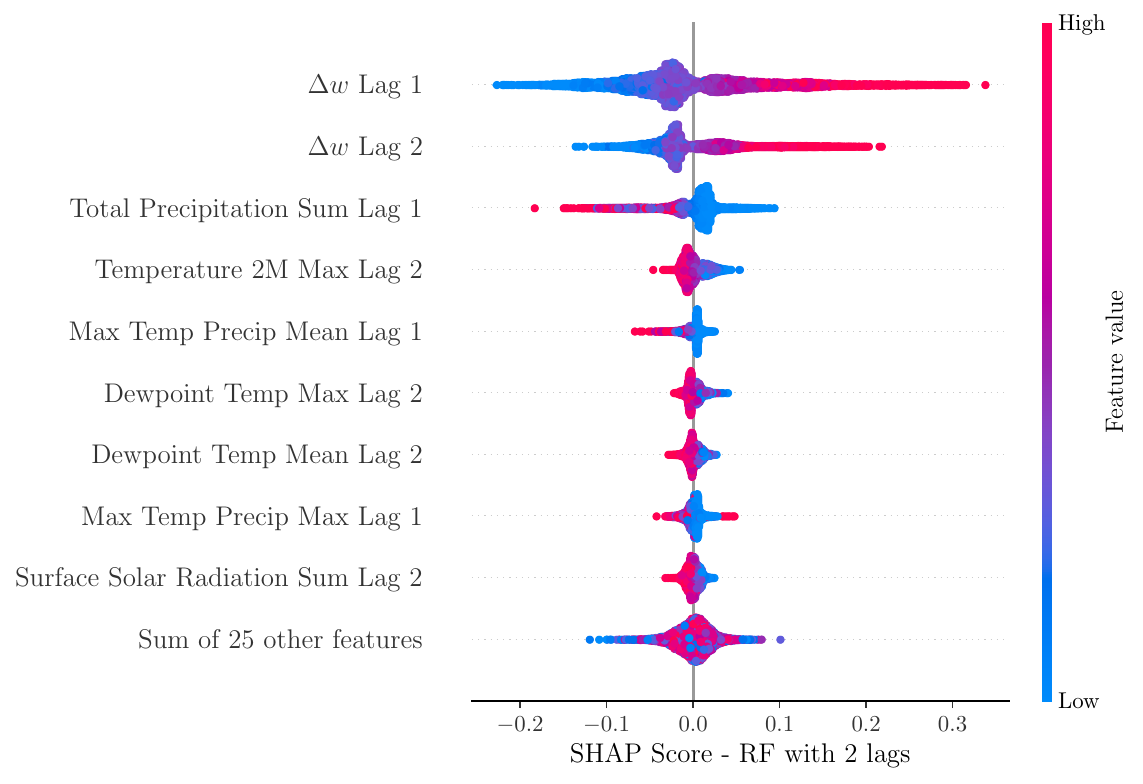} &
        \includegraphics[width=0.45\textwidth]{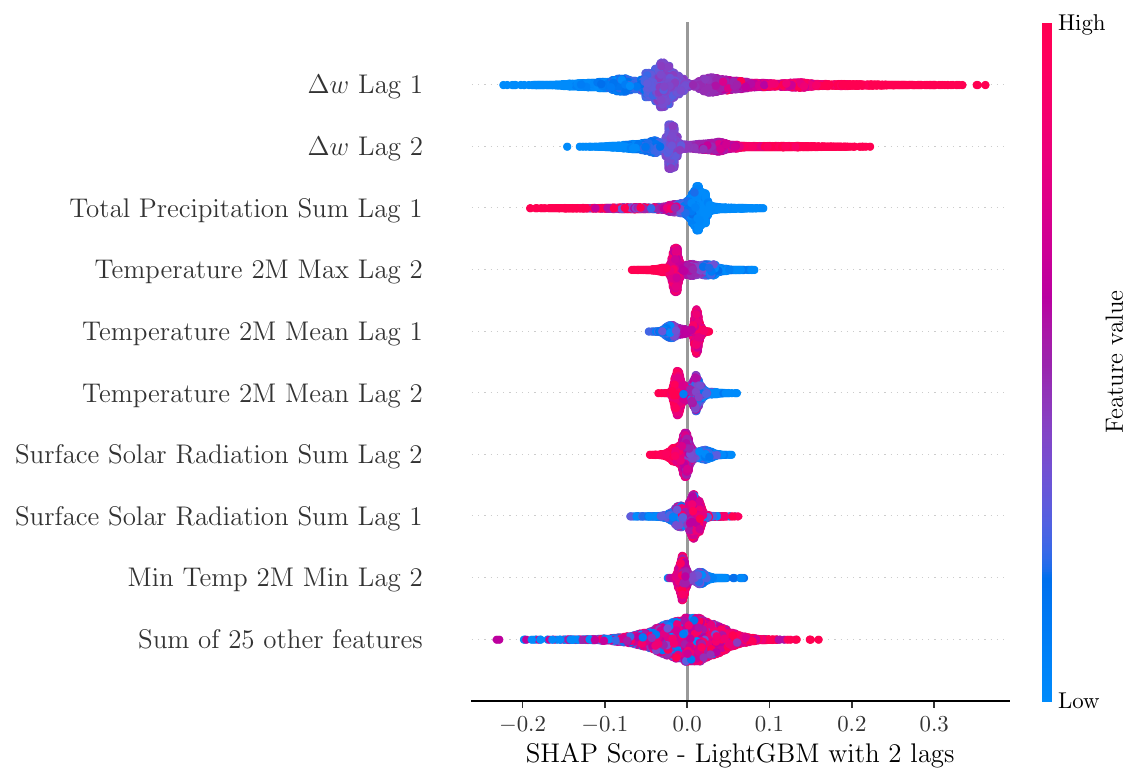} \\
        \includegraphics[width=0.45\textwidth]{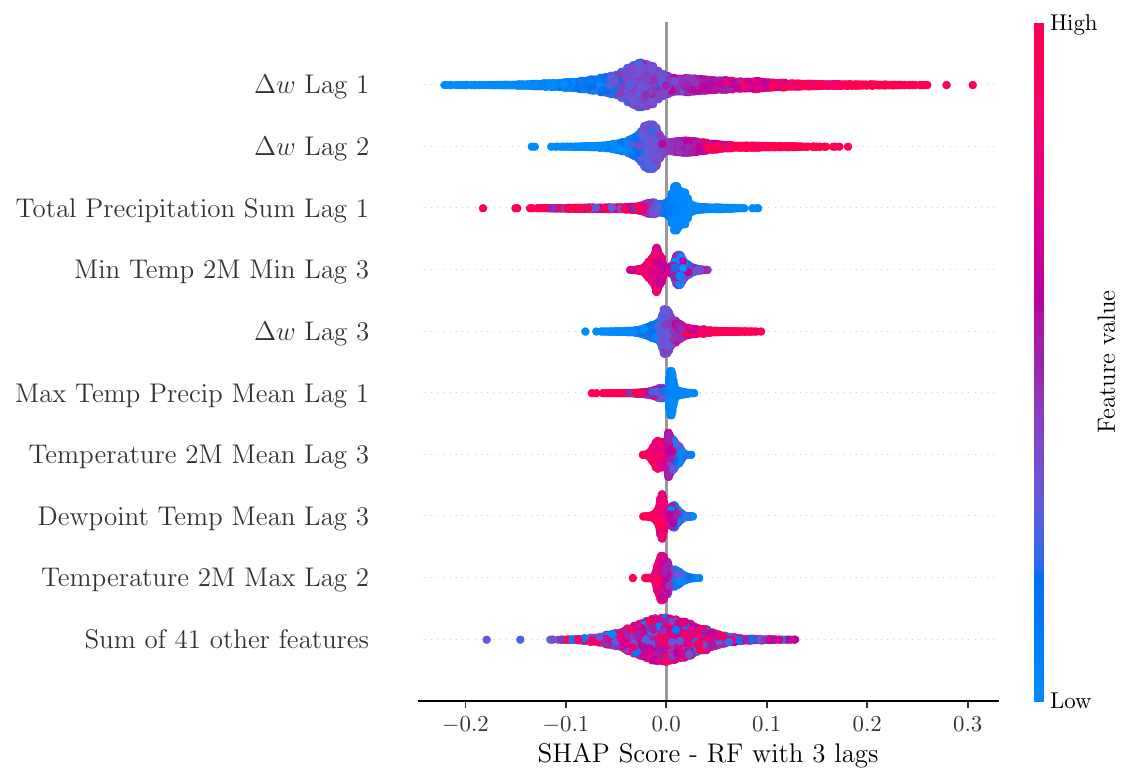} &
        \includegraphics[width=0.45\textwidth]{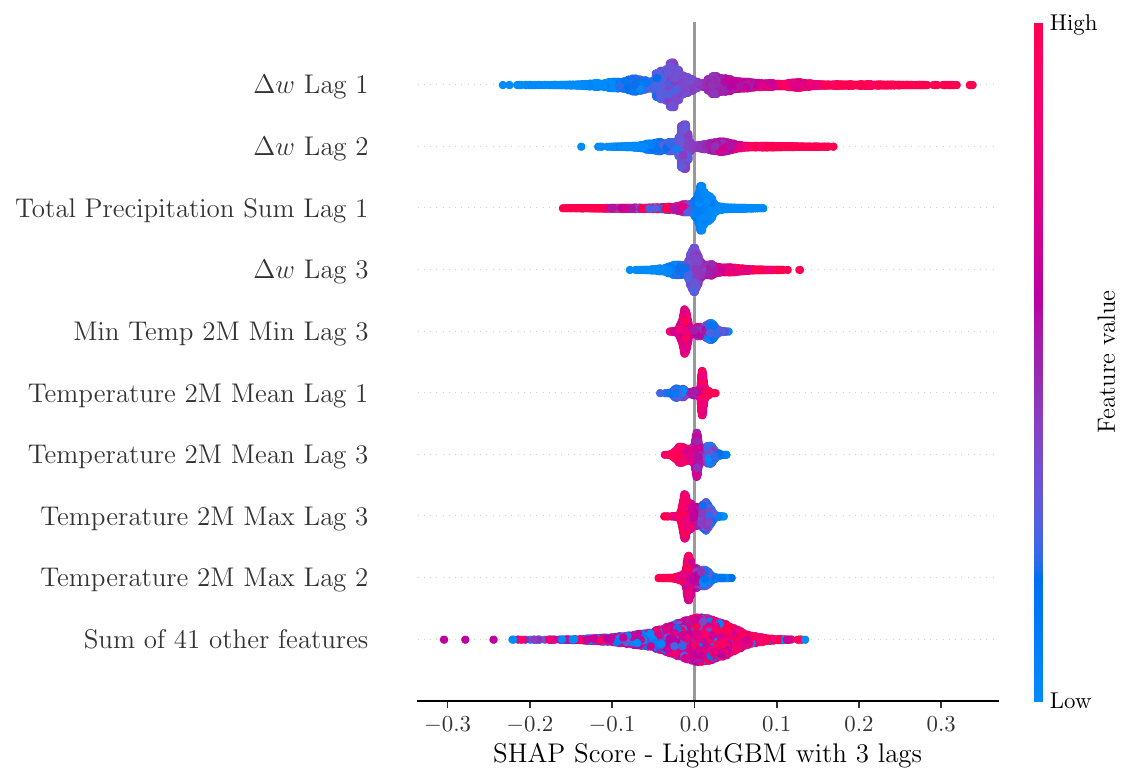} \\
        \includegraphics[width=0.45\textwidth]{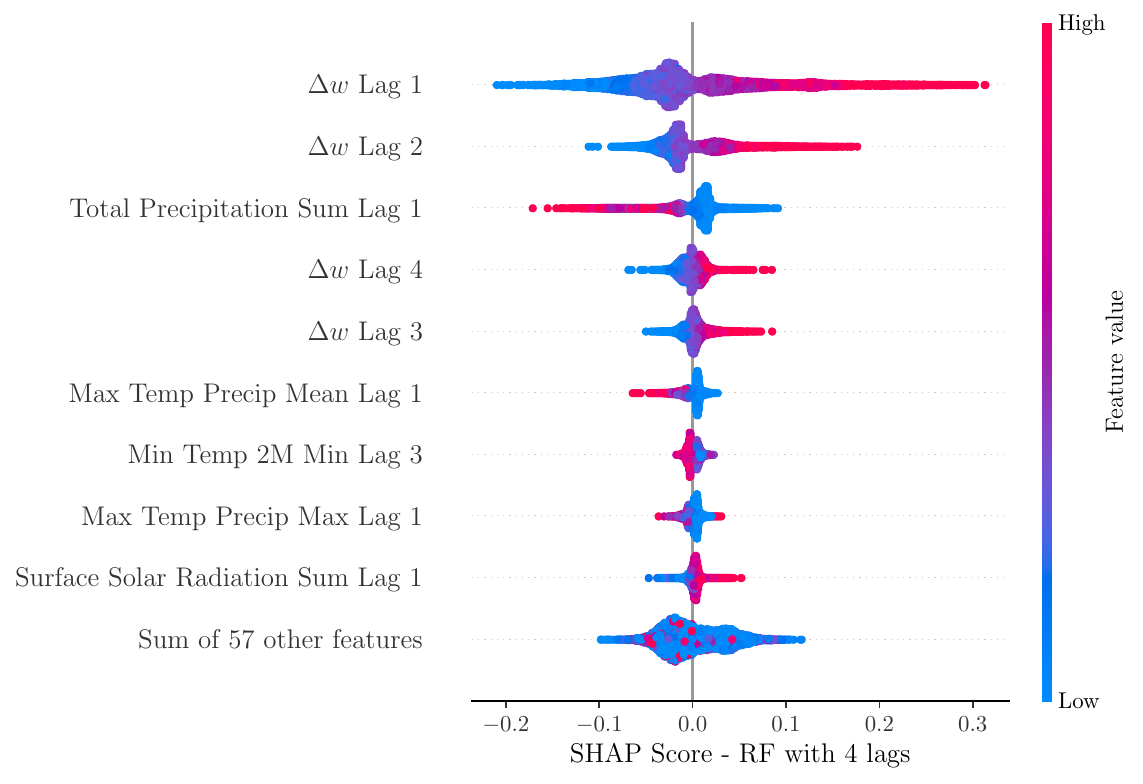} &
        \includegraphics[width=0.45\textwidth]{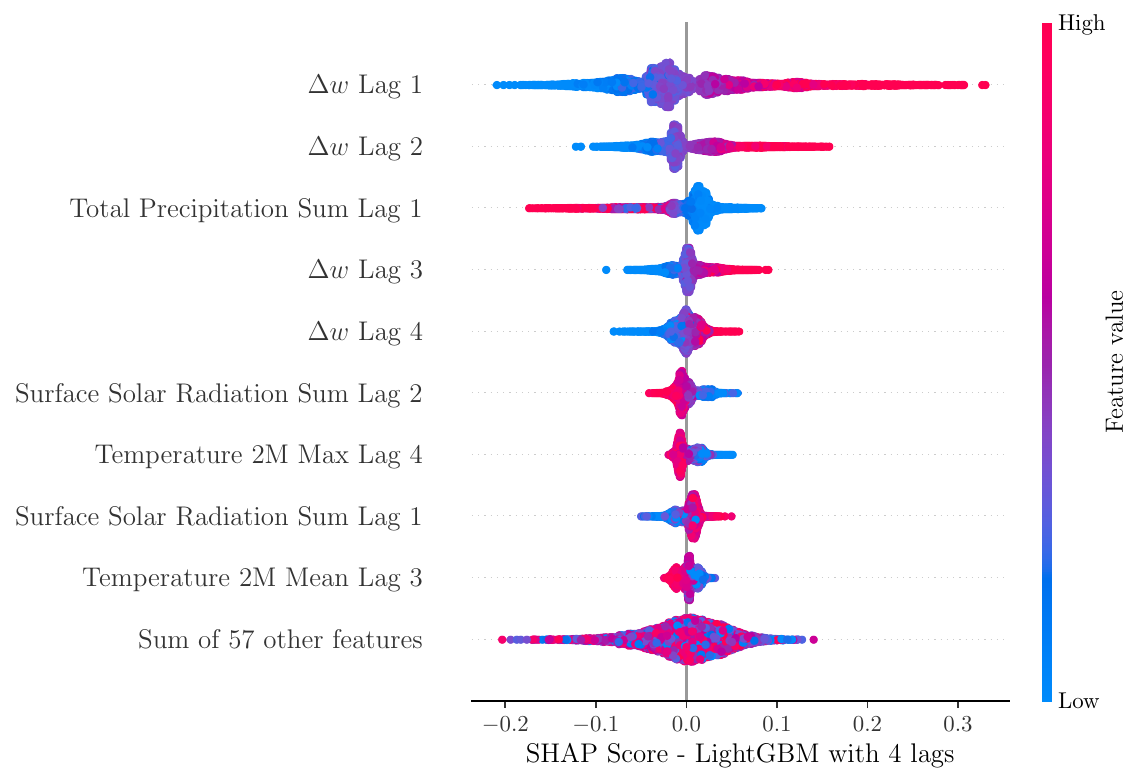}
    \end{tabular}
    \caption{SHAP beeswarm plots for RF (left) and LGBM (right) across different lag configurations (top to bottom: 1 to 4 lags).}
    \label{fig:shap_grid}
\end{figure}

Fig.~\ref{fig:shap_grid} presents SHAP beeswarm plots for RF and LightGBM across all lag configurations. SHAP values decompose each prediction into additive feature contributions, highlighting both the magnitude and direction of each input’s effect. In both models, lagged hive weight variations consistently emerge as the most influential predictors, with the first lag ranking highest across configurations and higher past values generally associated with increased predicted weight. Meteorological variables such as air temperature and total precipitation also appear prominently, preserving their relevance even as more lags are introduced. While RF tends to yield more compact SHAP distributions, LightGBM shows broader spreads for certain features, reflecting more heterogeneous local effects. Despite these differences, the two models rely on similar informative predictors to drive their forecasts.

It is worth noting that the patterns uncovered through SHAP confirm the trends observed in both impurity-based and permutation importance scores. Despite the methodological differences between Elastic Net, RF, and LightGBM, the three approaches consistently highlight the dominant role of autoregressive features and the complementary influence of selected weather variables.

Overall, the interpretability analysis reveals high consistency across modeling approaches. Lagged features related to hive weight dynamics dominate all importance rankings, a pattern that also emerges from the Elastic Net coefficients. These insights carry practical value for stakeholders connected to the beekeeping sector, including beekeepers, insurers, and policy designers, as they reveal the central importance of short-term hive weight dynamics and selected weather conditions, such as solar radiation and precipitation, in shaping predictive outcomes and guiding risk management strategies.

\FloatBarrier

\section{Ensemble Modeling and Aggregated Predictions}\label{Sec:ensemble}

After evaluating the performance and interpretability of individual models, we investigate whether combining different models through ensemble techniques can further improve predictive accuracy. By aggregating the predictions of multiple trained models, ensemble methods can exploit complementary strengths and mitigate individual weaknesses.

To construct ensemble models \citep{acar2009ensemble, gastinger2021study, wu2021ensemble}, we first collect the stored predictions from all individual models trained on the test set across different lag configurations. For each lag, we perform a temporal pseudo-split of the test data at the hive level, allocating the earlier 50\% of observations as a pseudo-training set and the remaining 50\% as a pseudo-test set. This split preserves the temporal ordering of the data to avoid information leakage.

We investigate two main classes of ensemble strategies: (i) basic combination methods and (ii) meta-modeling approaches. As a simple baseline, we adopt a weighted averaging scheme, assigning weights inversely proportional to each individual model’s RMSE, thereby giving more influence to better-performing models.

Beyond basic combinations, we implement meta-modeling techniques through stacking, where a secondary model (meta-learner) is trained to optimally combine the base model predictions. In this framework, the meta-learner is trained on out-of-sample predictions generated by the base models, learning how to best aggregate them into a final forecast. We adopt several meta-modeling strategies. First, we apply stacking using Ridge regression ($\lambda_1=0.1$) as the meta-learner. To further explore the benefits of sparsity and shrinkage, we apply stacking with LASSO regression ($\lambda_2=0.001$) and Elastic Net regression ($\lambda_1=0.001$, $\lambda_2=0.5$). These linear meta-models provide a trade-off between robustness and sparsity when combining base predictions.

In addition, we explore a residual-based ensemble, fitting a Gradient Boosting Regressor directly on the individual models’ predictions. The residual model is trained with 50 estimators, a learning rate of 0.01, and a fixed random seed for reproducibility.

Finally, we implement stacking via a small FNN. Prior to training, base model predictions are standardized by removing the mean and scaling to unit variance. The FNN consists of one hidden layer with 20 neurons and sigmoid activation, followed by a linear output layer. The model is optimized with Adam (learning rate of 0.0001) to minimize the mean squared error loss over 100 epochs with a batch size of 64.

Tab.~\ref{tab:ensemble_performance} reports the performance of all ensemble methods on the pseudo-test set across the four evaluation metrics introduced in Sec.~\ref{sec:empirical}. Among the methods considered, stacking with Elastic Net achieves the best overall performance in most cases, consistently yielding the lowest MAE and Log-Cosh Loss across lag configurations, and the lowest RMSE for most lags. Ridge and LASSO stacking also perform competitively, particularly for higher lag configurations. The residual-based ensemble and FNN stacking slightly lag behind but still improve over basic weighted averaging.

We do not optimize the hyperparameters of the aggregation and stacking methods, as splitting the pseudo-test set further to construct an additional validation set would substantially reduce the available data. As a result, performance may improve beyond our reported results if hyperparameters were tuned more carefully, rather than relying on the heuristic choices adopted in this analysis.

Overall, these results highlight the effectiveness of meta-modeling approaches, particularly those leveraging linear regularization, in enhancing predictive performance. Stacking with Elastic Net consistently achieves the best overall performance across different lag configurations, confirming that appropriately balancing sparsity and shrinkage leads to robust and accurate ensemble predictions. Comparing ensemble results with the best individual models reported in Tab.~\ref{Tab:model_performance_test} further reinforces this conclusion: ensemble techniques, particularly Elastic Net stacking, systematically improve upon the top-performing single models across RMSE, MAE, and Log-Cosh metrics. These findings underscore the value of combining complementary models to capture richer patterns in hive weight dynamics.

\begin{table}[htbp]
\centering
\begin{tabular}{c c | cccc}
\toprule
 &  & \multicolumn{4}{c}{Test} \\
\midrule
Ensemble Method & Lag & RMSE & NRMSE & MAE & Log-Cosh \\
\midrule
\multirow{4}{*}{Weighted Averaging} 
    & 1 & 0.273778 & 14.104577 & 0.195929 & 0.035482 \\
    & 2 & 0.273667 & 14.098892 & 0.195907 & 0.035466 \\
    & 3 & 0.273697 & 14.100426 & 0.195401 & 0.035494 \\
    & 4 & 0.278297 & 14.337406 & 0.196087 & 0.036147 \\
\midrule
\multirow{4}{*}{Stacking (Ridge)} 
    & 1 & 0.249169 & 12.836792 & 0.170604 & 0.029450 \\
    & 2 & 0.239308 & 12.328729 & 0.163603 & 0.027244 \\
    & 3 & 0.238669 & 12.295837 & 0.163208 & 0.027075 \\
    & 4 & \textit{0.235122} & \textit{12.113097} & 0.161039 & \textit{0.026337} \\
\midrule
\multirow{4}{*}{Stacking (LASSO)} 
    & 1 & 0.249411 & 12.849252 & 0.169832 & 0.029435 \\
    & 2 & 0.239503 & 12.338773 & 0.163433 & 0.027289 \\
    & 3 & 0.241061 & 12.419048 & 0.163485 & 0.027498 \\
    & 4 & 0.235628 & 12.139181 & \textit{0.160893} & 0.026443 \\
\midrule
\multirow{4}{*}{Stacking (Elastic Net)} 
    & 1 & 0.248797 & 12.817595 & 0.169663 & 0.029335 \\
    & 2 & 0.239021 & 12.313948 & 0.163041 & 0.027180 \\
    & 3 & 0.237740 & 12.247952 & 0.162306 & 0.026885 \\
    & 4 & \textbf{0.235090} & \textbf{12.111453} & \textbf{0.160556} & \textbf{0.026324} \\
\midrule
\multirow{4}{*}{Residual-Based Ensemble} 
    & 1 & 0.260089 & 13.399337 & 0.176805 & 0.032009 \\
    & 2 & 0.254784 & 13.126032 & 0.173560 & 0.030770 \\
    & 3 & 0.252206 & 12.993218 & 0.171765 & 0.030175 \\
    & 4 & 0.251649 & 12.964552 & 0.171514 & 0.030051 \\
\midrule
\multirow{4}{*}{Stacking (FNN)} 
    & 1 & 0.248693 & 12.812238 & 0.170264 & 0.029343 \\
    & 2 & 0.242615 & 12.499103 & 0.165045 & 0.027942 \\
    & 3 & 0.236552 & 12.186765 & 0.161813 & 0.026646 \\
    & 4 & 0.236513 & 12.184761 & 0.161640 & 0.026630 \\
\midrule
\bottomrule
\end{tabular}
\caption{The table reports the predictive performance of each ensemble method across different feature sets defined by the number of included lags. For each method and lag configuration, we report RMSE, NRMSE, MAE, and Log-Cosh evaluated on the pseudo-test set. The best-performing ensemble method for each metric is bolded, while the second-best is italicized.}
\label{tab:ensemble_performance}
\end{table}
\FloatBarrier

\section{Conclusions}\label{sec:conclus}
This paper investigates the forecasting of hive weight variations using a dataset combining precision beekeeping and weather observations across Italy. By modeling hive dynamics at a daily frequency, we evaluate a broad set of predictive models, including regularized linear methods, tree-based ensembles, and feedforward neural networks. We further explore ensemble learning strategies to enhance forecast accuracy and analyze feature importance to gain insights into the drivers of hive weight variation.

Our findings highlight the critical role of lagged hive weight dynamics and weather conditions, particularly surface solar radiation, precipitation and minimum temperature, in shaping predictive outcomes. Elastic Net and Random Forest models consistently achieve strong out-of-sample performance, with ensemble techniques, especially Elastic Net stacking, offering systematic improvements over individual models across multiple evaluation metrics.

Interpretability analyses across linear and nonlinear models confirm the dominant influence of short-term autoregressive components, supported by complementary meteorological indicators. These insights have potential applications in improving beekeeping practices and designing risk management tools for mitigating climate-induced honey production losses.

Several limitations emerge. The relatively limited number of hives constrains our ability to construct dedicated validation sets for hyperparameter optimization in the ensemble models. Additionally, our modeling framework assumes static hive locations and does not incorporate interactions across hives or spatial dependencies.

Future work can extend this framework by integrating multi-seasonal data, modeling hive relocations explicitly, and exploring spatially aware machine learning approaches. Incorporating additional predictors such as vegetation indices or disease markers can further enrich the predictive models and support the development of robust early-warning systems for beekeepers.

\section*{Acknowledgements}
\noindent This work is part of BEEkeepers Weather indexed INsurance project (BEEWIN), "Bando Miele 2021", funded by the Italian Ministry of Agricultural, Food and Forestry Policies. The person responsible for the project is Prof. Maria Elvira Mancino from DISEI, Department of Economics and Management, University of Florence, to whom our heartfelt thanks go.

\vspace{1em}

\noindent The authors thank 3Bee S.R.L. for providing the bee hives data. 

\vspace{1em}

\noindent E.G. thanks INdAM for the support of applied mathematical research activity. Hereby, E.G. also expresses gratitude to Guido Cioni and Matteo Puglini, Ph.D. alumni of Max Planck Institute f\"{u}r Meteorologie, for having friendly shared their expertise in weather data mining.

\bibliographystyle{apalike}
\bibliography{biblio}

\newpage
\appendix

\section{Hyperparameter and Design Choices}
\label{App:hyper}

This appendix outlines the hyperparameter tuning procedures and design decisions adopted throughout the empirical analysis. For every lag configuration considered, the dataset is consistently split into training, validation, and test sets using a time-aware strategy. The data is partitioned independently for each hive, ensuring that validation and test samples occur strictly after the training period to prevent information leakage. We reserve $30\%$ of the data for testing and allocate $10\%$ of the remaining data to validation. We adopt this setup consistently for all models.

\textit{Linear models with regularization.} For LASSO, Ridge, and Elastic Net, we perform a grid search over the regularization parameters, selecting the configuration that minimizes validation mean squared error (MSE). For Ridge, we tune $\lambda_1 \in \{10^{-6}, 5 \cdot 10^{-6}, 10^{-5}, 5 \cdot 10^{-5}, 10^{-4}, 5 \cdot 10^{-4}, 10^{-3}, 5 \cdot 10^{-3}, 10^{-2}, 5 \cdot 10^{-2}, 0.1, 0.5, 1.0\}$. For LASSO, we tune $\lambda_2 \in \{10^{-6}, 5 \cdot 10^{-6}, 10^{-5}, 5 \cdot 10^{-5}\}$. For Elastic Net, we search over $\lambda_1$ using the same grid as Ridge, and tune the $\lambda_2$ ratio over the same grid as LASSO. All the regularized linear models are trained on standardized features to ensure consistent application of regularization across predictors.

\textit{Tree-based ensemble models.} For Random Forest (RF), we tune the number of estimators $n \in \{50, 100, 200, 1000\}$ and the maximum tree depth $d \in \{5, 10, 20, 50\}$. For XGBoost and LightGBM, we follow a similar approach, exploring $n$, $d$, and the learning rate $\text{lr} \in \{0.001, 0.01, 0.1\}$. No feature scaling is applied to tree-based models.

\textit{Feedforward neural network.} We explore four architectures with one or two hidden layers, using either 64 or 128 units per layer, i.e., $\{[64], [128], [64, 64], [128, 128]\}$. Models are trained using the Adam optimizer for 250 epochs with mini-batches of size 64. We tune the learning rate $\text{lr}$ over $\{10^{-5}, 10^{-4}, 10^{-3}\}$ and select the best configuration based on validation MSE. Input features are standardized before training.

The optimal hyperparameter configurations identified through this process are summarized in Tab.~\ref{tab:optimal_hyperparameters}.

\begin{table}[H]
\centering
\begin{tabular}{lcccc}
\toprule
\textbf{Model} & \textbf{Lag 1} & \textbf{Lag 2} & \textbf{Lag 3} & \textbf{Lag 4} \\
\midrule
Elastic Net ($\lambda_1$, $\lambda_2$) & (0.5, $10^{-6}$) & (0.5, $10^{-6}$) & (0.5, $10^{-6}$) & (1.0, $10^{-6}$) \\
LASSO ($\lambda_2$) & $5 \cdot 10^{-5}$ & $5 \cdot 10^{-5}$ & $5 \cdot 10^{-5}$ & $5 \cdot 10^{-5}$ \\
Ridge ($\lambda_1$) & 1.0 & 1.0 & 1.0 & 1.0 \\
RF ($n$, $d$) & (1000, 10) & (1000, 10) & (1000, 10) & (1000, 10) \\
XGBoost ($n$, $d$, lr) & (1000, 10, 0.001) & (1000, 10, 0.001) & (1000, 10, 0.001) & (1000, 5, 0.01) \\
LightGBM ($n$, $d$, lr) & (1000, 5, 0.01) & (1000, 5, 0.01) & (1000, 5, 0.01) & (1000, 20, 0.01) \\
FNN (layers, lr) & [128], $10^{-5}$ & [64, 64], $10^{-5}$ & [64, 64], $10^{-5}$ & [64, 64], $10^{-5}$ \\
\bottomrule
\end{tabular}
\caption{Optimal Hyperparameters by Lag Configuration}
\label{tab:optimal_hyperparameters}
\end{table}

\section{Validation Results}\label{App:val}

\begin{table}[h]
\centering
\begin{tabular}{c c | cccc}
\toprule
Model & Lag & RMSE & NRMSE & MAE & Log-Cosh \\
\midrule
\multirow{4}{*}{OLS} & 1 & 0.279024 & \textbf{-2.790994} & 0.195235 & 0.036488 \\
& 2 & 0.272536 & -2.639839 & 0.189224 & 0.034849 \\
& 3 & 0.267248 & -2.682899 & 0.184987 & 0.033573 \\
& 4 & 0.267511 & -2.748509 & 0.184897 & 0.033605 \\
\midrule
\multirow{4}{*}{LASSO} & 1 & 0.278836 & -2.785322 & 0.194966 & 0.036440 \\
& 2 & 0.272224 & -2.634440 & 0.188846 & 0.034774 \\
& 3 & 0.266661 & -2.662884 & 0.184279 & 0.033431 \\
& 4 & 0.266806 & -2.722094 & 0.184139 & 0.033437 \\
\midrule
\multirow{4}{*}{Ridge} & 1 & 0.279003 & \textit{-2.790161} & 0.195206 & 0.036483 \\
& 2 & 0.272507 & -2.638912 & 0.189186 & 0.034842 \\
& 3 & 0.267202 & -2.680912 & 0.184931 & 0.033561 \\
& 4 & 0.267455 & -2.745862 & 0.184834 & 0.033592 \\
\midrule
\multirow{4}{*}{Elastic Net} & 1 & 0.277660 & -2.629812 & 0.205036 & 0.036535 \\
& 2 & 0.267190 & -2.444872 & 0.190927 & 0.033791 \\
& 3 & 0.260412 & -2.378592 & 0.183342 & 0.032116 \\
& 4 & \textit{0.258178} & -2.382576 & 0.180283 & 0.031556 \\
\midrule
\multirow{4}{*}{RF} & 1 & 0.272832 & -2.596831 & 0.184702 & 0.034872 \\
& 2 & 0.264483 & -2.514725 & 0.178365 & 0.032869 \\
& 3 & 0.259412 & -2.458934 & 0.176311 & 0.031679 \\
& 4 & 0.258767 & -2.433015 & \textbf{0.174410} & \textit{0.031521} \\
\midrule
\multirow{4}{*}{XGBoost} & 1 & 0.273143 & -2.452090 & 0.193189 & 0.035209 \\
& 2 & 0.264191 & -2.338493 & 0.179441 & 0.032823 \\
& 3 & 0.259698 & -2.406412 & 0.176374 & 0.031765 \\
& 4 & 0.259287 & -2.408342 & 0.176132 & 0.031674 \\
\midrule
\multirow{4}{*}{LightGBM} & 1 & 0.273172 & -2.619571 & 0.185647 & 0.035000 \\
& 2 & 0.264305 & -2.375417 & 0.179436 & 0.032836 \\
& 3 & 0.259264 & -2.360151 & 0.175883 & 0.031647 \\
& 4 & \textbf{0.257919} & -2.375884 & \textit{0.175067} & \textbf{0.031353} \\
\midrule
\multirow{4}{*}{FNN} & 1 & 0.273193 & -2.766911 & 0.186358 & 0.034979 \\
& 2 & 0.265596 & -2.497455 & 0.181159 & 0.033143 \\
& 3 & 0.261846 & -2.369601 & 0.179548 & 0.032294 \\
& 4 & 0.260518 & -2.353211 & 0.179269 & 0.031958 \\
\bottomrule
\end{tabular}
\caption{Model Performance Comparison on the Validation Set. The table reports the predictive performance of each model across different feature sets defined by the number of included lags. For each model and feature configuration, we report RMSE, NRMSE, MAE, and Log-Cosh evaluated on the test set. The best-performing model for each metric is bolded, while the second-best is italicized.}
\label{Tab:model_performance_val}
\end{table}
\FloatBarrier

\section{Errors Distributions}\label{App:error_distributions}

\begin{figure}[h]
    \centering
    \includegraphics[width=0.35\textwidth]{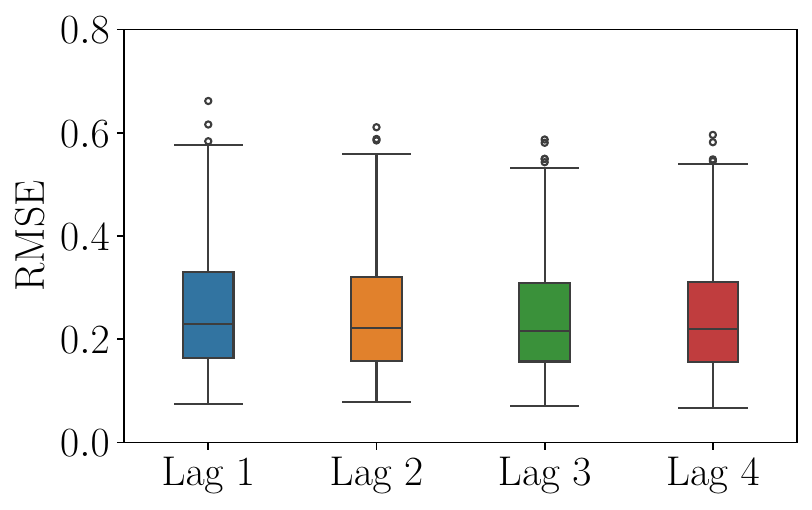}
    \hspace{0.04\textwidth}
    \includegraphics[width=0.35\textwidth]{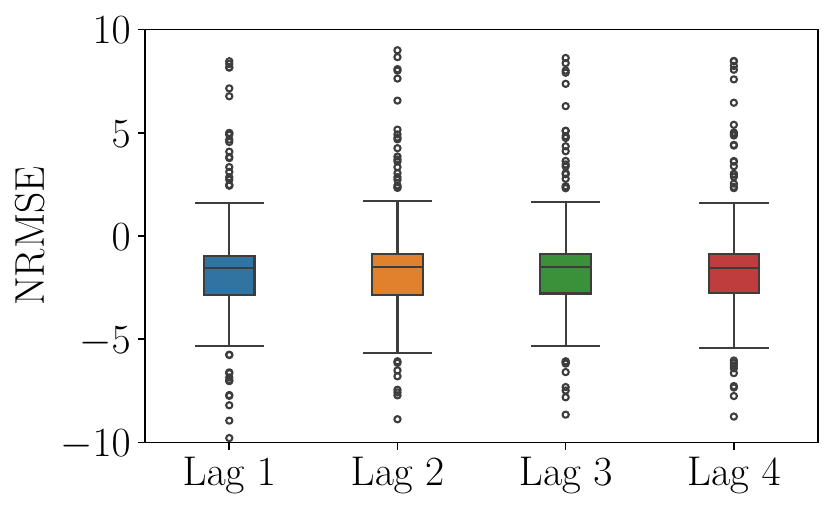}
    
    \vskip\baselineskip
    
    \includegraphics[width=0.35\textwidth]{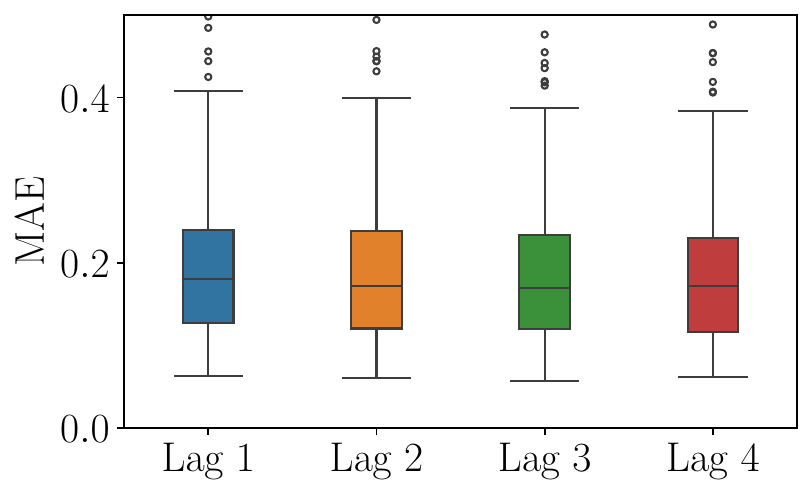}
    \hspace{0.04\textwidth}
    \includegraphics[width=0.35\textwidth]{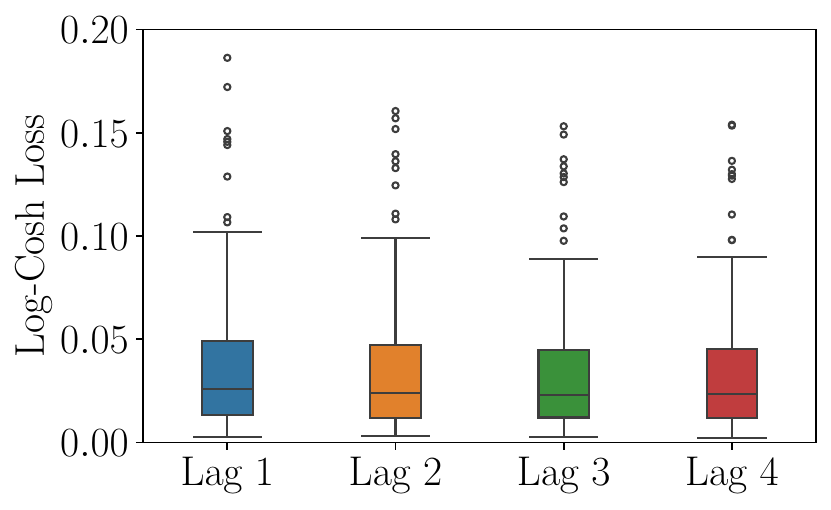}
    \caption{Prediction error distributions on the validation set for OLS across the four evaluation metrics.}
    \label{fig:ols_val_errors}
\end{figure}
\FloatBarrier

\begin{figure}[h]
    \centering
    \includegraphics[width=0.35\textwidth]{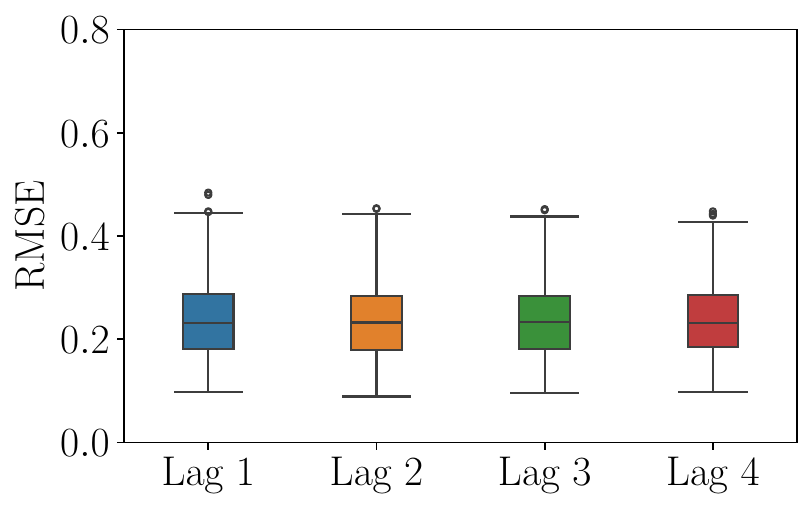}
    \hspace{0.04\textwidth}
    \includegraphics[width=0.35\textwidth]{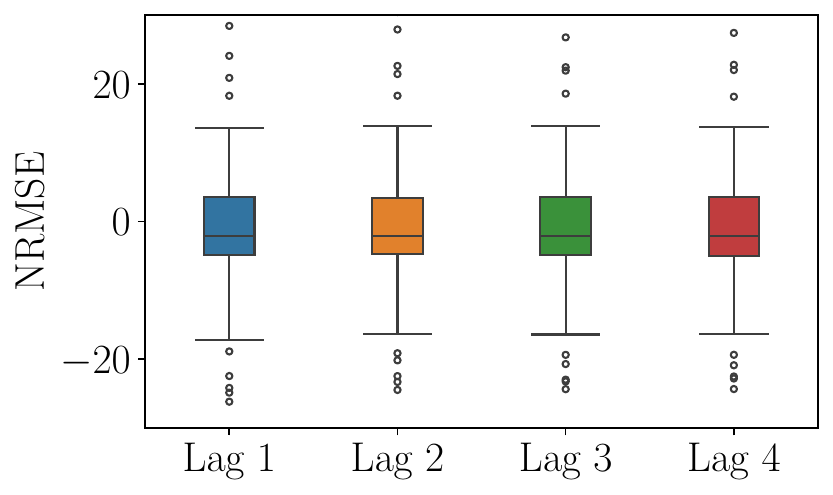}
    
    \vskip\baselineskip
    
    \includegraphics[width=0.35\textwidth]{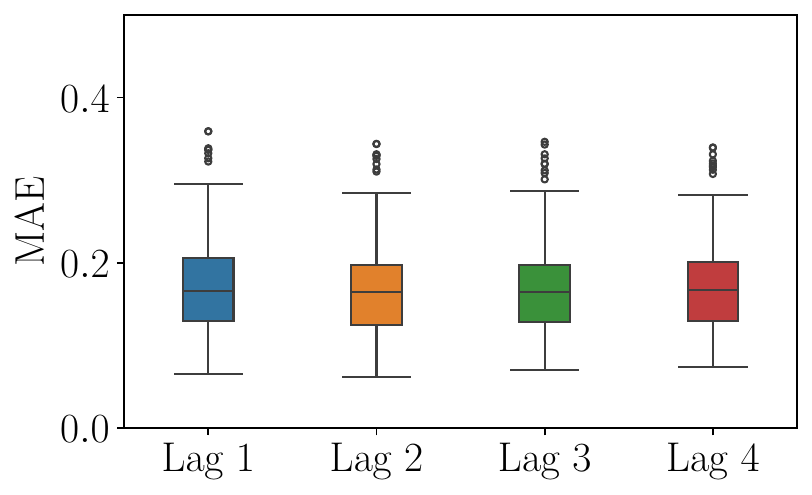}
    \hspace{0.04\textwidth}
    \includegraphics[width=0.35\textwidth]{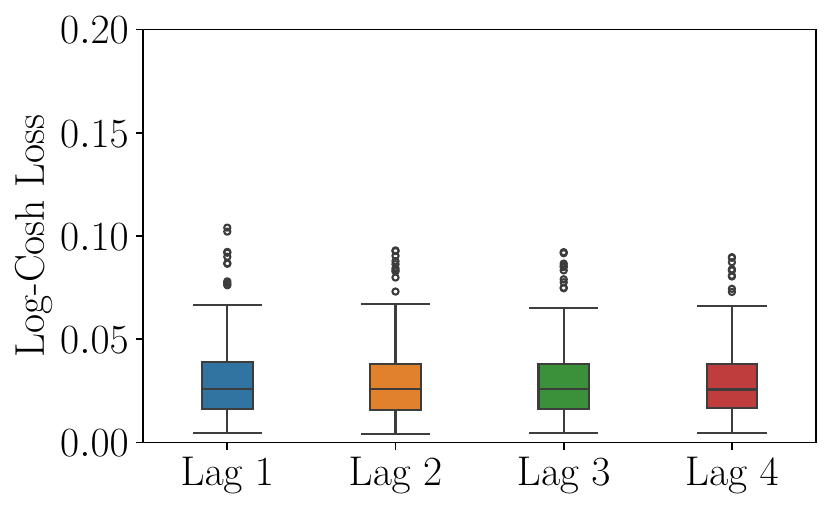}
    \caption{Prediction error distributions on the test set for OLS across the four evaluation metrics.}
    \label{fig:ols_test_errors}
\end{figure}
\FloatBarrier

\begin{figure}[h]
    \centering
    \includegraphics[width=0.35\textwidth]{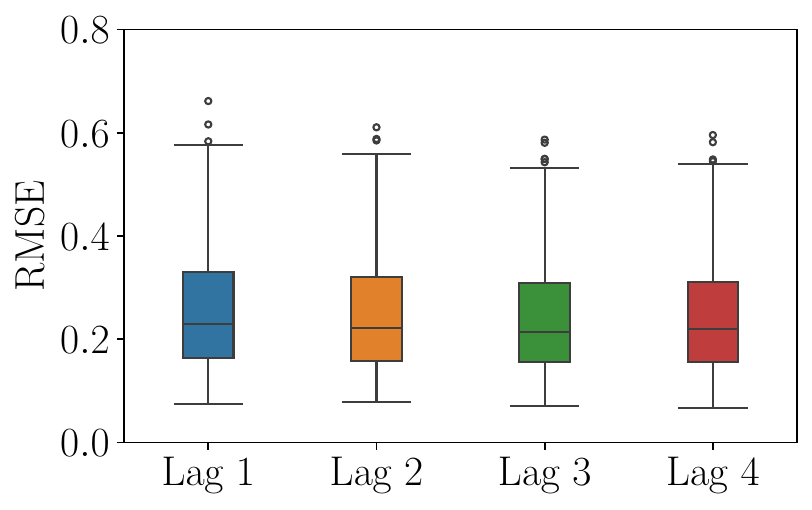}
    \hspace{0.04\textwidth}
    \includegraphics[width=0.35\textwidth]{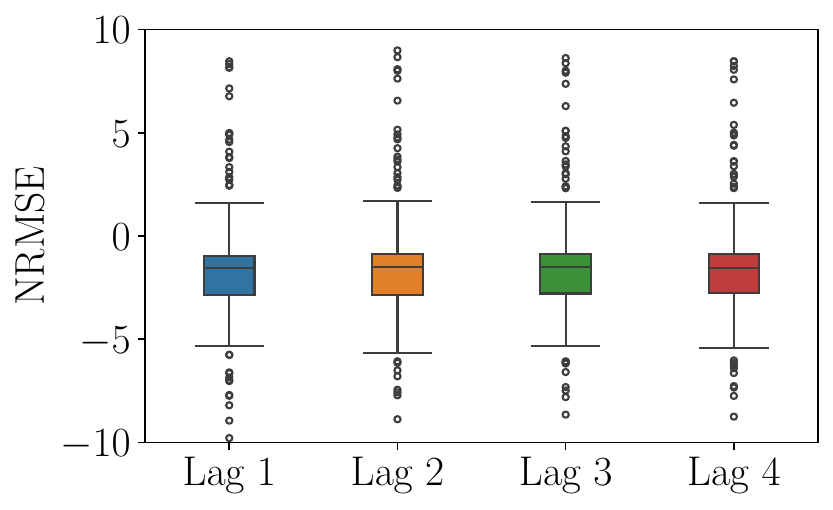}
    
    \vskip\baselineskip
    
    \includegraphics[width=0.35\textwidth]{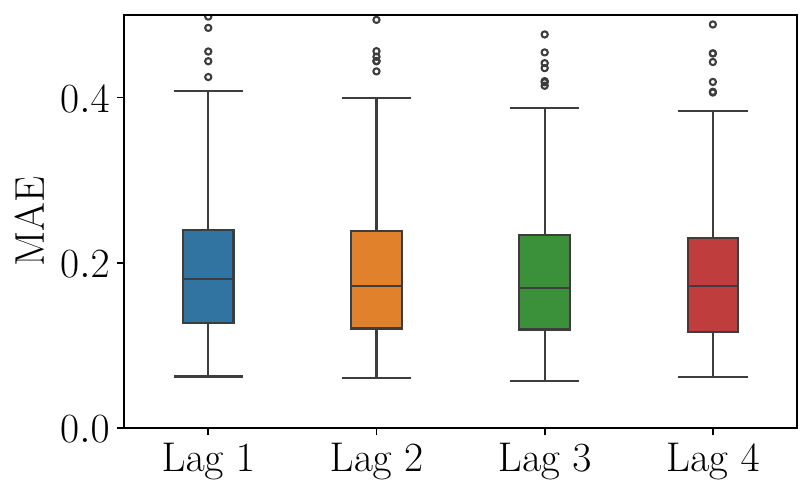}
    \hspace{0.04\textwidth}
    \includegraphics[width=0.35\textwidth]{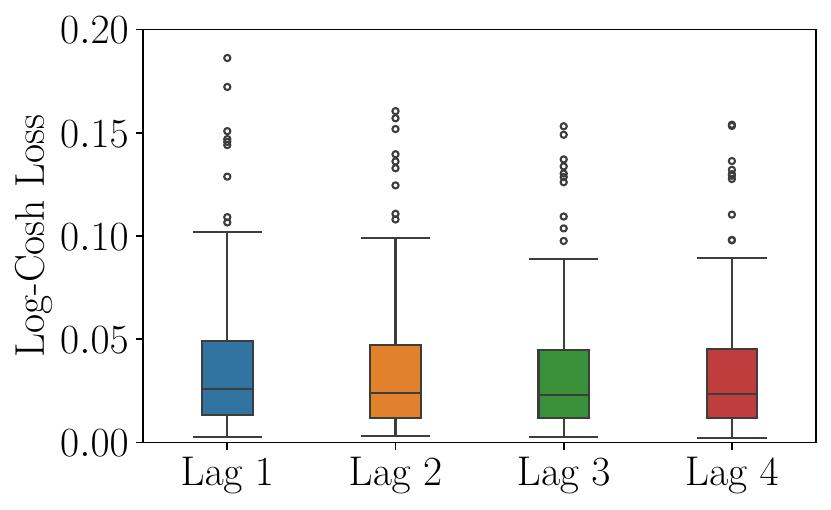}
    \caption{Prediction error distributions on the validation set for Ridge across the four evaluation metrics.}
    \label{fig:ridge_val_errors}
\end{figure}
\FloatBarrier

\begin{figure}[h]
    \centering
    \includegraphics[width=0.35\textwidth]{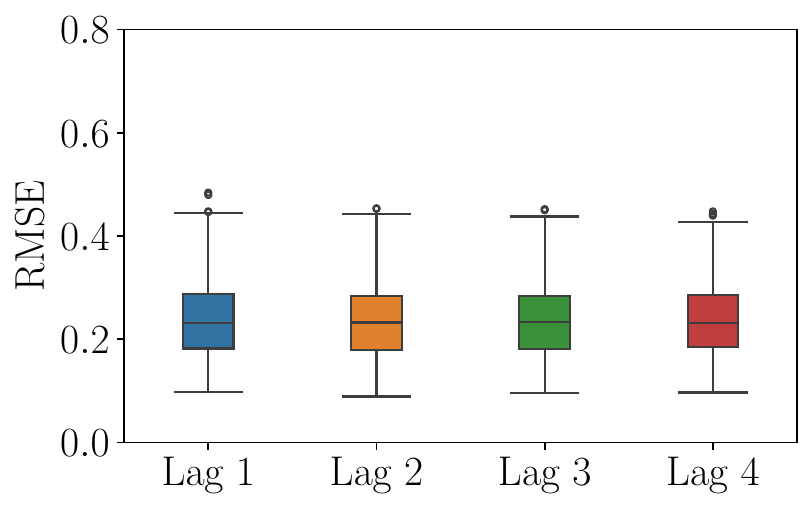}
    \hspace{0.04\textwidth}
    \includegraphics[width=0.35\textwidth]{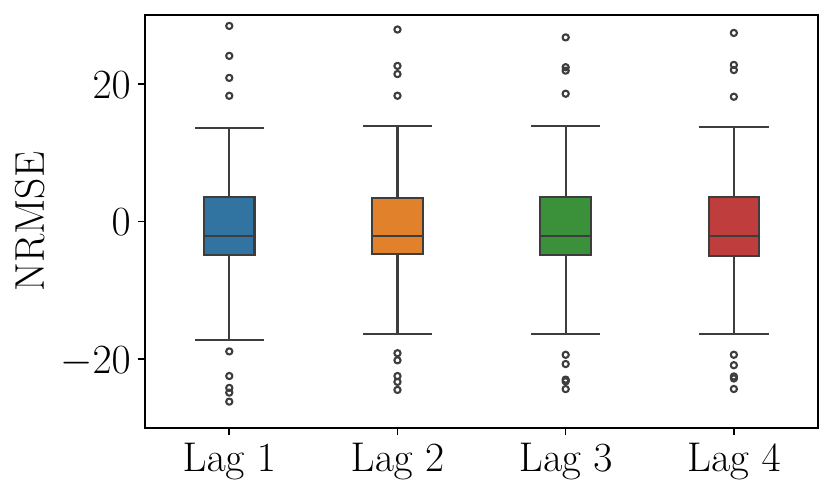}
    
    \vskip\baselineskip
    
    \includegraphics[width=0.35\textwidth]{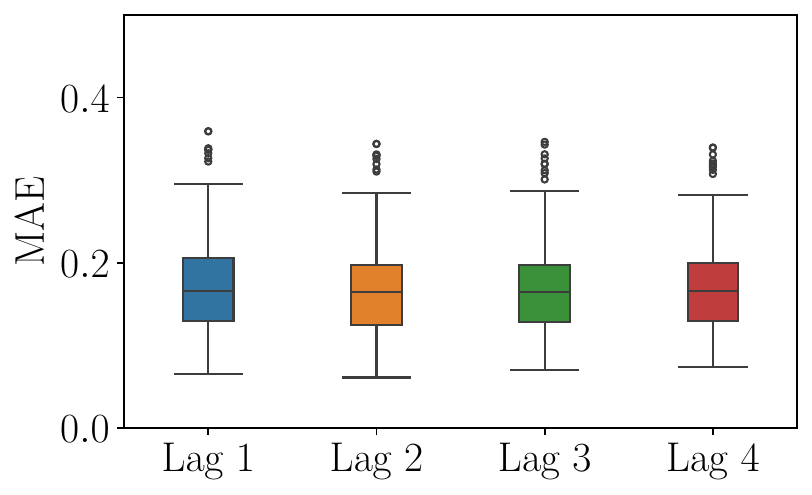}
    \hspace{0.04\textwidth}
    \includegraphics[width=0.35\textwidth]{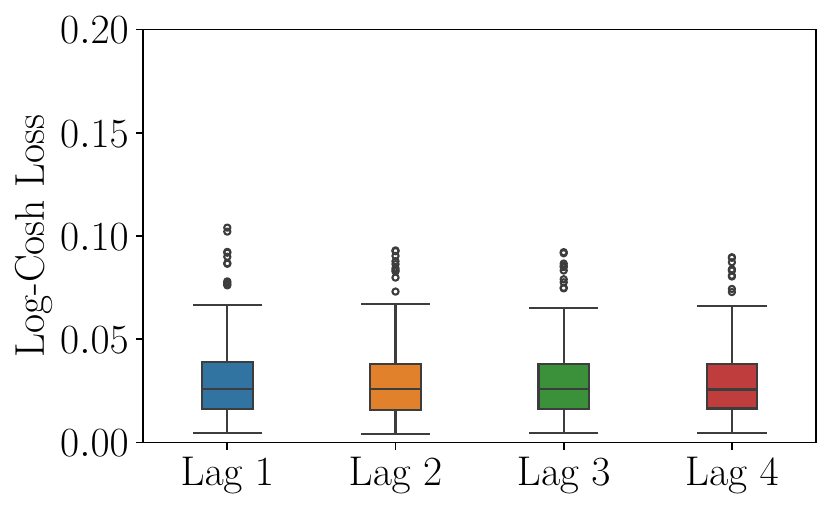}
    \caption{Prediction error distributions on the test set for Ridge across the four evaluation metrics.}
    \label{fig:ridge_test_errors}
\end{figure}
\FloatBarrier

\begin{figure}[h]
    \centering
    \includegraphics[width=0.35\textwidth]{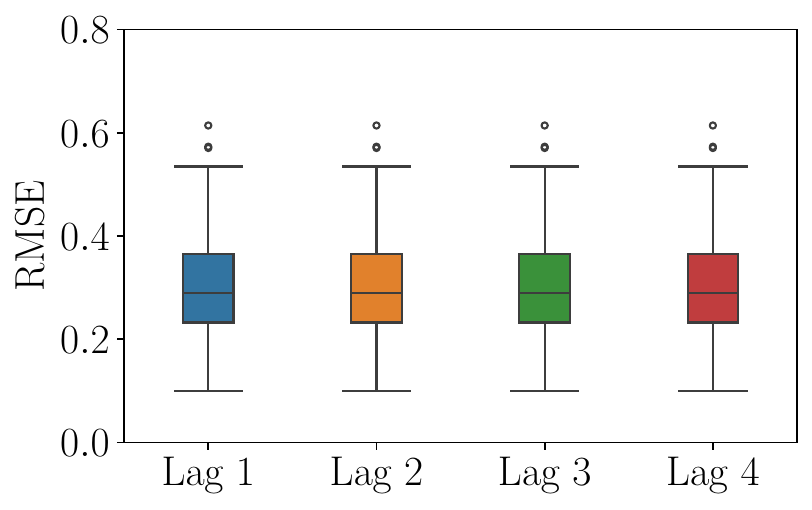}
    \hspace{0.04\textwidth}
    \includegraphics[width=0.35\textwidth]{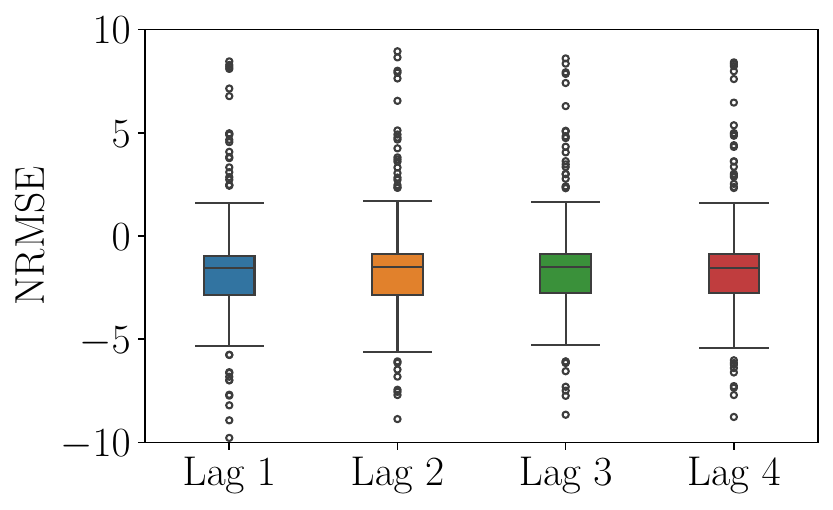}
    
    \vskip\baselineskip
    
    \includegraphics[width=0.35\textwidth]{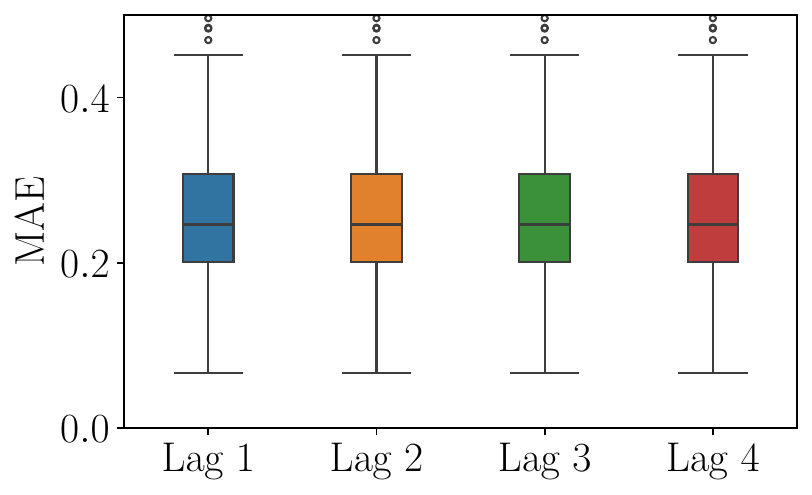}
    \hspace{0.04\textwidth}
    \includegraphics[width=0.35\textwidth]{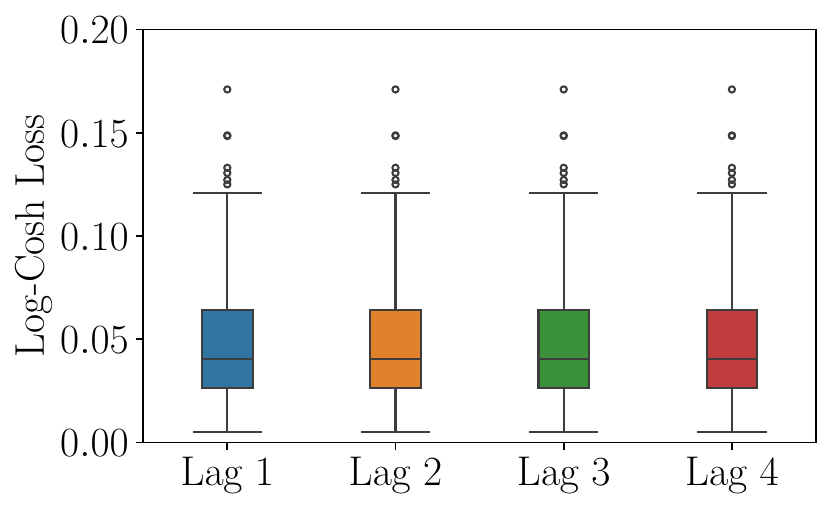}
    \caption{Prediction error distributions on the validation set for LASSO across the four evaluation metrics.}
    \label{fig:lasso_val_errors}
\end{figure}
\FloatBarrier

\begin{figure}[h]
    \centering
    \includegraphics[width=0.35\textwidth]{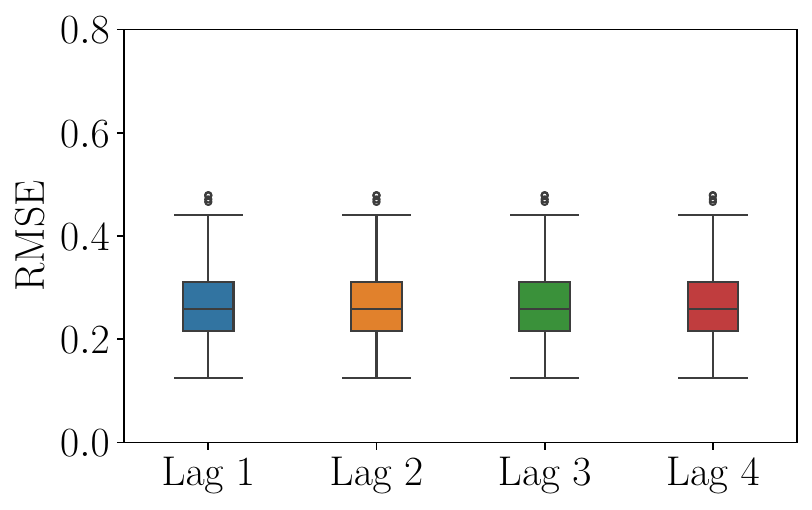}
    \hspace{0.04\textwidth}
    \includegraphics[width=0.35\textwidth]{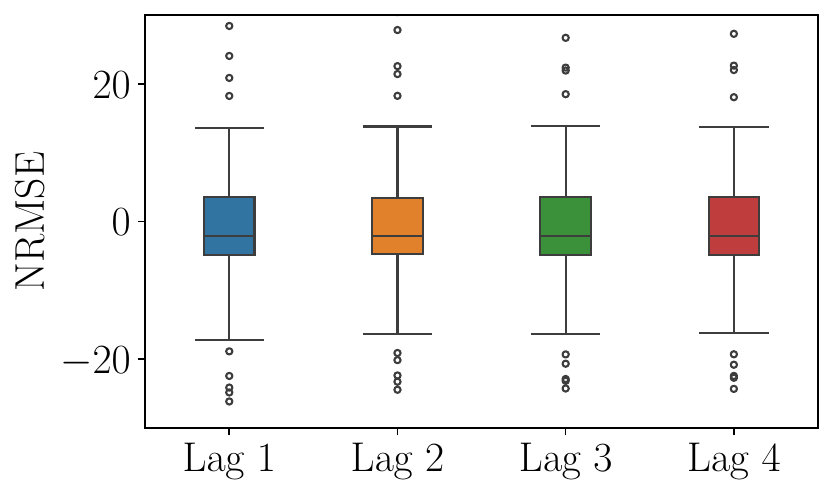}
    
    \vskip\baselineskip
    
    \includegraphics[width=0.35\textwidth]{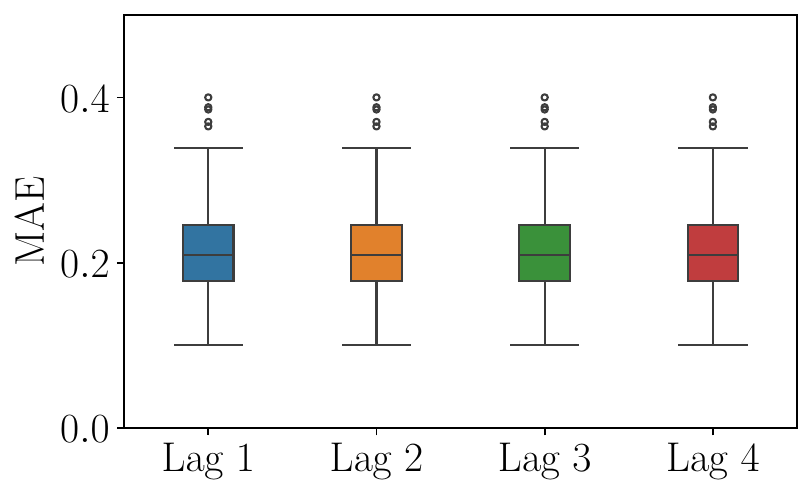}
    \hspace{0.04\textwidth}
    \includegraphics[width=0.35\textwidth]{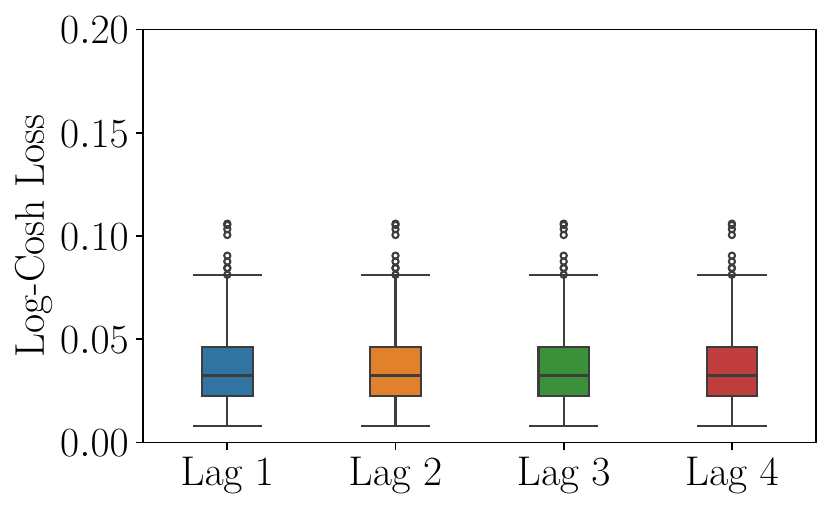}
    \caption{Prediction error distributions on the test set for LASSO across the four evaluation metrics.}
    \label{fig:lasso_test_errors}
\end{figure}
\FloatBarrier

\begin{figure}[h]
    \centering
    \includegraphics[width=0.35\textwidth]{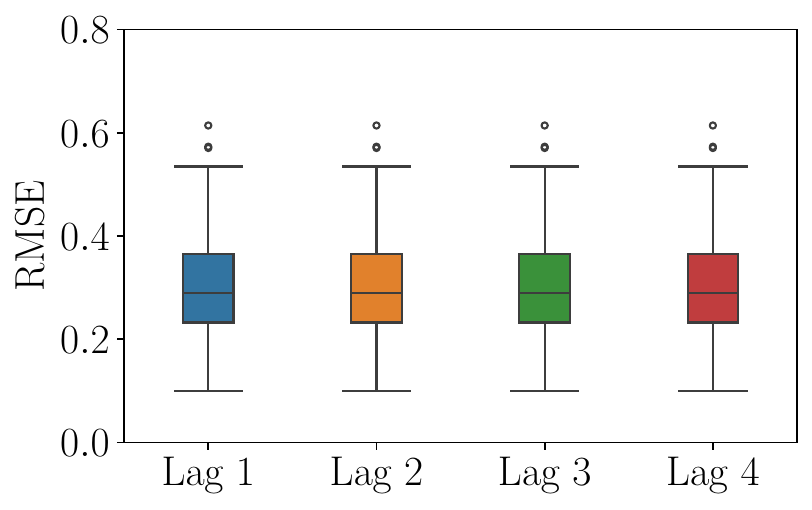}
    \hspace{0.04\textwidth}
    \includegraphics[width=0.35\textwidth]{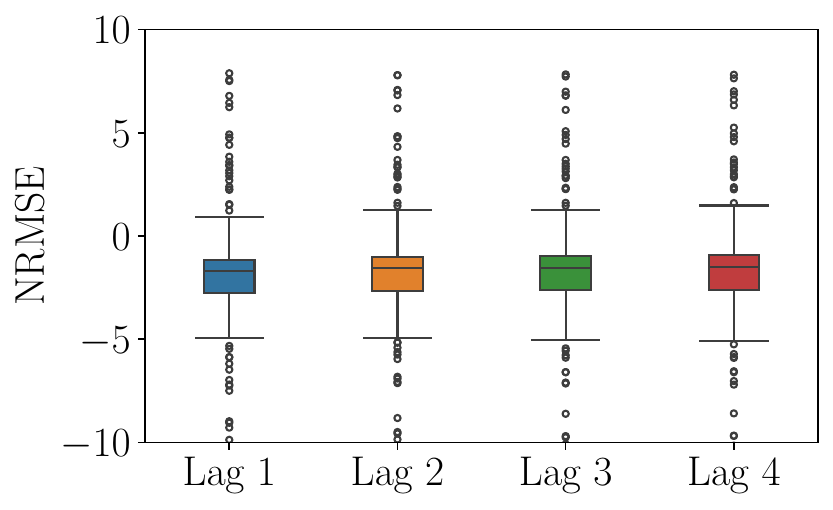}
    
    \vskip\baselineskip
    
    \includegraphics[width=0.35\textwidth]{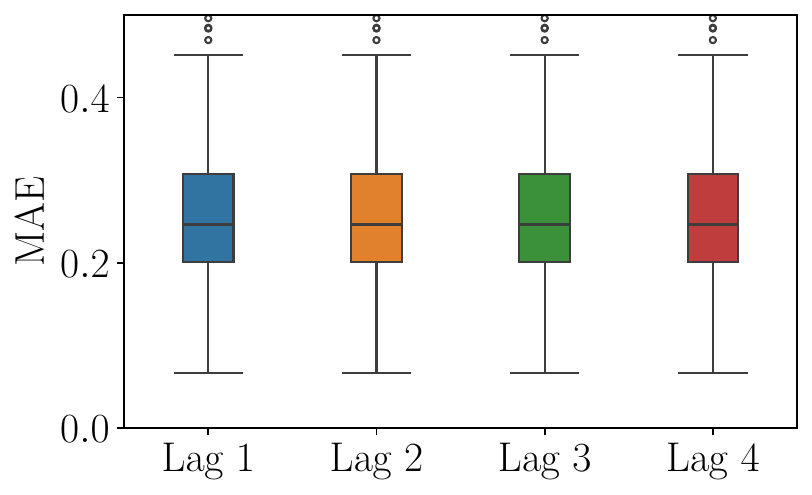}
    \hspace{0.04\textwidth}
    \includegraphics[width=0.35\textwidth]{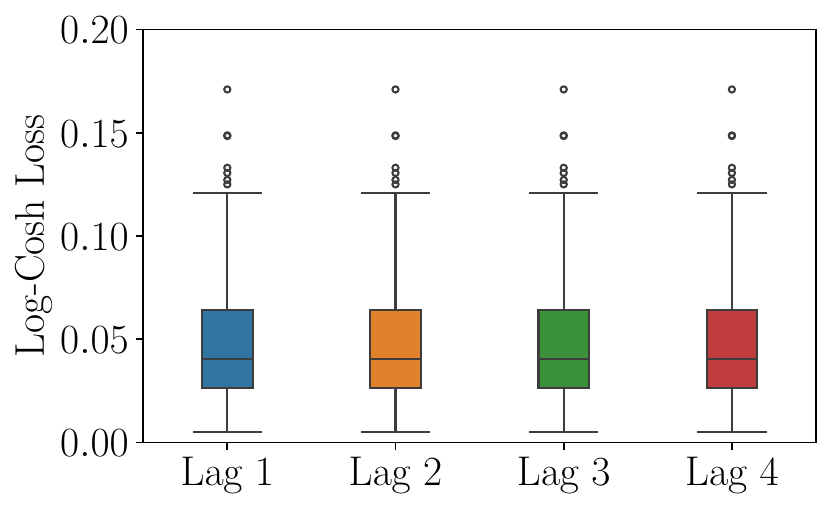}
    \caption{Prediction error distributions on the validation set for Elastic Net across the four evaluation metrics.}
    \label{fig:elasticnet_val_errors}
\end{figure}
\FloatBarrier

\begin{figure}[h]
    \centering
    \includegraphics[width=0.35\textwidth]{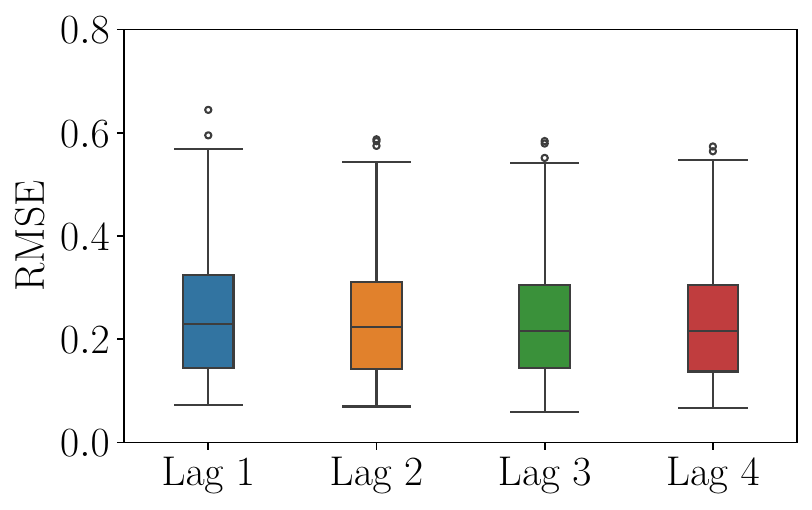}
    \hspace{0.04\textwidth}
    \includegraphics[width=0.35\textwidth]{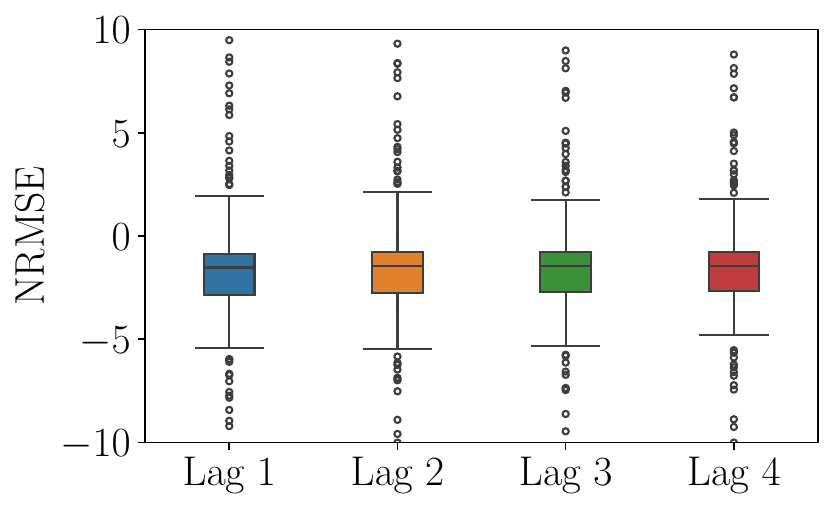}
    
    \vskip\baselineskip
    
    \includegraphics[width=0.35\textwidth]{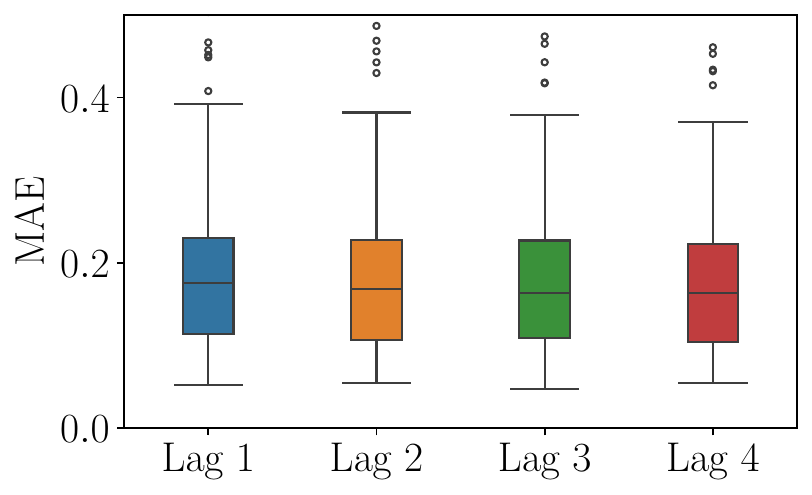}
    \hspace{0.04\textwidth}
    \includegraphics[width=0.35\textwidth]{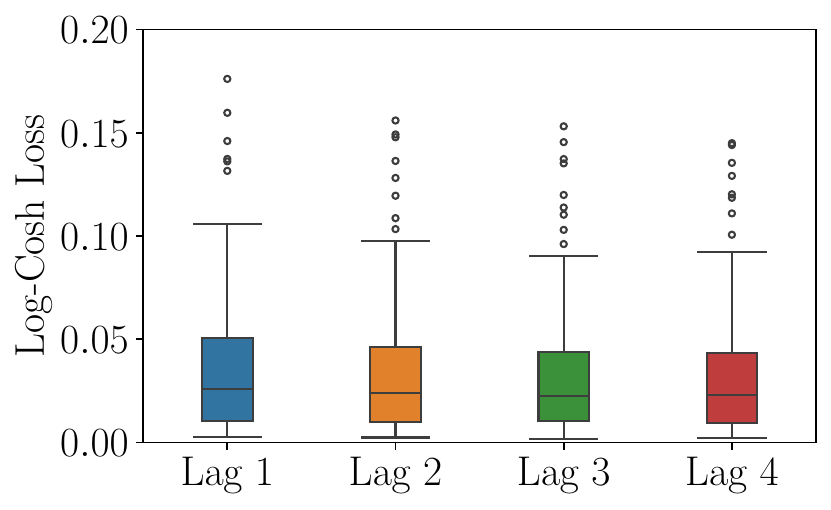}
    \caption{Prediction error distributions on the validation set for RF across the four evaluation metrics.}
    \label{fig:rf_val_errors}
\end{figure}
\FloatBarrier

\begin{figure}[h]
    \centering
    \includegraphics[width=0.35\textwidth]{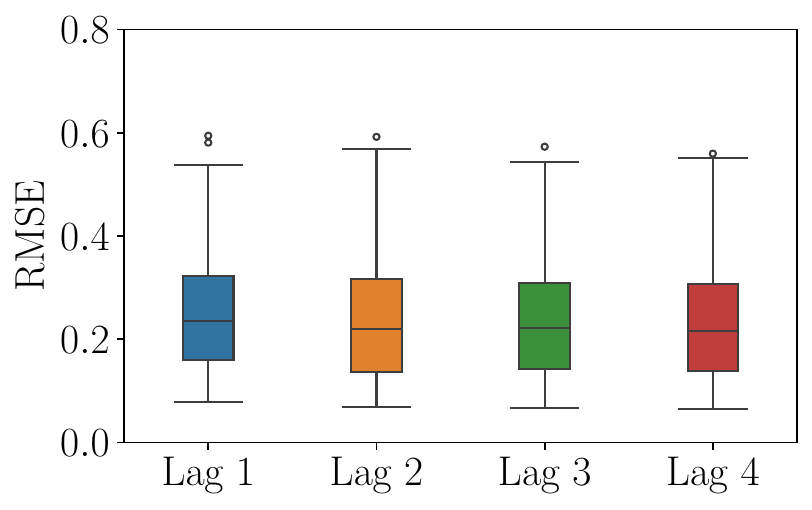}
    \hspace{0.04\textwidth}
    \includegraphics[width=0.35\textwidth]{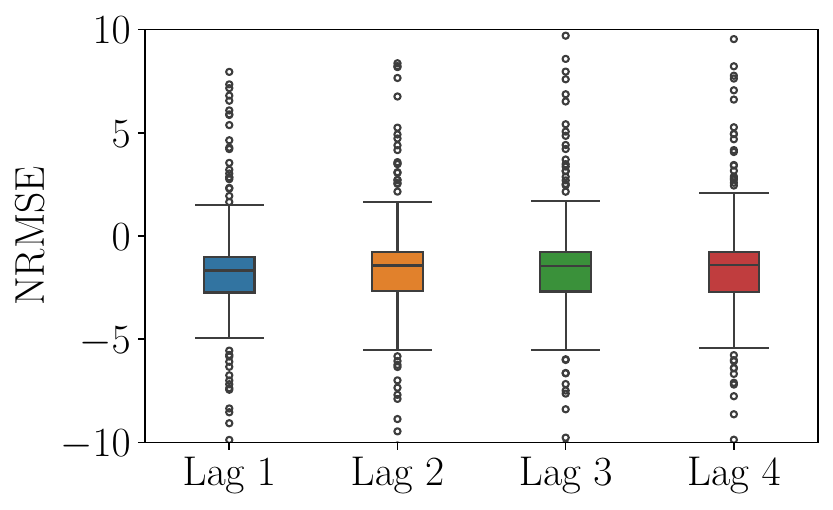}
    
    \vskip\baselineskip
    
    \includegraphics[width=0.35\textwidth]{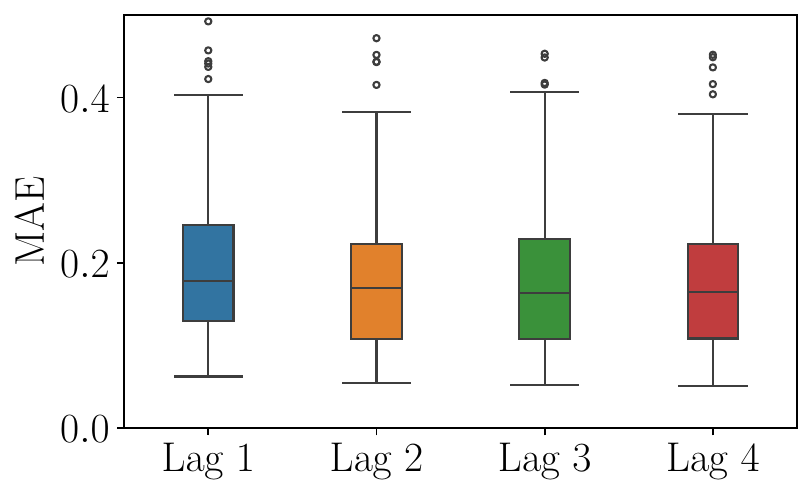}
    \hspace{0.04\textwidth}
    \includegraphics[width=0.35\textwidth]{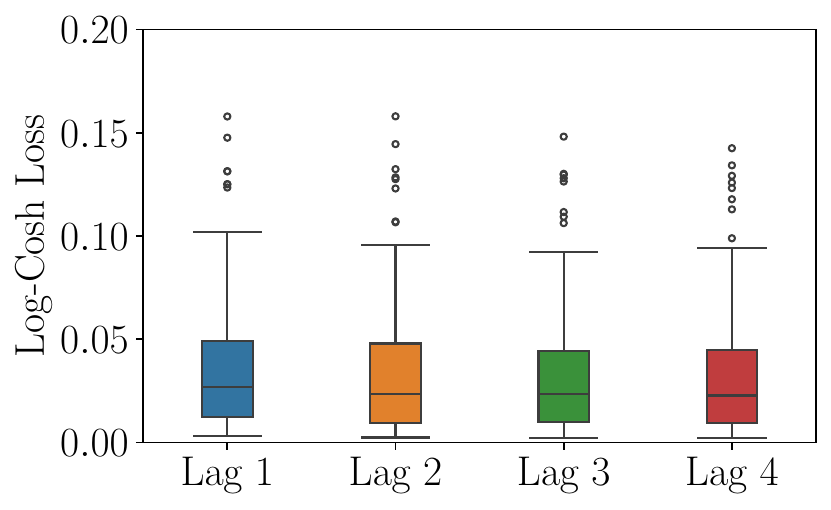}
    \caption{Prediction error distributions on the validation set for XGBoost across the four evaluation metrics.}
    \label{fig:xgb_val_errors}
\end{figure}
\FloatBarrier

\begin{figure}[h]
    \centering
    \includegraphics[width=0.35\textwidth]{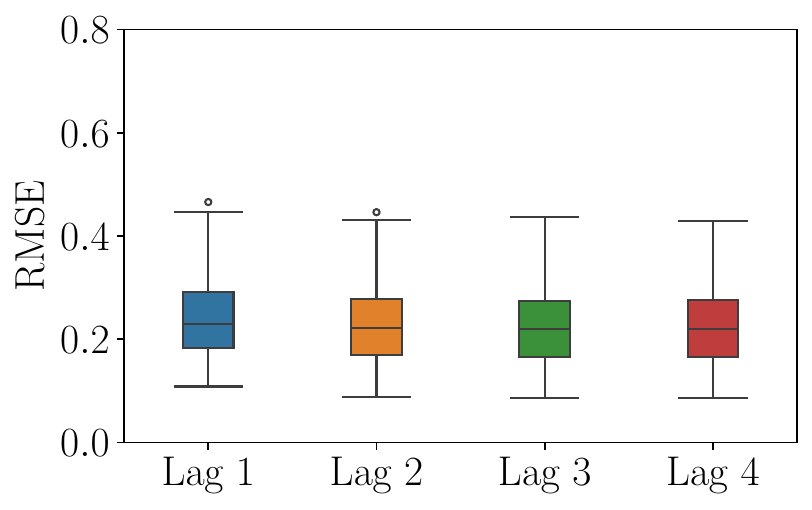}
    \hspace{0.04\textwidth}
    \includegraphics[width=0.35\textwidth]{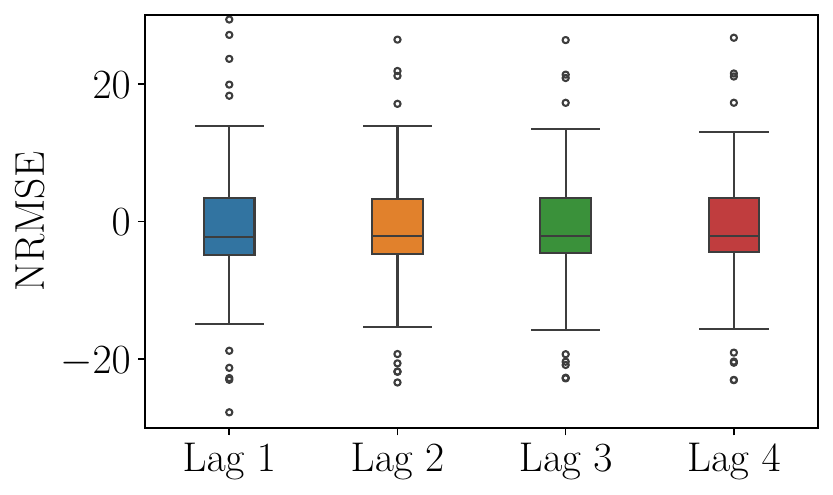}
    
    \vskip\baselineskip
    
    \includegraphics[width=0.35\textwidth]{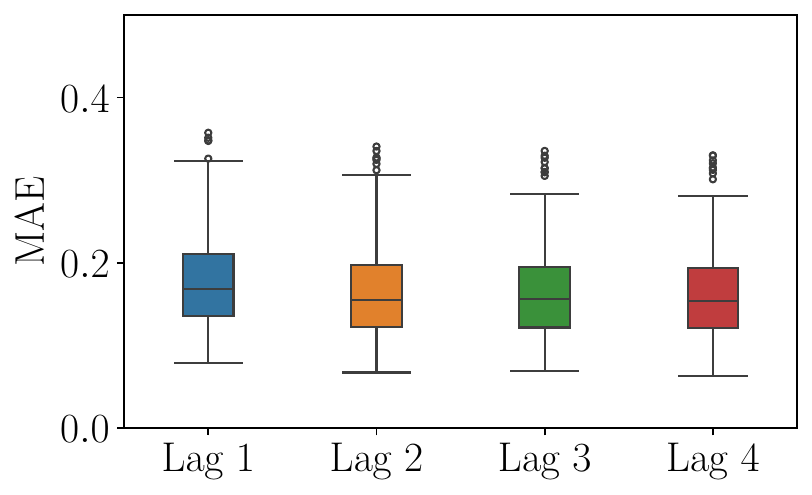}
    \hspace{0.04\textwidth}
    \includegraphics[width=0.35\textwidth]{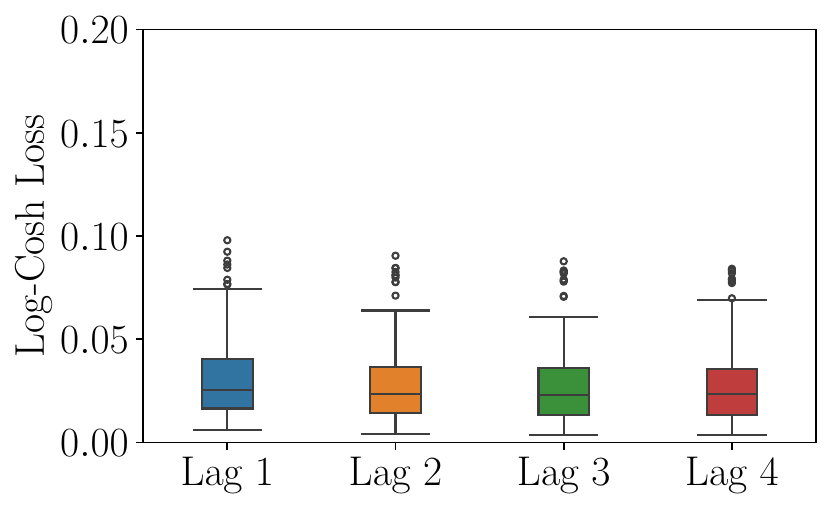}
    \caption{Prediction error distributions on the test set for XGBoost across the four evaluation metrics.}
    \label{fig:xgb_test_errors}
\end{figure}
\FloatBarrier

\begin{figure}[h]
    \centering
    \includegraphics[width=0.35\textwidth]{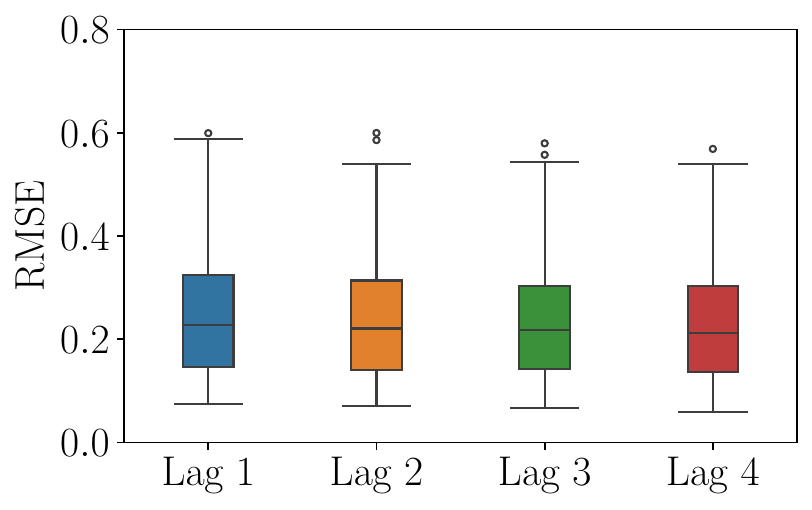}
    \hspace{0.04\textwidth}
    \includegraphics[width=0.35\textwidth]{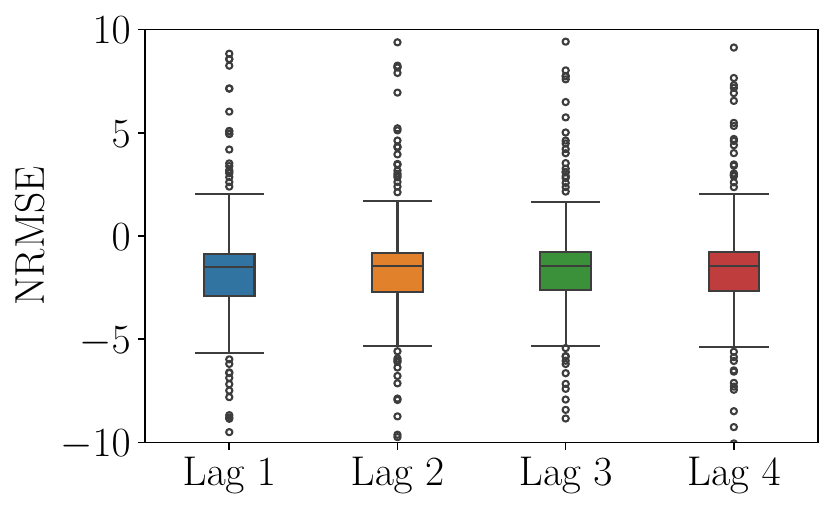}
    
    \vskip\baselineskip
    
    \includegraphics[width=0.35\textwidth]{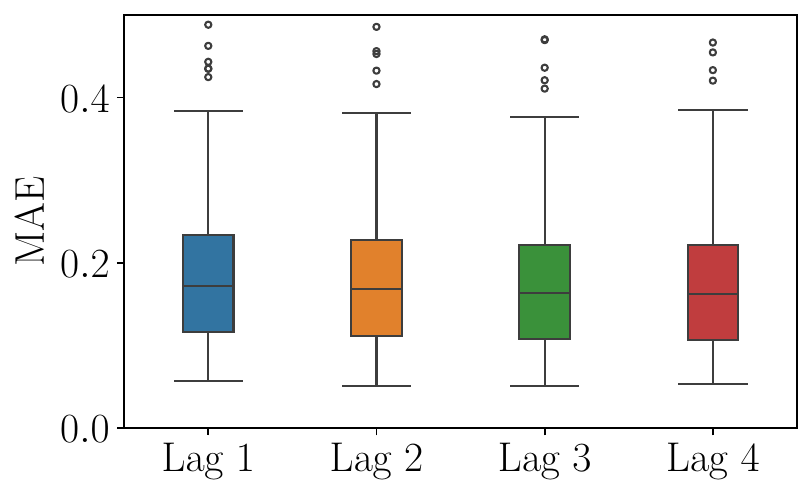}
    \hspace{0.04\textwidth}
    \includegraphics[width=0.35\textwidth]{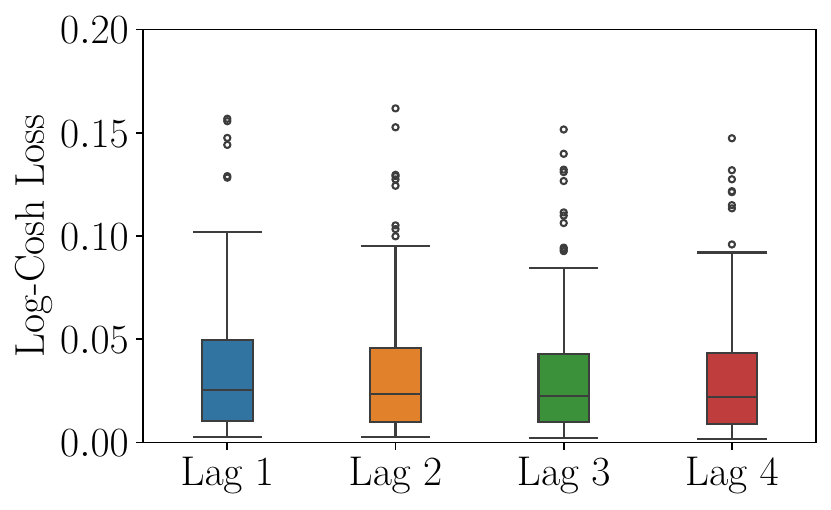}
    \caption{Prediction error distributions on the validation set for LightGBM across the four evaluation metrics.}
    \label{fig:lightgbm_val_errors}
\end{figure}
\FloatBarrier

\begin{figure}[h]
    \centering
    \includegraphics[width=0.35\textwidth]{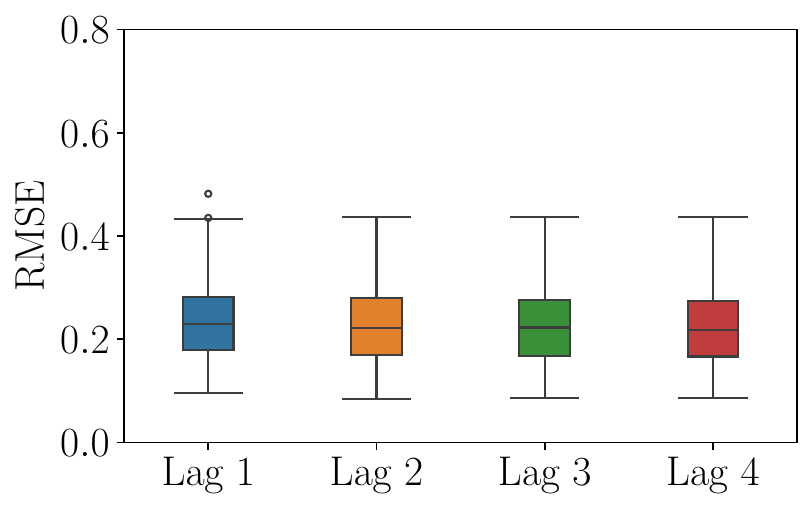}
    \hspace{0.04\textwidth}
    \includegraphics[width=0.35\textwidth]{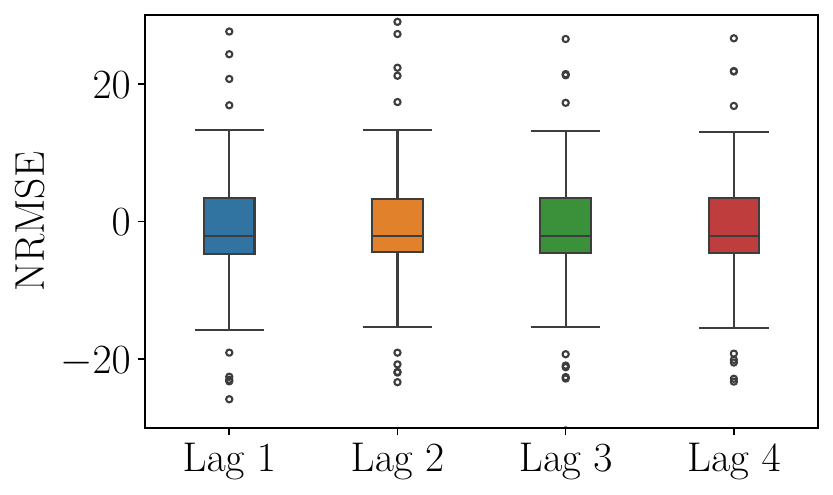}
    
    \vskip\baselineskip
    
    \includegraphics[width=0.35\textwidth]{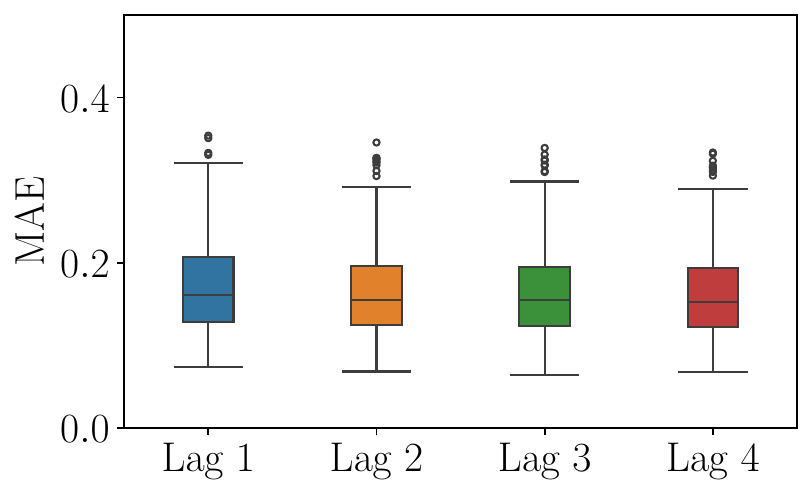}
    \hspace{0.04\textwidth}
    \includegraphics[width=0.35\textwidth]{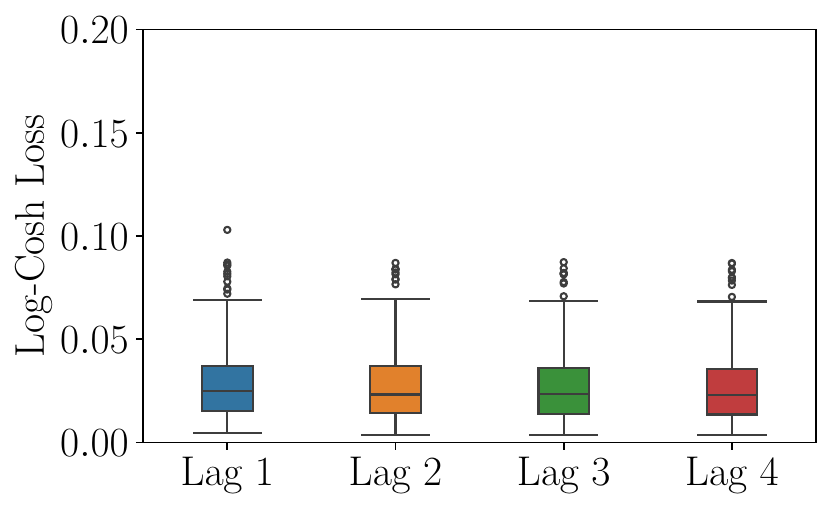}
    \caption{Prediction error distributions on the test set for LightGBM across the four evaluation metrics.}
    \label{fig:lightgbm_test_errors}
\end{figure}
\FloatBarrier

\begin{figure}[h]
    \centering
    \includegraphics[width=0.35\textwidth]{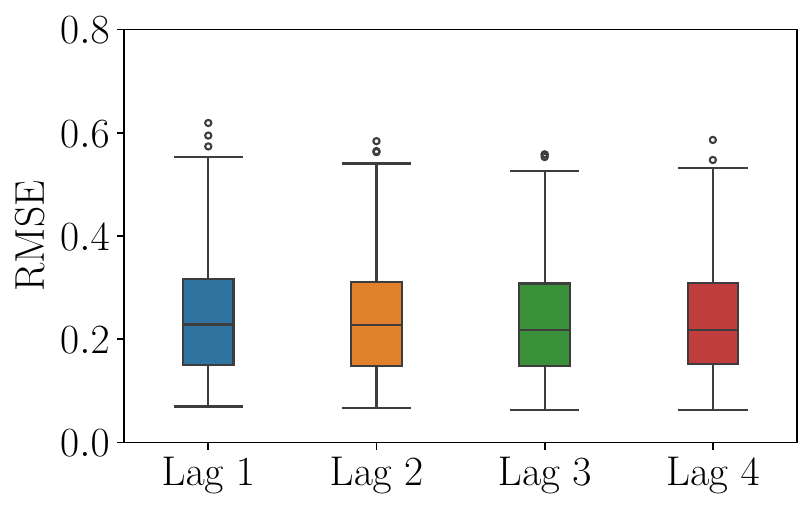}
    \hspace{0.04\textwidth}
    \includegraphics[width=0.35\textwidth]{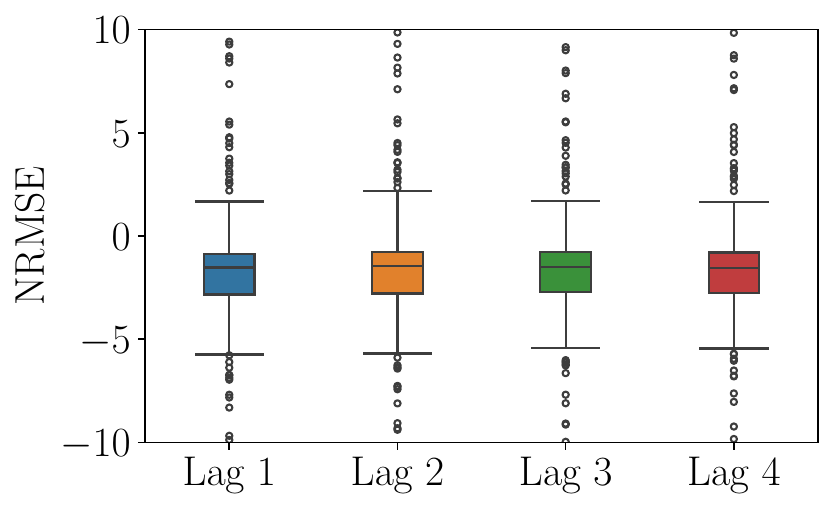}
    
    \vskip\baselineskip
    
    \includegraphics[width=0.35\textwidth]{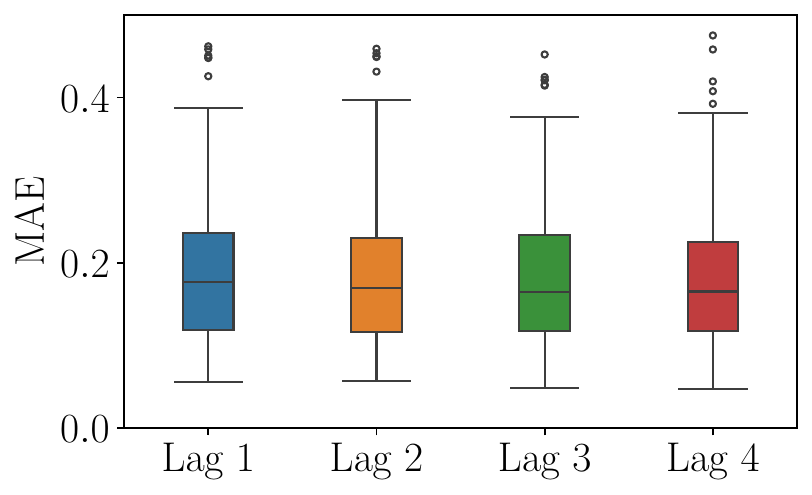}
    \hspace{0.04\textwidth}
    \includegraphics[width=0.35\textwidth]{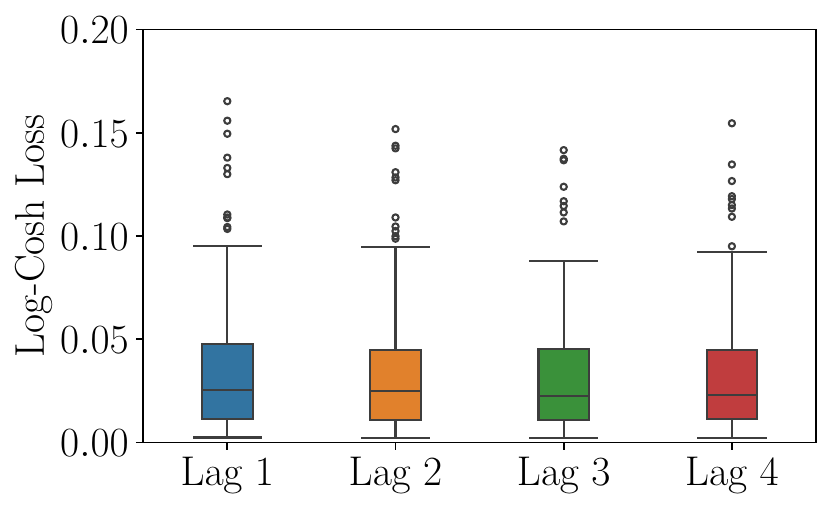}
    \caption{Prediction error distributions on the validation set for NN across the four evaluation metrics.}
    \label{fig:nn_val_errors}
\end{figure}
\FloatBarrier

\begin{figure}[h]
    \centering
    \includegraphics[width=0.35\textwidth]{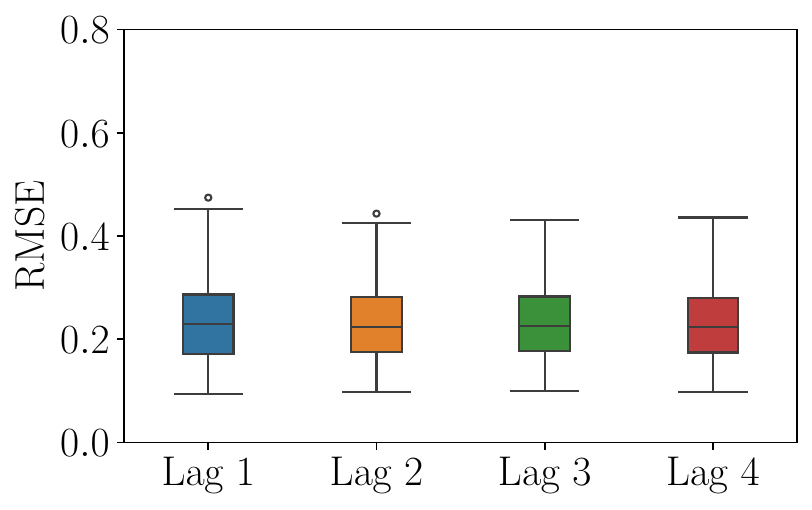}
    \hspace{0.04\textwidth}
    \includegraphics[width=0.35\textwidth]{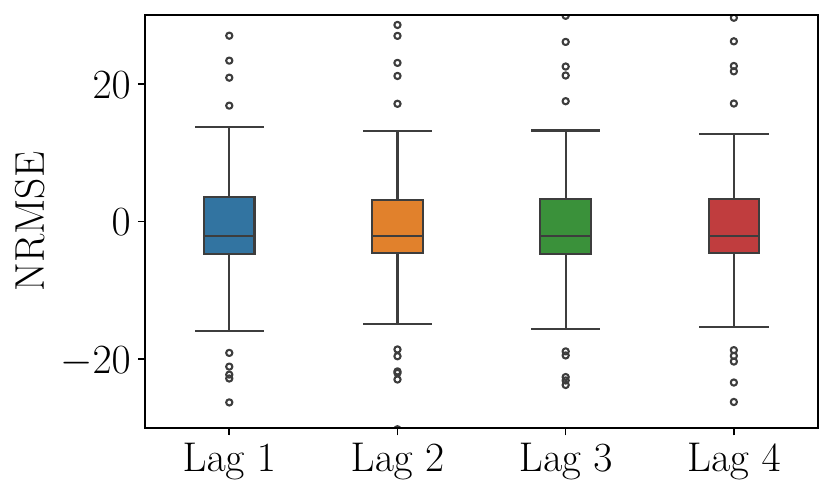}
    
    \vskip\baselineskip
    
    \includegraphics[width=0.35\textwidth]{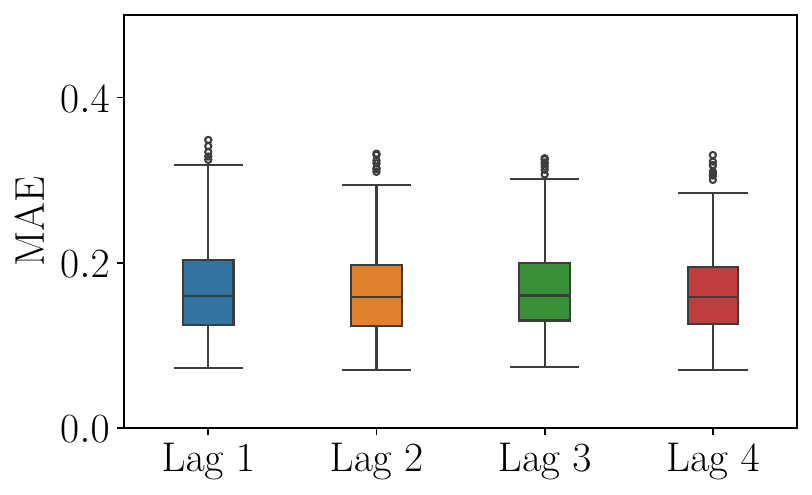}
    \hspace{0.04\textwidth}
    \includegraphics[width=0.35\textwidth]{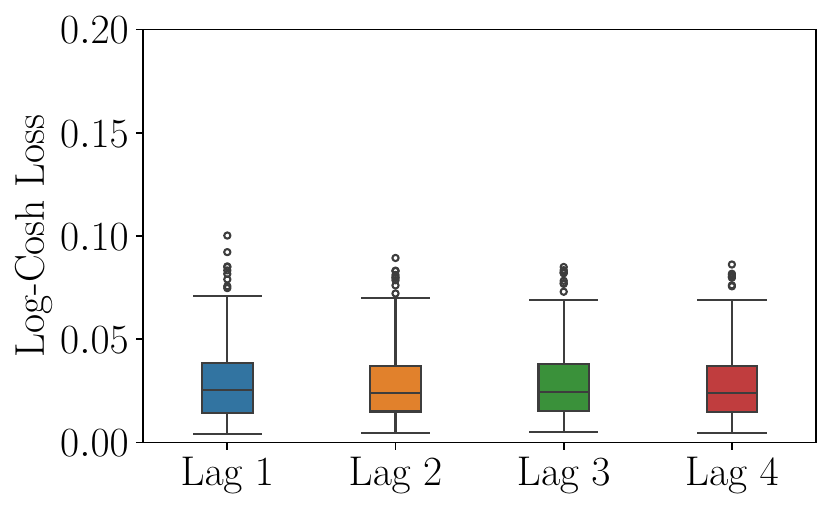}
    \caption{Prediction error distributions on the test set for NN across the four evaluation metrics.}
    \label{fig:nn_test_errors}
\end{figure}
\FloatBarrier

\end{document}